\documentclass{article}

\usepackage[utf8]{inputenc}
\usepackage[colorlinks = true,
            linkcolor = blue,
            urlcolor  = blue,
            citecolor = blue,
            anchorcolor = blue]{hyperref}
\usepackage{amsmath,amssymb,amsfonts,amsthm,bm}
\usepackage{color,graphicx}
\usepackage[a4paper, total={6.5in,8.5in}]{geometry}

\usepackage{amssymb}
\usepackage{empheq}
\usepackage{mathtools}
\usepackage{graphicx}
\usepackage{bbm}
\usepackage{hyperref}
\usepackage{dirtytalk}
\usepackage{stmaryrd}
\usepackage{fancyhdr}
\usepackage{microtype}
\usepackage{graphicx}
\usepackage{subcaption}
\usepackage{booktabs}
\usepackage{makecell}
\usepackage{caption}
\usepackage{amsmath}
\usepackage{booktabs} 
\usepackage{tablefootnote}
\usepackage{footnote}
\usepackage{hyperref}
\usepackage{tabularx}
\usepackage[ruled,vlined]{algorithm2e}
\usepackage{epigraph}

\usepackage{verbatim}
\usepackage{listings}
\usepackage{algpseudocode}
\usepackage[affil-it]{authblk} 
\usepackage{textcomp}
\usepackage{multirow}
\usepackage[toc,page]{appendix}
\usepackage{xspace}
\usepackage{wrapfig}
\usepackage[numbers, sort, comma, square]{natbib}
\usepackage{float}
\usepackage{todonotes}

\theoremstyle{plain}
\newtheorem{theorem}{Theorem}[section]

\newtheorem{lemma}[theorem]{Lemma}
\newtheorem{corollary}[theorem]{Corollary}
\theoremstyle{definition}

\theoremstyle{remark}

\newcommand{\Hquad}{\hspace{0.5em}} 

\DeclareMathOperator*{\argmax}{arg\,max}
\DeclareMathOperator*{\argmin}{arg\,min}
\DeclareMathOperator*{\sign}{sign}

\newcommand{\segmentation}{seg\xspace}
\newcommand{\epscutoff}{\tilde{\epsilon}}

\newcommand{\randvarphi}{\tilde{\varphi}}
\newcommand{\varphimin}{\varphi_{\text{min}}}
\newcommand{\varphimax}{\varphi_{\text{max}}}

\newcommand{\suscept}[1]{\text{Susc}(#1;\epstest)}
\newcommand{\motionblurkernel}{M}
\newcommand{\thetastar}{\theta^\star}
\newcommand{\nameofattack}{directed attack\xspace}
\newcommand{\nameofattacks}{directed attacks\xspace}

\newcommand{\nameofattackscapital}{Directed attacks\xspace}
\newcommand{\numlabels}{K}
\newcommand{\roberr}[1]{\text{Err}(#1;\epstest)}
\newcommand{\stderr}[1]{\text{Err}(#1;0)}

\newcommand{\sigseptest}{\sigsep_{\text{test}}}
\newcommand{\dims}{d}
\newcommand{\numsamp}{n}
\newcommand{\xnonsig}{\tilde{x}}
\newcommand{\marginnonsig}{\tilde{\gamma}}
\newcommand{\eps}{\epsilon}

\newcommand{\robacc}[1]{\text{Acc}(#1;\epstest)}
\newcommand{\Loss}{\mathcal{L}}
\newcommand{\robloss}[1]{\Loss_{\epstrain}(#1)}
\newcommand{\prob}{\mathbb{P}}

\newcommand{\EE}{\mathbb{E}}
\newcommand{\thetahat}[1]{\widehat{\theta}^{#1}}

\newcommand{\thetatrue}{\theta^{\star}}
\newcommand{\pertset}[2]{T({#1};{#2})}
\newcommand{\Indi}[1]{\mathbb{I}\{#1\}}
\newcommand{\epstrain}{\eps_{\text{tr}}}
\newcommand{\epstest}{\eps_{\text{te}}}
\newcommand{\mixvar}{\sigma}
\newcommand{\E}{\text{e}}
\newcommand{\data}{D}
\newcommand{\datanonsig}{\widetilde{D}}
\newcommand{\tconst}{\alpha}
\newcommand{\maxmargin}{\tilde{\gamma}_{\max}}
\newcommand{\minmargin}{\tilde{\gamma}_{\min}}
\newcommand{\sigsep}{r}
\newcommand{\thetatilde}{\tilde{\theta}}
\newcommand{\Dshift}{D_{\epstrain}}
\newcommand{\thetaA}{\thetahat{\epstrain}}

\newcommand{\maxind}{j^\star}
\newcommand{\maxindA}{\maxind}

\newcommand{\xind}[1]{x_{[#1]}}
\newcommand{\thetaind}[1]{\theta_{[#1]}}
\newcommand{\thetahatind}[2]{\widehat{\theta}_{[#2]}^{#1}}
\newcommand{\indof}[2]{{#1}_{[#2]}}

\newcommand{\robness}[1]{\text{Rob}(#1;\epstest)}
\newcommand{\R}{\mathbb{R}}
\newcommand{\Normal}{\mathcal{N}}
\newcommand{\loss}{L}

\newcommand{\cst}{c}

\newcommand{\constone}{a_1}
\newcommand{\constwo}{a_2}
\newcommand{\decplanegen}[1]{\mathcal{H}(#1)}
\newcommand{\randdatamatr}{X}

\newcommand{\gaussianmatrix}{Q}

\newif\ifshownotes
\shownotestrue

\newif\ifarxiv
\arxivfalse

\usepackage{longtable}
\date{}
\author[]{Jacob Clarysse}
\author[]{Julia Hörrmann}
\author[]{Fanny Yang}
\affil[]{Department of Computer Science, ETH Zürich}

\title{Why adversarial training can hurt robust accuracy}

\begin{document}

\maketitle
\setcounter{page}{1}

\begin{abstract}
Machine learning classifiers with high test accuracy often perform
poorly under adversarial attacks.
It is commonly believed that 
adversarial training 
alleviates this issue.
In this paper, we demonstrate that,
surprisingly, the opposite may be true --- Even though adversarial training helps when enough data is  available, it may hurt robust generalization in the small sample size regime. 
We first prove this phenomenon for a high-dimensional linear
classification setting with noiseless observations. Our proof provides explanatory insights that may also transfer to feature learning models. 
Further, we observe in experiments on standard image datasets that the same behavior occurs 
for perceptible attacks
that effectively reduce class information such as mask attacks and object corruptions. 
\end{abstract}

\section{Introduction}
\label{sec:intro}
\begin{wrapfigure}{r}{0.43\textwidth}
\centering
\vspace{-0.1in}
\includegraphics[width=0.99\linewidth]{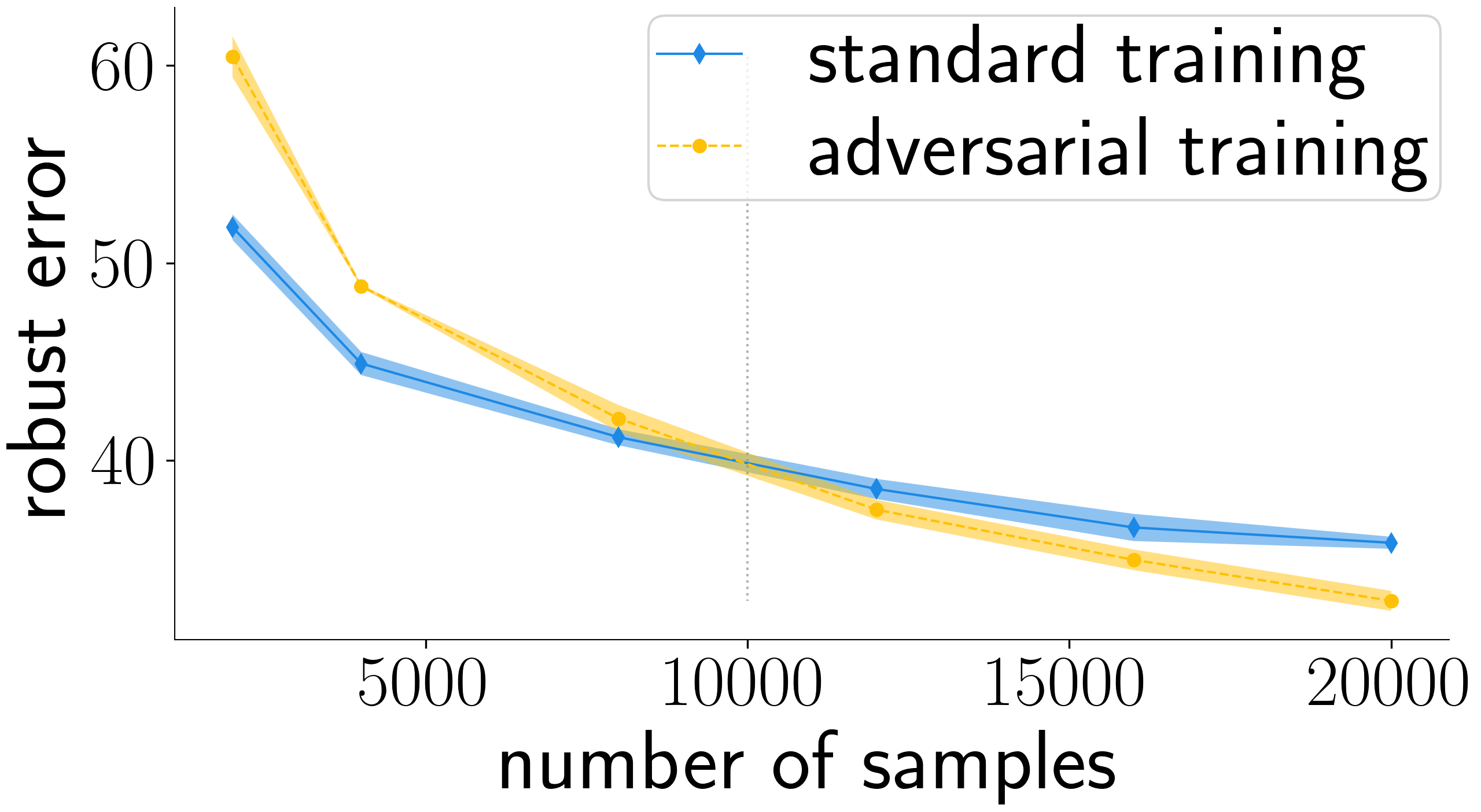}
\caption{
  On subsampled \mbox{CIFAR10} attacked by $2\times 2$ masks, adversarial training yields higher robust error than standard training
  when the sample size is small, even though it helps for large sample sizes.
  (see Sec.~\ref{sec:app_cifar10} for details).}
  \vspace{-0.2in}
\label{fig:teaserplot}
\end{wrapfigure}

Today's best-performing classifiers are vulnerable to adversarial attacks
\cite{goodfellow15, szegedy14} and exhibit high \emph{robust error}: for many inputs, their predictions change under adversarial perturbations,
even though the true class stays the same. 
For example, in image classification tasks, we distinguish between two categories of
such attacks that are content-preserving \cite{gilmer18b} (or consistent \cite{raghunathan20}) if their strength is limited --- perceptible and imperceptible perturbations.
Most work to date studies imperceptible attacks such as 
bounded $\ell_p$-norm perturbations \cite{goodfellow15, madry18, moosavi16}, small transformations using image processing
techniques \cite{ghiasi19, zhao20, laidlaw21, Luo18} or 
nearby samples on the data manifold \cite{Lin20, Zhou20}.
They can often use their limited budget to successfully fool a learned classifier but, by definition, do not visibly reduce information about the actual class: the object in the perturbed image looks exactly the same as in the original version.

\begin{figure}[t]
\vskip 0.2in
\begin{center}
\begin{subfigure}[b]{0.2\textwidth}
  \includegraphics[width=0.99\linewidth]{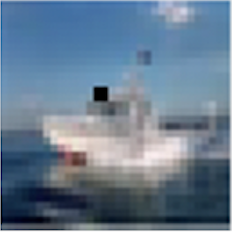}
  \caption{Masking}
  \label{fig:CIFAR10_boat}
\end{subfigure}
\begin{subfigure}[b]{0.2\textwidth}
  \includegraphics[width=0.99\linewidth]{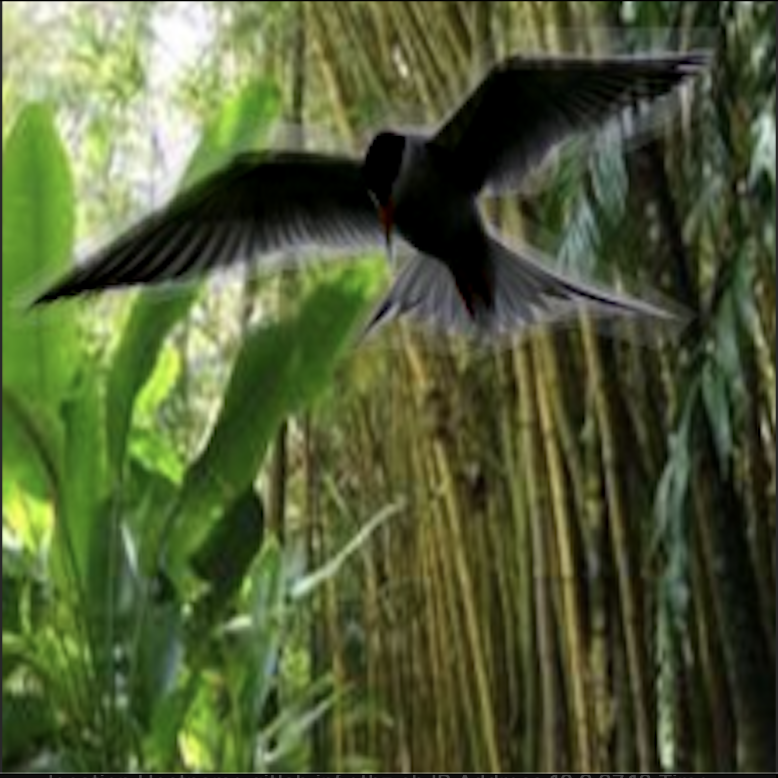}
  \caption{Illumination}
  \label{fig:WB_light_dark}
\end{subfigure}
\begin{subfigure}[b]{0.2\textwidth}
  \includegraphics[width=0.99\linewidth]{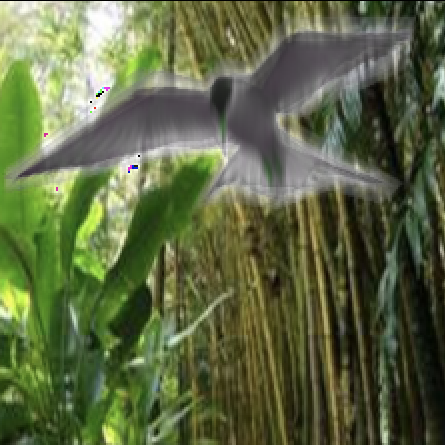}
  \caption{Motion blur}
  \label{fig:WB_motion_blur}
\end{subfigure}
\begin{subfigure}[b]{0.2\textwidth}
  \includegraphics[width=0.99\linewidth]{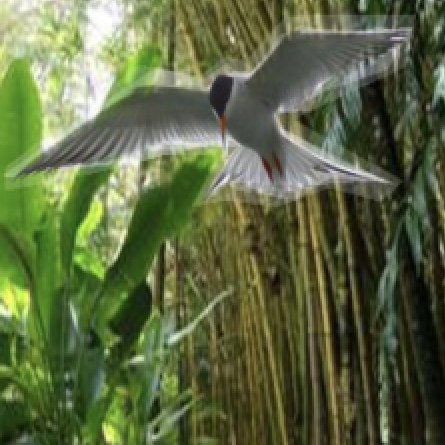}
  \caption{Original}
  \label{fig:fig:WB_original}
\end{subfigure}
\caption{Examples of \nameofattacks on CIFAR10 and the
  Waterbirds dataset. In Figure \ref{fig:CIFAR10_boat}, we corrupt the image with a black mask of size $2 \times 2$ and in Figure \ref{fig:WB_light_dark} and \ref{fig:WB_motion_blur} we change the lighting conditions (darkening) and apply motion blur on the bird in the image respectively. 
  All perturbations effectively reduce the information about the class in the images: they are the result of \nameofattacks.}
\label{fig:sig_att_examples}
\end{center}
\vskip -0.2in
\end{figure}

On the other hand, perceptible perturbations may occur more naturally in practice or are physically realizable. 
For example, stickers can be placed on traffic signs \cite{Eykholt18},
masks of different sizes may cover important features of human faces
\cite{Wu20}, images might be rotated or translated \cite{Logan19},
animals in motion may appear blurred in photographs
depending on the shutter speed, or the lighting conditions could be poor (see Figure~\ref{fig:sig_att_examples}).
Some perceptible attacks can effectively
use the perturbation budget to reduce actual class information
in the input (the \emph{signal}) while still preserving the original class.
For example, a stop sign with a small sticker doesn't lose its semantic meaning
or a flying bird does not become a different species because it induces motion blur
in the image.
We refer to these attacks as \emph{\nameofattacks} (see
Section~\ref{sec:robustness} for a more formal
characterization). 

In this paper, we
demonstrate that one of the most common beliefs 
for adversarial attacks does not transfer to \nameofattacks, in
particular when the sample size is small. Specifically, it is widely acknowledged that adversarial training often achieves significantly lower adversarial error than standard
training. This holds in particular if 
the perturbation type 
\cite{madry18, zhang19, Bai21} and perturbation budget match the attack during test time. 
Intuitively, the improvement is a result of decreased
\emph{attack-susceptibility}: independent of the true class,
adversarial training explicitly encourages the classifier to predict
the same class for all perturbed points.

In this paper, we question the efficacy of adversarial training to increase
robust accuracy for \nameofattacks.
In particular, we show that adversarial training not only increases standard test error as noted in \cite{zhang19, tsipras19, Stutz19, raghunathan20}), but surprisingly,
\begin{center}
 \emph{adversarial training may even increase the robust test error compared to standard training!}
\end{center}
Figure \ref{fig:teaserplot} illustrates the main message of our paper for CIFAR10 subsets: Although adversarial training
outperforms standard training when enough training samples are available, it is inferior 
in the low-sample regime.  
More specifically, our contributions are as follows:
\begin{itemize}
\item We prove that, almost surely, adversarially training a linear classifier on separable data yields a monotonically increasing robust error as the perturbation budget grows. 
We further establish high-probability non-asymptotic lower bounds on the robust error gap between adversarial and standard training.
\item Our proof provides intuition for why this phenomenon is particularly prominent for \nameofattacks in the small sample size regime. 
\item We show that this phenomenon occurs on a variety of real-world datasets and perceptible \nameofattacks in the small sample size regime.
\end{itemize}


\section{Robust classification}
\label{sec:robustness}


We first introduce our robust classification setting more formally by defining
the notions of  adversarial robustness, \nameofattacks and adversarial training
used throughout the paper.

\paragraph{Adversarially robust classifiers}

For inputs $x \in \R^\dims$, we consider multi-class classifiers
associated with parameterized functions $f_\theta:\R^\dims \to
\R^\numlabels$, where $\numlabels$ is the number of labels. In the special case of binary classification ($\numlabels = 2$), we use the output predictions $y=\textrm{sign}(f_\theta(x))$. For example, $f_\theta(x)$ could be linear models (as in Section~\ref{sec:theoryresults}) or
neural networks (as in Section~\ref{sec:realworldexpapp}).

One key step to encourage deployment of machine learning based classification in real-world applications, is to increase the
robustness of classifiers against perturbations that do not
change the ground truth label. 
Mathematically speaking, we would like to have a small
\emph{$\epstest$-robust error}, defined as
\begin{equation}
  \label{eq:roberr}
  \roberr{\theta} := \EE_{(x, y)\sim \prob} \max_{x' \in \pertset{x}{\epstest}} \ell(f_\theta (x'),y),
\end{equation}
where $\ell$ is the multi-class zero-one loss, which only equals $1$ if the predicted
output using $f_\theta(x)$ does not match the true label $y$.
Further, $\pertset{x}{\epstest}$ is a perturbation set associated with a \emph{transformation type} and size $\epstest$. 
Note that the \emph{(standard) error} of a classifier corresponds to evaluating $\roberr{\theta}$ at $\epstest = 0$, yielding the standard error $\stderr{\theta} =\EE_{(x, y)\sim \prob} \ell(f_\theta (x),y)$.

\paragraph{(Signal)-Directed attacks}
Most works in the existing literature consider consistent perturbations where
$\epstest$ is small enough such that all samples in the perturbation set
have the same ground truth or expert label. 
Note that the ground truth model $f_{\theta^{\star}}$ is therefore robust against perturbations and achieves the same error for standard and adversarial evaluation. 
The inner maximization in Equation~\eqref{eq:roberr} is often called the adversarial \emph{attack} of the model $f_\theta$ and the corresponding solution is referred to as the adversarial example.
In this paper, we consider \emph{\nameofattacks}, as described in Section~\ref{sec:intro}, that effectively reduce the information about the ground truth classes.
Formally, we characterize \emph{\nameofattacks} by the following property: 
for any model $f_\theta$ with low standard error, the corresponding adversarial example is well-aligned with the adversarial example found using the ground truth model 
$f_{\theta^{\star}}$.
An example for such an attack are additive perturbations that are constrained to the direction of the ground truth decision boundary.  We provide concrete examples for linear classification in  Section~\ref{logreg_linear_model}.

\paragraph{Adversarial training}



In order to obtain classifiers with a good robust accuracy, it is
common practice to minimize a (robust) training objective $\mathcal{L}_{\epstrain}$ with a surrogate
classification loss $\loss$ such as
\begin{equation}
  \label{eq:emploss}
  \robloss{\theta} :=  \frac{1}{n} \sum_{i=1}^n \max_{x_i' \in \pertset{x_i}{\epstrain}} \loss(f_\theta(x_i') y_i),
\end{equation}
which is called adversarial training.  In practice, we often use the
cross entropy loss $\loss(z) = \log (1+ \E^{-z})$ and minimize the
robust objective by using first order optimization methods such as
(stochastic) gradient descent.  SGD is also the algorithm that we
focus on in both the theoretical and experimental sections.

When the desired type of robustness is known in advance, it is
standard practice to use the same perturbation set for training as for
testing, i.e. $\pertset{x}{\epstrain}=\pertset{x}{\epstest}$. For example, \citet{madry18} shows that the robust error sharply increases for $\epstrain < \epstest$.
In this paper, we show that for \nameofattacks in the small sample size regime, in fact, the opposite is true.


\section{Theoretical results}
\label{sec:theoryresults}
In this section, we prove for linear functions $f_\theta(x) =
\theta^\top x$ that  in the case of directed attacks, robust
generalization deteriorates with increasing $\epstrain$.
The proof, albeit in a simple setting, provides
explanations for why adversarial training fails in the
high-dimensional regime for such attacks.

\subsection{Setting}
\label{logreg_linear_model}

We now introduce the precise linear setting used in our theoretical results.




\paragraph{Data model}
In this section, we assume that the ground truth and hypothesis class
are given by linear functions $f_\theta(x) = \theta^\top x$ and the
sample size $\numsamp$ is lower than the ambient dimension $\dims$.  In
particular, the generative distribution $\prob_\sigsep$ is similar to
\cite{tsipras19, kolter19}: The label $y \in \{+1, -1\}$ is drawn with
equal probability and the covariate vector is sampled as $x =
[y\frac{\sigsep}{2}, \xnonsig]$ with the random vector $\xnonsig \in
\R^{\dims-1}$ drawn from a standard normal distribution,
i.e. $\xnonsig \sim \Normal(0, \sigma^2 I_{d-1})$. We would like to
learn a classifier that has low robust error by using a dataset
$\data = {(x_i, y_i)}_{i=1}^n$ with $\numsamp$ i.i.d. samples from
$\prob_{\sigsep}$.

Notice that the distribution $\prob_{\sigsep}$ is noiseless: for a given input
$x$, the label $y = \sign(\xind{1})$ is deterministic. Further, the
optimal linear classifier (also referred to as the \emph{ground
  truth}) is parameterized by $\thetatrue = e_1$.\footnote{Note that the result more generally holds for non-sparse models that are not axis aligned by way of a simple rotation $z = U x$. In that case the distribution is characterized by $\thetastar = u_1$ and a rotated Gaussian in the $\dims-1$ dimensions orthogonal to $\thetastar$.} By definition, the ground truth is
robust against all consistent perturbations and hence the optimal
robust classifier.



\paragraph{\nameofattackscapital}  
The focus in this paper lies on consistent \nameofattacks that by
definition efficiently concentrate their attack budget in the
direction of the signal.  For our linear setting, we can model such
attacks by  additive perturbations in the first dimension
\begin{equation}
  \label{eq:linfmaxpert}
  \pertset{x}{\eps} = \{x'=x+\delta  \mid \delta = \beta e_1 \text{ and } -\eps \leq \beta\leq \eps\}.
\end{equation}
Note that this attack is always in the direction of the true signal dimension, i.e. the ground truth. Furthermore, when  $\epsilon < \frac{r}{2}$, it is a consistent \nameofattack.
Observe how this is different from $\ell_p$ attacks - an $\ell_p$ attack, depending on the model, may add a perturbation that only has a very small component in the signal direction. 


\paragraph{Robust max-$\ell_2$-margin classifier}

A long line of work studies the implicit bias of interpolators
that result from applying stochastic gradient descent on the logistic loss until convergence \cite{liu20, Ji19, Chizat20, nacson19}.
For linear models, we obtain the $\epstrain$-robust maximum-$\ell_2$-margin solution (\emph{robust max-margin} in short) 
\begin{equation}
  \label{eq:maxmargin}
  \thetahat{\epstrain} := \argmax_{\|\theta\|_2\leq 1} \min_{i\in [n], x_i' \in \pertset{x_i}{\epstrain}} y_i \theta^\top x_i'.
\end{equation}
This can for example be shown by a simple rescaling
argument using Theorem 3.4 in \cite{liu20}.  Even though our result is proven for the max-$\ell_2$-margin classifier,
it can easily be extended to other interpolators.

\begin{figure*}[!t]
  \centering
\begin{subfigure}[b]{0.3\textwidth}
  \includegraphics[width=0.99\linewidth]{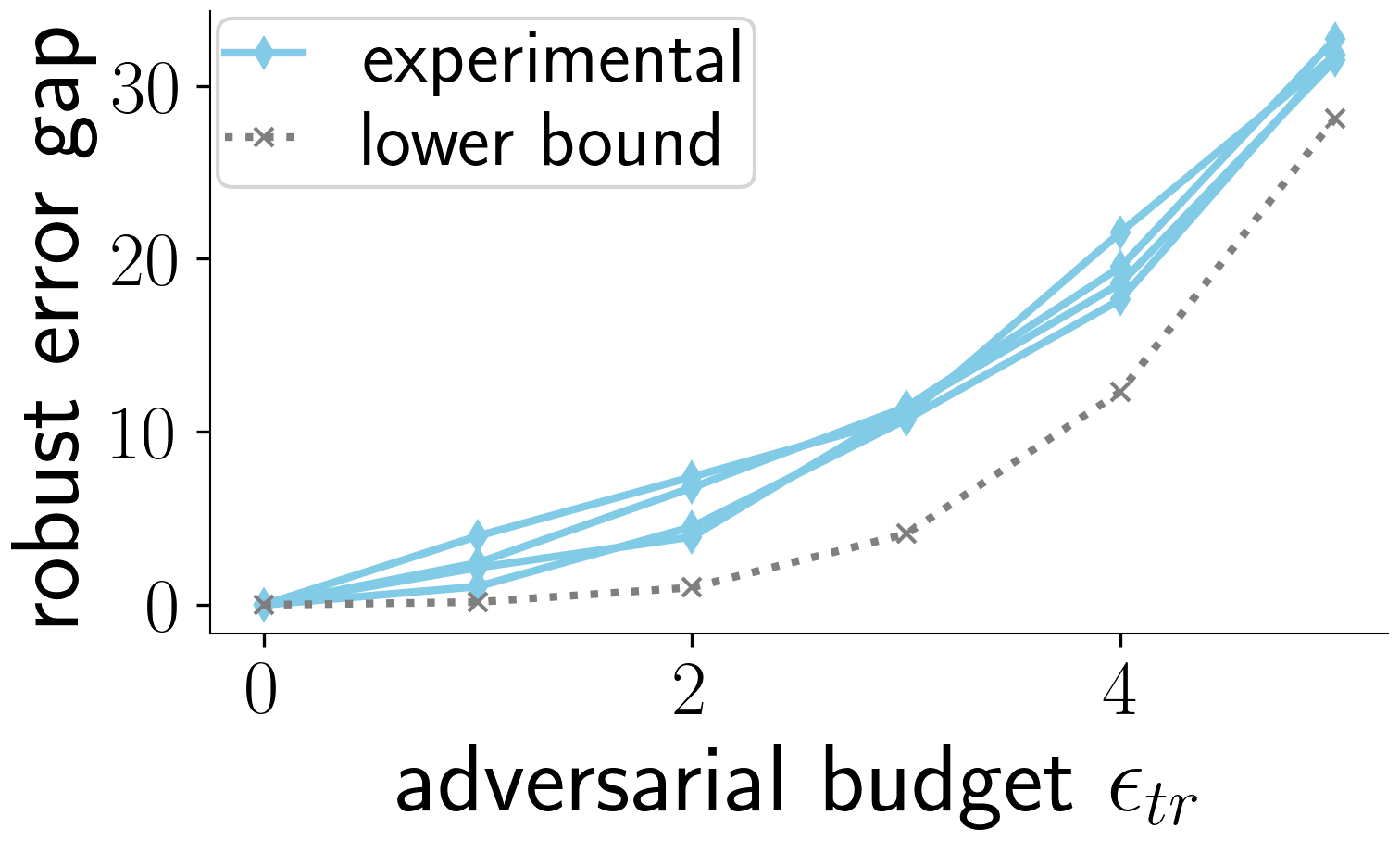}
  \caption{Robust error increase with $\epstrain$}
  \label{fig:main_lower_bound_eps}
\end{subfigure}
\begin{subfigure}[b]{0.3\textwidth}
  \includegraphics[width=0.99\linewidth]{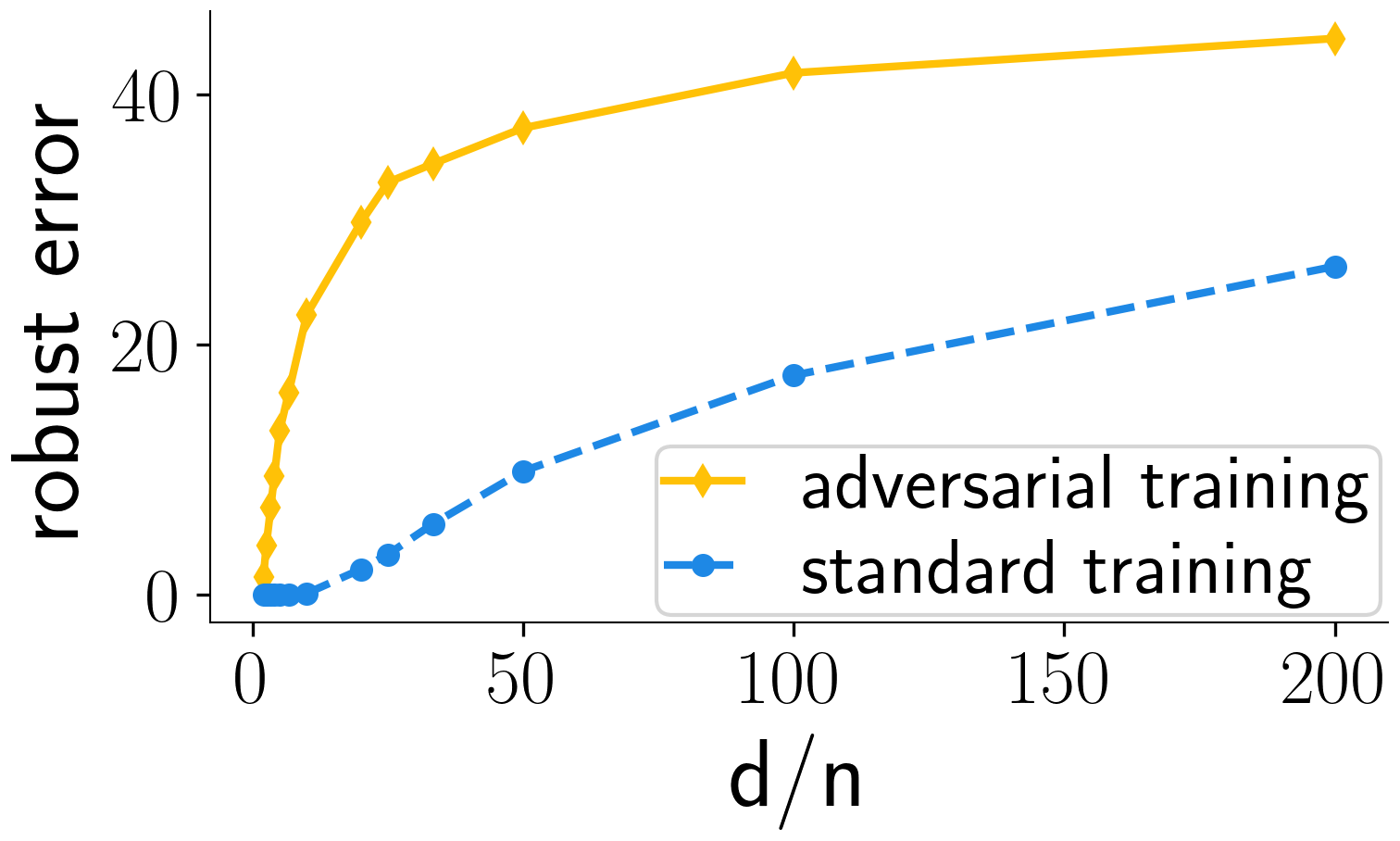}
  \caption{Standard-adversarial training}
  \label{fig:main_numobs}
\end{subfigure}
\begin{subfigure}[b]{0.3\textwidth}
  \includegraphics[width=0.99\linewidth]{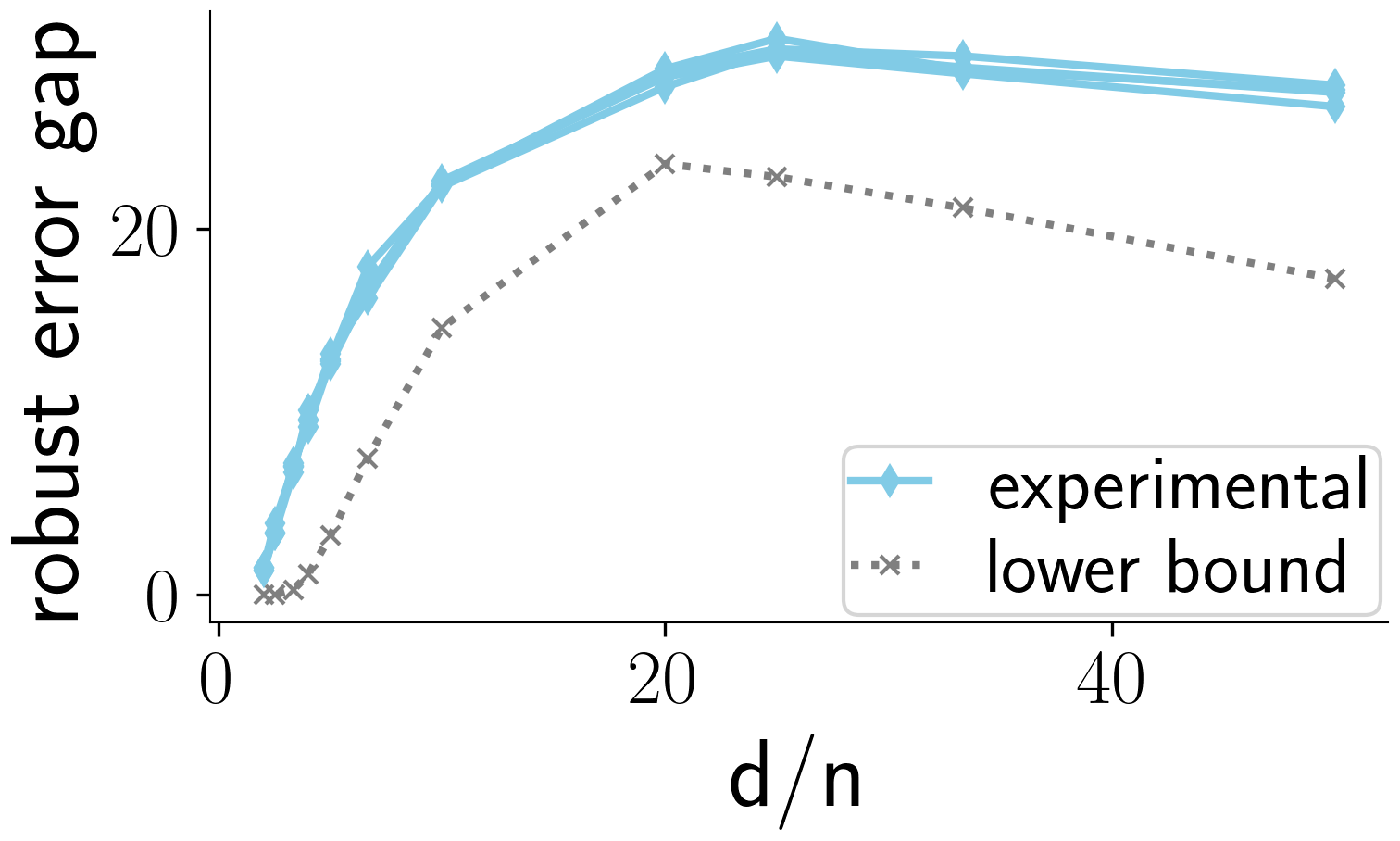}
  \caption{Effect of over-parameterization}
  \label{fig:main_numobs_bound}
\end{subfigure}
\caption{Experimental verification of Theorem \ref{thm:linlinf}.
(a) We set $\dims = 1000$, $\sigsep = 12$, $\numsamp = 50$ and plot the robust error gap between standard and adversarial training with increasing adversarial budget $\epstrain$ of $5$ independent experiments. For comparison, we also plot the lower bound given in Theorem \ref{thm:linlinf}. In (b) and (c), we set $\dims = 10000$ and vary the number of samples $\numsamp$. (b) We plot the robust error of standard and adversarial training ($\epstrain = 4.5$). (c) We compute the error gap and the lower bound of Theorem \ref{thm:linlinf}. For more experimental details see Appendix~\ref{sec:logregapp}.}
  \vspace{-0.2in}
\label{fig:main_theorem}
\end{figure*}

\subsection{Main results}
\label{logreg_main_theorem}

We are now ready to characterize the
$\epstest$-robust error as a function of $\epstrain$, the separation
$\sigsep$, the dimension $\dims$ and sample size $\numsamp$ of the
data. In the theorem statement we use the following quantities
\begin{align*}
      \varphimin &= \frac{\mixvar}{r/2-\epstest}  \left(  \sqrt{\frac{\dims-1}{\numsamp}} - \left(1 + \sqrt{\frac{2 \log (2/\delta)}{\numsamp}}\right)\right)\\
      \varphimax &= \frac{\mixvar}{r/2-\epstest}  \left(  \sqrt{\frac{\dims-1}{\numsamp}} + \left(1 + \sqrt{\frac{2 \log (2/\delta)}{\numsamp}}\right)\right)
\end{align*}
that arise from concentration bounds for the singular values of the random data matrix. Further, let $\epscutoff := \frac{\sigsep}{2} - \frac{\varphimax}{\sqrt{2}}$ and denote by
 $\Phi$ the cumulative distribution function of a standard normal.
\begin{theorem}
  \label{thm:linlinf}
  Assume $d-1>n$. 
  For any $\epstest \geq 0$, the $\epstest$-robust error on test samples from $\prob_{\sigsep}$ with $2 \epstest < \sigsep$ and perturbation sets in Equation~\eqref{eq:linfmaxpert} and~\eqref{eq:l1maxpert}, the following holds:
  \begin{enumerate}
  \item
      The $\epstest$-robust error of the $\epstrain$-robust max-margin estimator reads
    \begin{equation}
      \roberr{\thetahat{\epstrain}} = \Phi \left( -\frac{\left( \frac{r}{2}-\epstrain \right) }{\randvarphi} \right)
    \end{equation}
    for a random quantity $\randvarphi>0$ depending on $\sigma, \sigsep,\epstest$, which is a strictly increasing function with respect to $\epstrain$.
     

  \item
    With probability at least $1-\delta$, we further have $\varphimin \leq \randvarphi\leq \varphimax$ and  the following lower bound on the robust error increase by adversarially training with size $\epstrain$
    \begin{equation}
      \roberr{\thetahat{\epstrain}} - \roberr{\thetahat{0}}
      \geq 
      \Phi \left(\frac{r/2}{\varphimin} \right) - \Phi \left(  \frac{r/2 -\min\{\epstrain, \epscutoff\}}{ \varphimin} \right).
    \end{equation}
  \end{enumerate}
\end{theorem}

The proof can be found in Appendix~\ref{sec:app_theorylinear} and
primarily relies on high-dimensional probability. Note that the
theorem holds for any $0\leq \epstest <\frac{\sigsep}{2}$ and hence
also directly applies to the standard error by setting $\epstest =
0$. In Figure~\ref{fig:main_theorem}, we empirically confirm the statements of Theorem \ref{thm:linlinf} by performing multiple experiments on synthetic datasets as described in Subsection \ref{logreg_linear_model} with different choices of $d/n$ and $\epstrain$. 
In the first statement, we prove that for small
sample-size ($n<d-1$) noiseless data,
almost surely, the robust error increases monotonically with
adversarial training budget $\epstrain >0$. 
In Figure~\ref{fig:main_lower_bound_eps}, we plot the robust error gap between standard and adversarial logistic regression in function of the adversarial training budget $\epstrain$ for $5$ runs. 

The second statement establishes a simplified lower bound on the
robust error increase for adversarial training (for a fixed
$\epstrain = \epstest$)  compared to standard training.
In Figures~\ref{fig:main_lower_bound_eps}  and \ref{fig:main_numobs_bound}, we show how the lower bound closely
predicts the robust error gap in our synthetic experiments.
Furthermore, by the dependence of $\varphimin$ on the overparameterization ratio $d/n$, the lower bound on the robust error gap is amplified for large $d/n$.
Indeed, Figure~\ref{fig:main_numobs_bound} shows how the error gap increases with $d/n$
both theoretically and experimentally. However, when $d/n$ increases above a certain threshold, the gap decreases again, as standard training fails to learn the signal and yields a high error (see Figure~\ref{fig:main_numobs}).

\begin{figure*}[!t]
\centering
\begin{subfigure}[b]{0.32\textwidth}
  \includegraphics[width=0.99\linewidth]{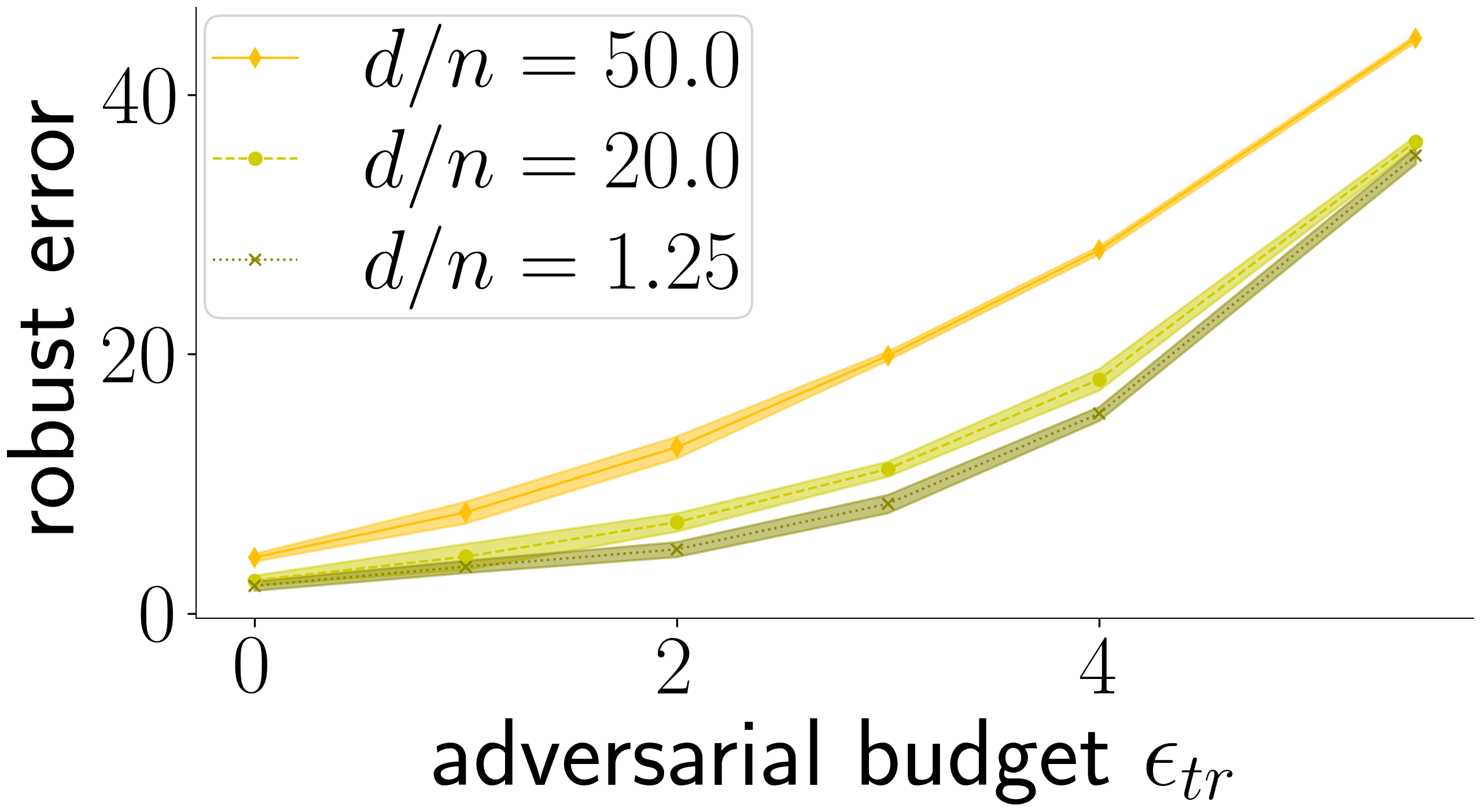}
  \caption{Robust error vs $\epstrain$}
  \label{fig:eps_logreg}
\end{subfigure}
\begin{subfigure}[b]{0.32\textwidth}
  \includegraphics[width=0.99\linewidth]{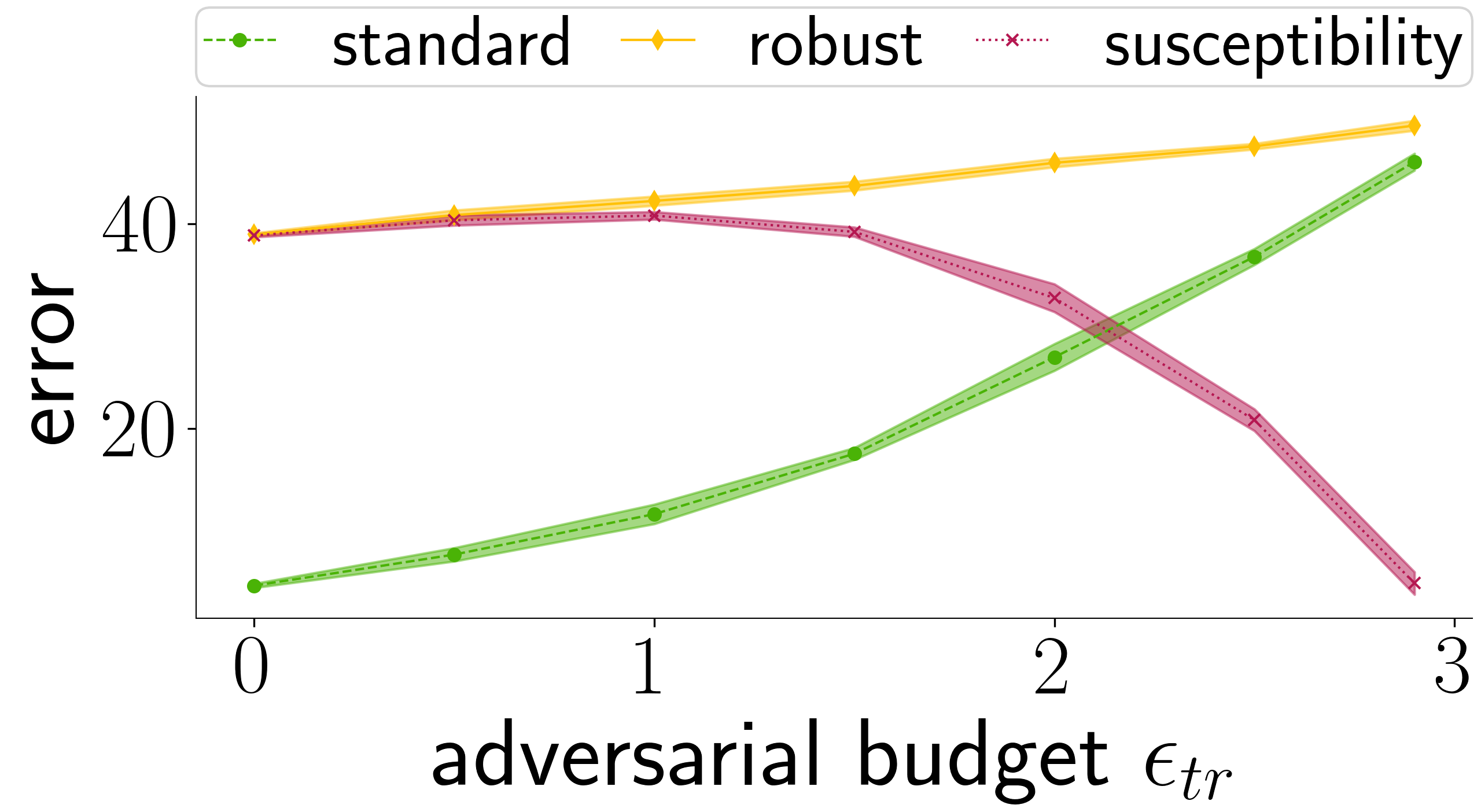}
  \caption{Robust error decomposition}
  \label{fig:main_robust}
\end{subfigure}
\begin{subfigure}[b]{0.32\textwidth}
  \includegraphics[width=0.99\linewidth]{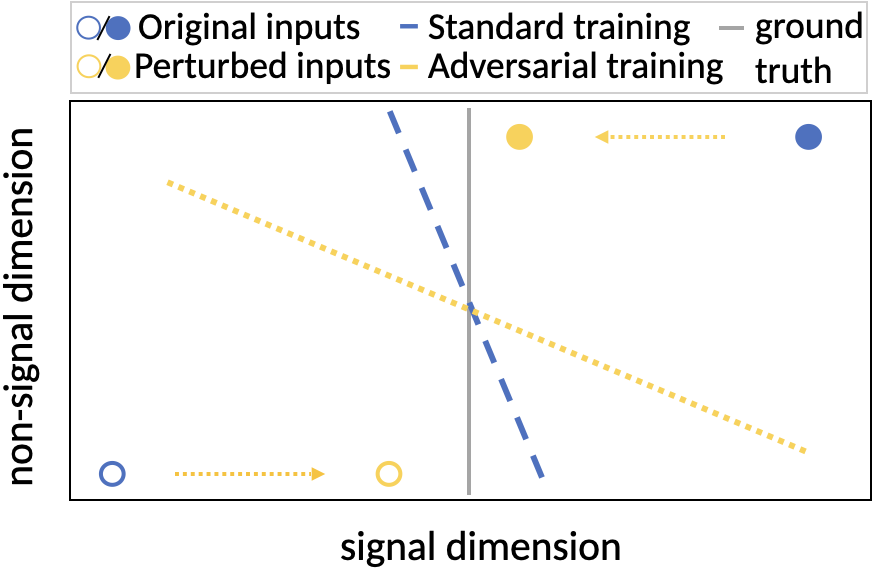}
  \caption{Intuition in 2D}
  \label{fig:2D_dataset_intuition}
\end{subfigure}
\caption{(a) We set $\dims=1000$ and $\sigsep = 12$ and plot the robust error with increasing adversarial training budget ($\epstrain$) and with increasing $\dims/\numsamp$.  (b) We plot the robust error decomposition in susceptibility and standard error for increasing adversarial budget $\epstrain$. 
  Full experimental details can be found in Section~\ref{sec:logregapp}. (c) 2D illustration providing intuition for the linear setting: Training on \nameofattacks (yellow) effectively corresponds to fiting the original datapoints (blue) after shifting them closer to the decision boundary. The robust max-$\ell_2$-margin (yellow dotted) is heavily tilted if the points are far apart in the non-signal dimension, while the standard max-$\ell_2$-margin solution (blue dashed) is much closer to the ground truth (gray solid). }
\label{fig:lineartradeoff}
\end{figure*}

\subsection{Proof idea: intuition and surprises}
\label{logreg_proof_sketch}

The reason that adversarial training hurts
robust generalization is based on an extreme robust vs. standard
error tradeoff. We provide intuition for the effect of
\nameofattacks and the small sample regime
on the solution of adversarial training by decomposing the
robust error $\roberr{\theta}$.
Notice that $\epstest$-robust error $\roberr{\theta}$ 
can be written as the probability of the union of two events: the
event that the classifier based on $\theta$ is wrong and the event
that the classifier is susceptible to attacks:
\begin{equation}
 \label{eq:decomposition}
\begin{aligned}
     \roberr{\theta} &=  \EE_{x, y\sim \prob}  \left[\Indi{y f_\theta (x) <0} \vee \max_{x' \in \pertset{x}{\epstest}} \Indi{f_\theta(x) f_\theta(x')<0} \right] \\
  &\leq \stderr{\theta} + \suscept{\theta}
\end{aligned}
\end{equation}
where $\suscept{\theta}$ is the expectation of the maximization term in Equation \eqref{eq:decomposition}.
$\suscept{\theta}$ represents the $\epstrain$-\emph{attack-susceptibility} of a classifier
induced by $\theta$ and $\stderr{\theta}$ its standard error.
Equation~\eqref{eq:decomposition} suggests that
the robust error can only be small if both the standard error and
susceptibility are small. In Figure~\ref{fig:main_robust}, we
plot the decomposition of the robust error in standard error and susceptibility for adversarial logistic regression with increasing $\epstrain$. We observe that increasing $\epstrain$
increases the standard error too drastically compared to the decrease
in susceptibility, leading to an effective drop in robust accuracy. For completeness, in Appendix \ref{app:susc}, we provide upper and lower bounds for the susceptibility score.  We
now explain why, in the small-sample size regime, adversarial training
with \nameofattacks ~\eqref{eq:linfmaxpert} may increase standard
error to the extent that it dominates the decrease in susceptibility.

A key observation is that the robust max-$\ell_2$-margin solution of a
dataset $\data= \{(x_i, y_i)\}_{i=1}^n$ 
maximizes the minimum margin that reads ${\min_{i\in [n]}
  y_i \theta^\top (x_i - y_i \epstrain |\thetaind{1}| e_1)}$, where
$\indof{\theta}{i}$ refers to the $i$-th entry of vector $\theta$. Therefore, it
simply corresponds to the max $\ell_2$-margin solution of the dataset
shifted towards the decision boundary ${\Dshift = \{(x_i - y_i \epstrain
  |\indof{\thetahat{\epstrain}}{1}| e_1, y_i)\}_{i=1}^n}$.
Using this fact, we obtain 
a closed-form expression of the (normalized) max-margin solution~\eqref{eq:maxmargin} as a function of
$\epstrain$ that reads
\begin{equation}
  \label{eq:maxmarginmaintext}
\thetahat{\epstrain} = \frac{1}{(r-2\epstrain)^2 + 4 \marginnonsig^2}
\left[\sigsep - 2\epstrain, 2 \marginnonsig \thetatilde \right],
\end{equation} 
where $\|\thetatilde\|_2 = 1$ and $\marginnonsig >0$ is a random quantity
associated with the max-$\ell_2$-margin solution of the
$\dims-1$ dimensional Gaussian inputs orthogonal
to the signal direction
(see Lemma~\ref{lem:maxmargin} in Section~\ref{sec:app_theorylinear}).

In high dimensions, with high probability any two
Gaussian random vectors are far apart -- in our
distributional setting, this corresponds to the vectors being far
apart in the non-signal directions. In
Figure~\ref{fig:2D_dataset_intuition}, we illustrate the phenomenon
using a simplified 2D cartoon, where the few samples
in the dataset are all far apart in the non-signal direction.
We see how shifting the dataset closer to the true decision boundary,
may result in a max-margin solution (yellow) that aligns much worse
with the ground truth (gray), compared to the estimator learned from
the original points (blue). Even though the new (robust max-margin)
classifier (yellow) is less susceptible to directed attacks in the
signal dimension, it also uses the signal dimension less.
Mathematically, this is directly
reflected in the expression of the max-margin solution in
Equation~\eqref{eq:maxmarginmaintext}: Even without the definition of
$\marginnonsig, \thetatilde$, we can directly see that the first
(signal) dimension is used less as $\epstrain$ increases.

\subsection{Generality of the results}

In this section we discuss how the theorem might generalize to
other perturbation sets, models and training procedures.
\paragraph{Signal direction is known}
The type of additive perturbations used in Theorem~\ref{thm:linlinf},
defined in Equation~\eqref{eq:linfmaxpert}, is explicitly constrained
to the direction of the true signal. This choice is reminiscent of
corruptions where every possible perturbation in the set is directly
targeted at the object to be recognized, such as motion blur of moving
objects.  Such corruptions are also studied in the context of domain
generalization and adaptation \cite{Schneider20}.

\nameofattackscapital in general, however, may also consist of
perturbation sets that are only strongly biased towards the true
signal direction, such as mask attacks.  They may find the true signal
direction only when the inner maximization is
exact. The following corollary extends Theorem~\ref{thm:linlinf} to
small $\ell_1$-perturbations
\begin{equation}
  \label{eq:l1maxpert}
  \pertset{x}{\eps} = \{x'=x+\delta \mid \|\delta\|_1 \leq \eps\},
\end{equation}
for $0<\eps<\frac{\sigsep}{2}$ that reflect such attacks. We state the corollary here and give the proof in Appendix \ref{sec:app_theorylinear}.
\begin{corollary}
\label{cor:l1extension}
  Theorem~\ref{thm:linlinf} also holds for ~\eqref{eq:maxmargin} with perturbation sets defined in \eqref{eq:l1maxpert}.
\end{corollary}
The proof uses the fact that the inner maximization
effectively results in a sparse perturbation equivalent to the attack
resulting from the perturbation set~\eqref{eq:linfmaxpert}.

\paragraph{Other models}
Motivated by the implicit bias results of (stochastic)
gradient descent on the logistic loss, Theorem~\ref{thm:linlinf} is proven for the max-$\ell_2$-margin
solution. We would like to conjecture
that for the data distribution in Section \ref{sec:theoryresults},
adversarial training can hurt robust generalization also for other models with zero
training error (\emph{interpolators} in short).

For example, Adaboost is a widely used algorithm that converges to the max-$\ell_1$-margin classifier \cite{telgarsky13}. One might argue that for a sparse ground truth, the max-$\ell_1$-margin classifier should (at least in the noiseless case) have the right inductive bias to alleviate large bias in high dimensions. Hence, in many cases the (sparse) max-$\ell_1$-margin solution might align with the ground
truth for a given dataset. However, we conjecture that even in this
case, the \emph{robust} max-$\ell_1$-margin solution (of the dataset
shifted towards the decision boundary) would be misled to choose a
wrong sparse solution. This can be seen with the help of the cartoon
illustration in Figure \ref{fig:2D_dataset_intuition}.

\begin{figure*}[!t]
\centering
\begin{subfigure}[b]{0.33\textwidth}
  \includegraphics[width=0.99\linewidth]{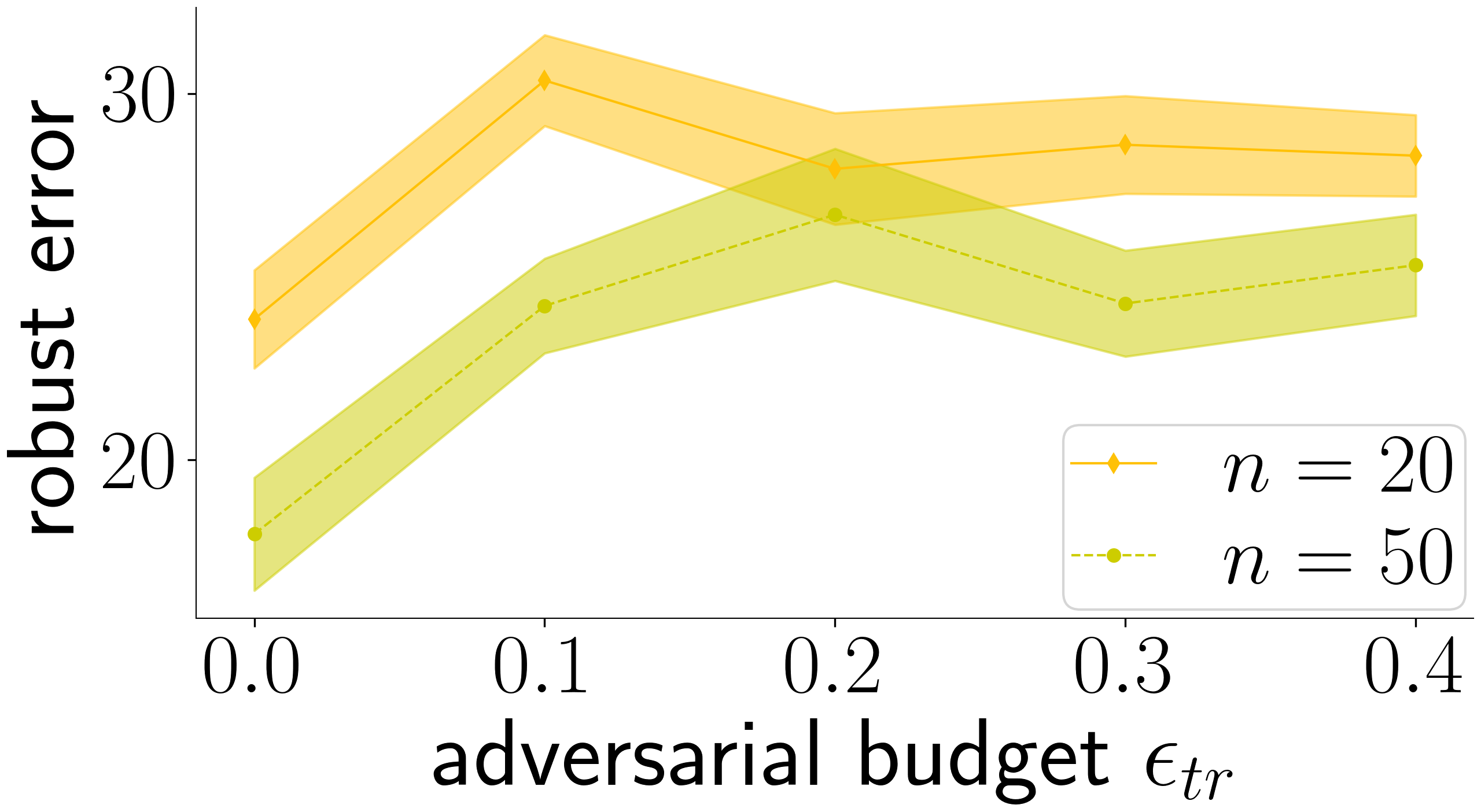}
  \caption{Robust error vs $\epstrain$}
  \label{fig:waterbirds_light_d_n}
\end{subfigure}
\begin{subfigure}[b]{0.32\textwidth}
  \includegraphics[width=0.99\linewidth]{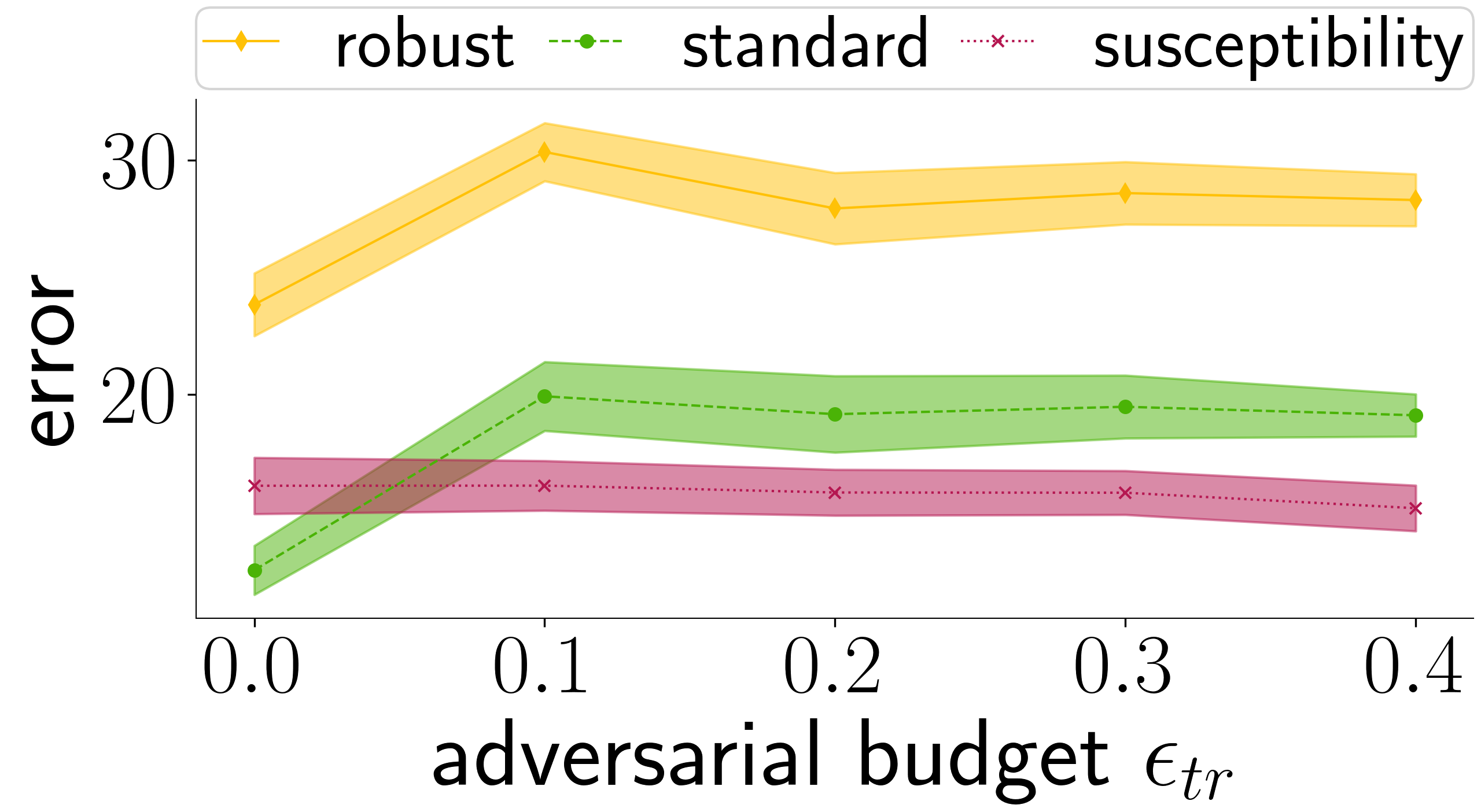}
  \caption{Robust error decomposition}
  \label{fig:light_trade_off}
\end{subfigure}
\begin{subfigure}[b]{0.33\textwidth}
  \includegraphics[width=0.99\linewidth]{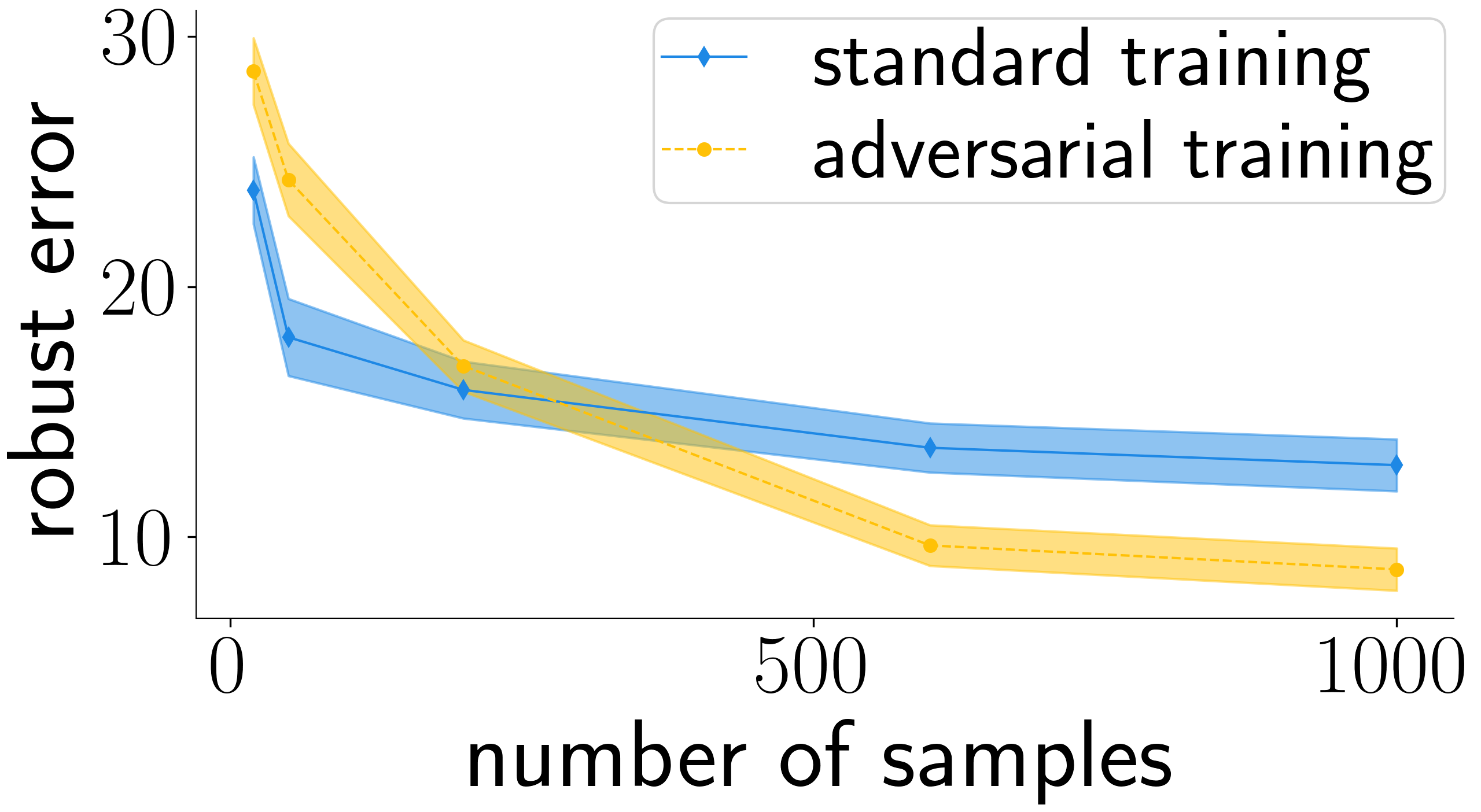}
  \caption{Number of samples}
  \label{fig:waterbirds_light_numobs}
\end{subfigure}
  \caption{Experiments on the Waterbirds dataset considering the adversarial illumination attack with $\epstest = 0.3$. We plot the mean and standard deviation of the mean of several independent experiments. (a) The robust error increases with larger $\epstrain$ in the low sample size regime. (b) We set $\numsamp=20$ and plot the robust error decomposition as in Equation $\eqref{eq:decomposition}$ with increasing $\epstrain$. While the susceptibility decreases slightly, the increase in standard error is much more severe, resulting in an increase in robust error. (c) Adversarial training hurts robust generalization in the low sample size regime $(\numsamp < 200)$, but helps when enough samples are available. For more experimental details see Section \ref{sec:waterbirds}.}
\label{fig:waterbirds_light}
\end{figure*}

\section{Real-world experiments}
\label{sec:realworldexpapp}

In this section, we demonstrate that adversarial training may
hurt robust accuracy in a variety of image attack scenarios
on the Waterbirds and CIFAR10 dataset.
The corresponding experimental details and more experimental results (including
on an additional hand gestures dataset) can be found in Appendices
 \ref{sec:waterbirds}, \ref{sec:app_cifar10} and \ref{sec:handgestures}.


\subsection{Datasets}

We now describe the datasets and models that we use for the
experiments. In all our experiments on CIFAR10, we vary the sample
size by subsampling the dataset and use a ResNet18 \cite{He16} as
model. We always train on the same (randomly subsampled) dataset,
meaning that the variances arise from the random seed of the model and
the randomness in the training algorithm. In Appendix
\ref{sec:app_cifar10}, we complement the results of this section by
reporting the results of similar experiments with different
architectures.

As a second dataset, we build a new version of the Waterbirds
dataset, consisting of images of water- and
landbirds of size $256 \times 256$ and labels that distinguish the
two types of birds. We construct the dataset as follows: First, we
sample equally many water- and landbirds from the CUB-200 dataset
\cite{Welinder10}. Then, we segment the birds and paste them onto a
background that is randomly sampled (without replacement) from the Places-256 dataset \cite{zhou17}.
For the implementation of the dataset we used the code provided by \citet{Sagawa20}. Also, following the choice of \citet{Sagawa20}, we use as model a ResNet50 that was pretrained on ImageNet and which achieves near perfect standard accuracy.

\subsection{Evaluation of \nameofattacks}

We consider three types of \nameofattacks on our real world datasets:
square masks, motion blur and adversarial illumination. The mask
attack is a model used to simulate sticker-attacks and general
occlusions of objects in images \cite{Eykholt18, Wu20}. On the other
hand, motion blur may arise naturally for example when photographing
fast moving objects with a slow shutter speed. Further, adversarial
illumination may result from adversarial lighting conditions or smart
image corruptions. Next, we describe the attacks in more detail.

\paragraph{Mask attacks}
On CIFAR10, we consider the square black mask attack: the adversary can set a mask
of size $\epstest \times \epstest$ to zero in the image. To ensure that the mask does not cover the whole signal in the image, we
restrict the size of the masks to be at most $2 \times 2$. Hence, the search space of the attack consists of all possible locations of the masks in the targeted image. For exact robust error evaluation, we perform a full grid search over all possible locations during test time. See Figure \ref{fig:CIFAR10_boat} for an example of a mask attack on CIFAR10.

\paragraph{Motion blur}
On the Waterbirds dataset we consider two \nameofattacks: motion blur and adversarial illumination. For the motion blur attack,
the bird may move at different speeds without changing the background. 
The aim is to be robust against all motion blur severity levels up to $\motionblurkernel_{max} = 15$. 
To simulate motion blur, we first segment the birds and then use a filter with a kernel of size $\motionblurkernel$ to apply motion blur on the bird only. Lastly, we paste the blurred bird back onto the background image. We can change the severity level of the motion blur by increasing the kernel size of the filter.
See Appendix \ref{sec:waterbirds} for an ablation study and concrete expressions of the motion blur kernel. At test time, we perform a full grid search over all kernel sizes to exactly evaluate the robust error. We refer to Figure \ref{fig:WB_motion_blur} and Section \ref{sec:waterbirds} for examples of our motion blur attack.

\paragraph{Adversarial illumination} As a second attack on the Waterbirds dataset, we consider adversarial illumination. The adversary can darken or brighten the bird without corrupting the background of the image. The attack aims to model images where the object at interest is hidden in shadows or placed against bright light. 
To compute the adversarial illumination attack, we segment the bird, then darken or brighten the it, by adding a constant $a \in [-\epstest, \epstest]$, before pasting the bird back onto the background image. We find the most adversarial lighting level, i.e. the value of $a$, by equidistantly partitioning the interval $[-\epstest, \epstest]$ in $K$ steps and performing a full list-search over all steps.
See Figure \ref{fig:WB_light_dark} and Section \ref{sec:waterbirds} for examples of the adversarial illumination attack.

\subsection{Adversarial training procedure}

For all datasets, we run SGD until convergence on the \emph{robust} cross-entropy
loss~\eqref{eq:emploss}. In each iteration, we search for an adversarial example
and update the weights using a gradient with respect to the resulting
perturbed example \cite{goodfellow15, madry18}.
For every experiment, we choose the learning
rate and weight decay parameters that minimize the robust error on a
hold-out dataset. We now describe the implementation of the
adversarial search for the three types of
\nameofattacks. 

\paragraph{Mask attacks}
Unless specified otherwise, we use an approximate attack similar to
\citet{Wu20} during training time:
First, we identify promising mask locations by analyzing the gradient, $\nabla_x \loss(f_\theta(x), y)$, of the cross-entropy loss with respect to the input. Masks that cover part of the image where the gradient is large, are more likely to increase the loss. Hence, we compute the $K$ mask locations $(i, j)$, where $\|\nabla_x \loss(f_\theta(x), y)_{[i:i+2, j:j+2]} \|_1$ is the largest and take using a full list-search the mask that incurs the highest loss.
Our intuition from the theory predicts that higher $K$,
and hence a more exact ``defense'', only increases the robust error of
adversarial training, since the mask could then more efficiently cover
important information about the class. We indeed confirm this effect
and provide more details in Section~\ref{sec:app_cifar10}.

\paragraph{Motion blur}
Intuitively the worst attack should be the most severe blur, rendering
a search over a range of severity superfluous.  However, similar to
rotations, this is not necessarily true in practice since the training loss on
neural networks is generally nonconvex. Hence, during training time,
we perform a search over kernels with sizes $2i$ for $i = 1,\dots,
\motionblurkernel_{max}/2$. Note that, at test time, we do an exact search
over all kernels of sizes in $[1, 2, \dots, \motionblurkernel_{max}]$.

\paragraph{Adversarial illumination}
Similar to the motion blur attack, intuitively the worst perturbation
should be the most severe lighting changes; either darkening or
illuminating the object maximally. However, again this is not
necessarily the case, since finding the worst attack is a nonconvex
problem. Therefore, during training and testing we partition the
interval $[-\epstrain, \epstrain]$ in $33$ and $65$ steps
respectively, and perform a full grid-search to find the worst
perturbation.

\subsection{Adversarial training can hurt robust generalization}

\begin{figure*}[!t]
\centering
\begin{subfigure}[b]{0.4\textwidth}
  \includegraphics[width=0.99\linewidth]{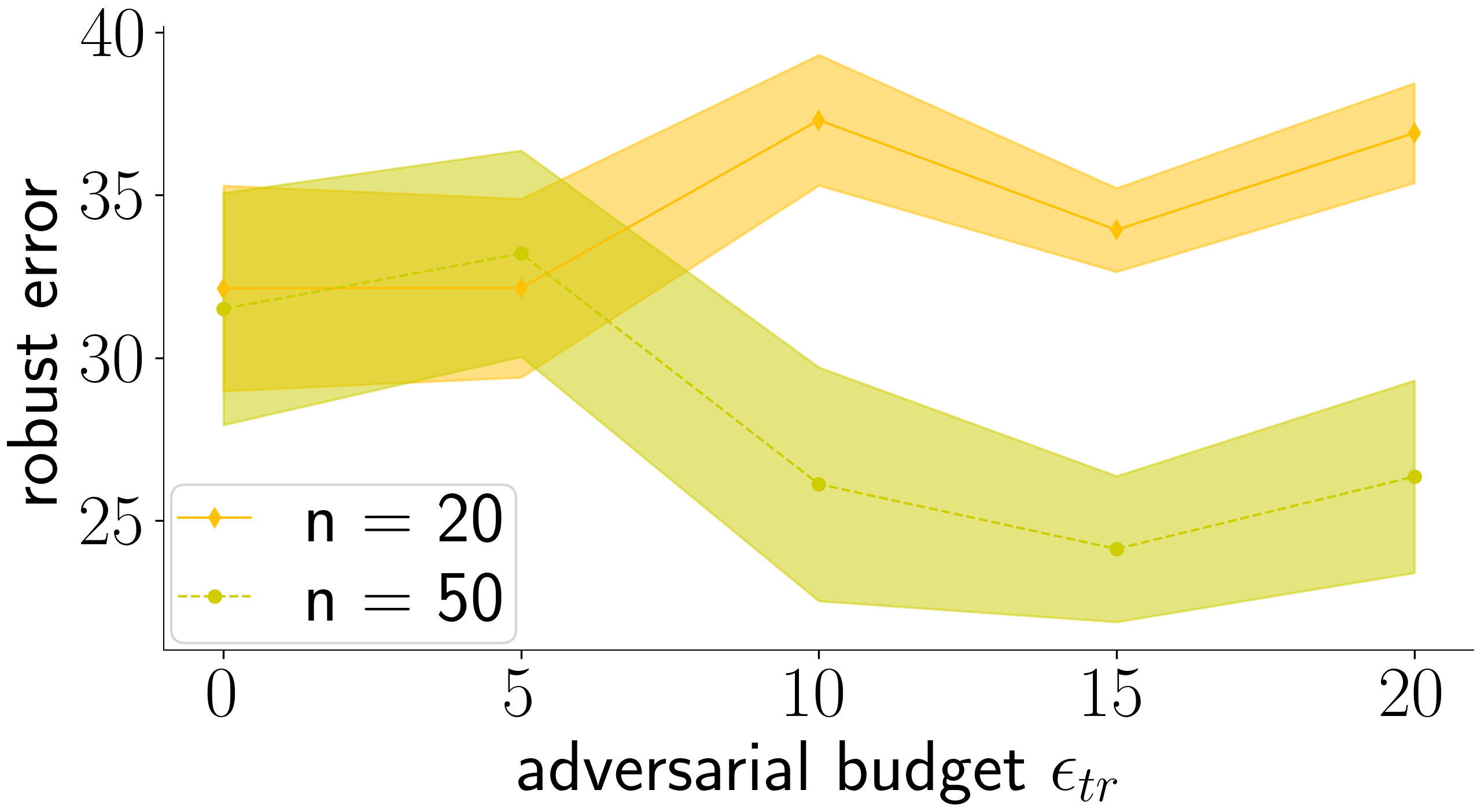}
  \caption{Robust error with increasing $\epstrain$}
  \label{fig:motion_lines}
\end{subfigure}
\begin{subfigure}[b]{0.4\textwidth}
  \includegraphics[width=0.99\linewidth]{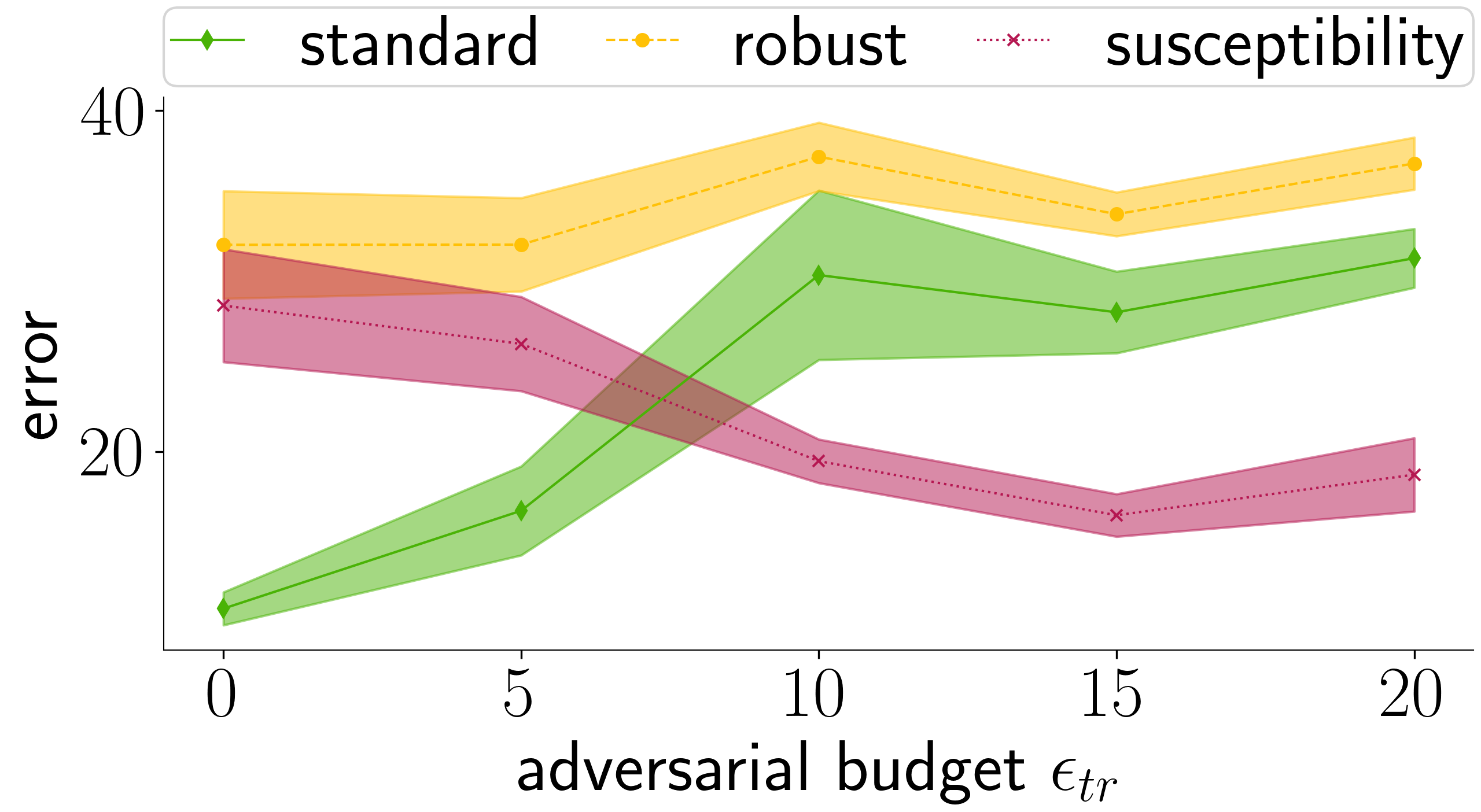}
  \caption{Robust error decomposition}
  \label{fig:motion_blur_trade_off}
\end{subfigure}
  \caption{ (a) We plot the robust error with increasing adversarial training budget $\epstrain$ of $5$ experiments on the subsampled Waterbirds datasets of sample sizes $20$ and $30$. Even though adversarial training hurts robust generalization for low sample size ($\numsamp = 20$), it helps for $\numsamp = 50$.  (b) We plot the decomposition of the robust error in standard error and susceptibility with increasing adversarial budget $\epstrain$. We plot the mean and standard deviation of the mean of $5$ experiments on a subsampled Waterbirds dataset of size $\numsamp = 20$. The increase in standard error is more severe than the drop in susceptibility, leading to a slight increase in robust error. For more experimental details see Section \ref{sec:waterbirds}.}
\label{fig:motion_blur_real_world}
\end{figure*}

Further, we perform the following experiments on the Waterbirds dataset using the motion blur and adversarial illumination attack. We vary the adversarial training  budget $\epstrain$, while keeping the number of samples fixed, and compute the resulting robust error.
We see in Figure \ref{fig:waterbirds_light_d_n} and \ref{fig:motion_lines} that, indeed, adversarial training can hurt robust generalization with increasing perturbation budget $\epstrain$.

Furthermore, to gain intuition as described in Section~\ref{logreg_proof_sketch} and, we also plot the robust error decomposition (Equation~\ref{eq:decomposition}) consisting of the standard error and susceptibility in Figure \ref{fig:light_trade_off} and \ref{fig:motion_blur_trade_off}. Recall that we measure susceptibility as the fraction of data points in the test set for
which the classifier predicts a different class under an adversarial attack.
As in our linear example, we observe an increase in robust error despite a slight drop
in susceptibility, because of the more severe increase in standard error. 
Similar experiments for the hand gesture dataset can be
found in~\ref{sec:handgestures}.

As predicted by our theorem, the phenomenon where adversarial training hurts robust generalization is most pronounced in the small sample size regime. Indeed, the experiments depicted in Figures \ref{fig:waterbirds_light_d_n} and \ref{fig:motion_lines} are conducted on small sample size datasets of $\numsamp = 20$ or $50$.
In Figure \ref{fig:teaserplot} and \ref{fig:waterbirds_light_numobs}, we
observe that the as sample size increases,  adversarial training does improve robust generalization compared to standard training, even for \nameofattacks. Moreover, on the experiments of CIFAR10 using the mask perturbation, which can be found in Figure \ref{fig:teaserplot} and Appendix \ref{sec:app_cifar10}, we observe the same behaviour: Adversarial training hurts robust generalization in the low sample size regime, but helps when enough samples are available. 


\subsection{Discussion}

In this section, we discuss how different algorithmic choices, motivated
by related work, affect when and how adversarial training hurts robust generalization. 

\paragraph{Strength of attack and catastrophic overfitting}
In many cases, the worst case perturbation during adversarial training is found using an approximate algorithm such as projected gradient descent. It is common belief  that using the strongest attack (in the mask-perturbation case, full grid search) during training should also result in better robust generalization. 
In particular, the literature on catastrophic overfitting shows that weaker attacks during training lead to bad performance on stronger attacks during testing  \cite{Wong20Fast, andriushchenko20, li21}.
Our result suggests the opposite is true in the low-sample size regime for
\nameofattacks : the weaker the attack, the better
adversarial training performs.

\paragraph{Robust overfitting}
Recent work observes empirically \cite{rice20} and theoretically
\cite{sanyal20, donhauser21}, that perfectly minimizing the
adversarial loss during training might in fact be suboptimal for
robust generalization; that is, classical regularization techniques
might lead to higher robust accuracy. The phenomenon is often referred
to as robust overfitting. May the phenomenon be mitigated using
standard regularization techniques?  In Appendix \ref{sec:waterbirds} we shed light on
this question and show that adversarial training hurts robust generalization even with standard regularization methods such as early stopping are used.





\section{Related work}
\label{sec:relatedwork}

We now discuss how our results relate to phenomena that have been observed or proven in the literature before.

\paragraph{Robust and non-robust useful features}
In the words of \citet{ilyas19, springer21}, for
directed attacks, all robust features become less useful, but adversarial
training uses robust features more.  In the small sample-size regime
$n<d-1$ in particular, robust learning assigns so much weight
on the robust (possibly non-useful) features, that the signal in the non-robust
features is drowned. This leads to an unavoidable and large increase
in standard error that dominates the decrease in susceptibility and
hence ultimately leads to an increase of the robust error.

\paragraph{Small sample size and robustness}
A direct consequence of Theorem~\ref{thm:linlinf} is that in order to
achieve the same robust error as standard training, adversarial
training requires more samples. This statement might remind the reader
of sample complexity results for robust generalization in
\citet{schmidt18, Yin19, Khim18}. While those results compare sample
complexity bounds for standard vs. robust error, our theorem
statement compares two algorithms, standard vs. adversarial training,
with respect to the robust error.

\paragraph{Trade-off between standard and robust error} 

Many papers observed that even though adversarial training decreases robust error compared to standard training, it may lead
to an increase in standard test error \cite{madry18, zhang19}.  
For example, \citet{tsipras19, zhang19, javanmard20, dobriban20, chen20} study settings where the Bayes optimal robust classifier is not equal to the Bayes optimal (standard)
classifier (i.e. the perturbations are inconsistent or the dataset is non-separable).
\cite{raghunathan20} study consistent perturbations, as in our paper,
and prove that for small sample size, fitting adversarial
examples can increase standard error even in the absence of
noise. In contrast to aforementioned works, which do not refute that
adversarial training decreases robust error, we prove that for
\nameofattacks perturbations, in the small sample regime adversarial training may also increase \emph{robust error}.

\paragraph{Mitigation of the trade-off} 
A long line of work has proposed procedures to 
mitigate the trade-off phenomenon.  For example \citet{alayrac19,
  Carmon19, zhai20, raghunathan20} study robust self training, which
leverages a large set of unlabelled data, while \citet{lee20, lamb19,
  xu20} use data augmentation by interpolation. \citet{Ding20,
  balaji19, Cheng20} on the other hand propose to use adaptive
perturbation budgets $\epstrain$ that vary across inputs. 
Our intuition from the theoretical analysis suggests that the standard
mitigation procedures for imperceptible perturbations may not work for
perceptible \nameofattacks, because all relevant features are non-robust.
We leave a thorough empirical study
as interesting future work.

\section{Future work}

This paper aims to caution the practitioner against blindly following
current widespread practices to increase the robust
performance of machine learning models.
Specifically, adversarial training is currently recognized to be one
of the most effective defense mechanisms for $\ell_p$-perturbations,
significantly outperforming robust performance of standard training.  However, we prove that
this common wisdom is not applicable for directed attacks -- that are perceptible (albeit consistent) but efficiently focus their
attack budget to target ground truth class information -- in the low-sample size regime.
In particular, in such settings adversarial training can in fact yield worse accuracy than standard training.





In terms of follow-up work on directed attacks in the low-sample
regime, there are some concrete questions that would be interesting to
explore.  For example, as discussed in Section~\ref{sec:relatedwork},
it would be useful to test whether some methods to mitigate the
standard accuracy vs. robustness trade-off would also relieve the
perils of adversarial training for directed attacks. Further, we
hypothesize, independent of the attack during test time, it is
important in the small sample-size regime to choose perturbation sets
during training that align with
the ground truth signal (such as rotations for data with inherent
rotation). If this hypothesis were to be confirmed, it would break
with yet another general rule that the best defense perturbation type
should always match the attack during evaluation.  The insights from
this study might also be helpful in the context of searching for
good defense perturbations.


\bibliographystyle{icml2022}
\bibliography{ICML}

\begin{thebibliography}{60}
\providecommand{\natexlab}[1]{#1}
\providecommand{\url}[1]{\texttt{#1}}
\expandafter\ifx\csname urlstyle\endcsname\relax
  \providecommand{\doi}[1]{doi: #1}\else
  \providecommand{\doi}{doi: \begingroup \urlstyle{rm}\Url}\fi

\bibitem[Alayrac et~al.(2019)Alayrac, Uesato, Huang, Fawzi, Stanforth, and
  Kohli]{alayrac19}
Alayrac, J.-B., Uesato, J., Huang, P.-S., Fawzi, A., Stanforth, R., and Kohli,
  P.
\newblock Are labels required for improving adversarial robustness?
\newblock \emph{Advances in Neural Information Processing Systems}, pp.\
  12214--12223, 2019.

\bibitem[Andriushchenko \& Flammarion(2020)Andriushchenko and
  Flammarion]{andriushchenko20}
Andriushchenko, M. and Flammarion, N.
\newblock Understanding and improving fast adversarial training.
\newblock \emph{Advances in Neural Information Processing Systems}, 2020.

\bibitem[Bai et~al.(2021)Bai, Luo, Zhao, Wen, and Wang]{Bai21}
Bai, T., Luo, J., Zhao, J., Wen, B., and Wang, Q.
\newblock Recent advances in adversarial training for adversarial robustness.
\newblock In \emph{International Joint Conference on Artificial Intelligence},
  pp.\  4312--4321, Aug 2021.

\bibitem[Balaji et~al.(2019)Balaji, Goldstein, and Hoffman]{balaji19}
Balaji, Y., Goldstein, T., and Hoffman, J.
\newblock Instance adaptive adversarial training: Improved accuracy tradeoffs
  in neural nets.
\newblock \emph{arXiv preprint arXiv:1910.08051}, 2019.

\bibitem[Bradski(2000)]{opencv_library}
Bradski, G.
\newblock {The OpenCV Library}.
\newblock \emph{Dr. {D}obb's Journal of Software Tools}, 2000.

\bibitem[Carmon et~al.(2019)Carmon, Raghunathan, Schmidt, Liang, and
  Duchi]{Carmon19}
Carmon, Y., Raghunathan, A., Schmidt, L., Liang, P., and Duchi, J.~C.
\newblock Unlabeled data improves adversarial robustness.
\newblock In \emph{International Conference on Neural Information Processing
  Systems}, pp.\  11192--11203, Dec 2019.

\bibitem[Chen et~al.(2020)Chen, Min, Zhang, and Karbasi]{chen20}
Chen, L., Min, Y., Zhang, M., and Karbasi, A.
\newblock More data can expand the generalization gap between adversarially
  robust and standard models.
\newblock In \emph{International Conference on Machine Learning}, pp.\
  1670--1680, Jun 2020.

\bibitem[Cheng et~al.(2020)Cheng, Lei, Chen, Dhillon, and Hsieh]{Cheng20}
Cheng, M., Lei, Q., Chen, P.-Y., Dhillon, I., and Hsieh, C.-J.
\newblock {CAT}: Customized adversarial training for improved robustness.
\newblock \emph{arXiv preprint arXiv:2002.06789}, 2020.

\bibitem[Chizat \& Bach(2020)Chizat and Bach]{Chizat20}
Chizat, L. and Bach, F.
\newblock Implicit bias of gradient descent for wide two-layer neural networks
  trained with the logistic loss.
\newblock In \emph{International Conference on Learning Theory}, pp.\
  1305--1338, Jul 2020.

\bibitem[Ding et~al.(2020)Ding, Sharma, Lui, and Huang]{Ding20}
Ding, G.~W., Sharma, Y., Lui, K. Y.~C., and Huang, R.
\newblock {MMA} training: Direct input space margin maximization through
  adversarial training.
\newblock In \emph{International Conference on Learning Representations}, Apr
  2020.

\bibitem[Dobriban et~al.(2020)Dobriban, Hassani, Hong, and Robey]{dobriban20}
Dobriban, E., Hassani, H., Hong, D., and Robey, A.
\newblock Provable tradeoffs in adversarially robust classification.
\newblock \emph{arXiv preprint arXiv:2006.05161}, 2020.

\bibitem[Donhauser et~al.(2021)Donhauser, Tifrea, Aerni, Heckel, and
  Yang]{donhauser21}
Donhauser, K., Tifrea, A., Aerni, M., Heckel, R., and Yang, F.
\newblock Interpolation can hurt robust generalization even when there is no
  noise.
\newblock In \emph{Advances in Neural Information Processing Systems}, Dec
  2021.

\bibitem[Engstrom et~al.(2019)Engstrom, Tran, Tsipras, Schmidt, and
  Madry]{Logan19}
Engstrom, L., Tran, B., Tsipras, D., Schmidt, L., and Madry, A.
\newblock Exploring the landscape of spatial robustness.
\newblock In \emph{International Conference on Machine Learning}, pp.\
  1802--1811, Jun 2019.

\bibitem[Eykholt et~al.(2018)Eykholt, Evtimov, Fernandes, Li, Rahmati, Xiao,
  Prakash, Kohno, and Song]{Eykholt18}
Eykholt, K., Evtimov, I., Fernandes, E., Li, B., Rahmati, A., Xiao, C.,
  Prakash, A., Kohno, T., and Song, D.
\newblock Robust physical-world attacks on deep learning visual classification.
\newblock In \emph{IEEE Conference on Computer Vision and Pattern Recognition},
  pp.\  1625--1634, Jun 2018.

\bibitem[Ghiasi et~al.(2019)Ghiasi, Shafahi, and Goldstein]{ghiasi19}
Ghiasi, A., Shafahi, A., and Goldstein, T.
\newblock Breaking certified defenses: semantic adversarial examples with
  spoofed robustness certificates.
\newblock In \emph{International Conference on Learning Representations}, Apr
  2019.

\bibitem[Gilmer et~al.(2018)Gilmer, Adams, Goodfellow, Andersen, and
  Dahl]{gilmer18b}
Gilmer, J., Adams, R.~P., Goodfellow, I., Andersen, D., and Dahl, G.~E.
\newblock Motivating the rules of the game for adversarial example research.
\newblock \emph{arXiv preprint arXiv:1807.06732}, 2018.

\bibitem[Goodfellow et~al.(2015)Goodfellow, Shlens, and Szegedy]{goodfellow15}
Goodfellow, I., Shlens, J., and Szegedy, C.
\newblock Explaining and harnessing adversarial examples.
\newblock In \emph{International Conference on Learning Representations}, pp.\
  1--10, Jan 2015.

\bibitem[He et~al.(2016)He, Zhang, Ren, and Sun]{He16}
He, K., Zhang, X., Ren, S., and Sun, J.
\newblock Deep residual learning for image recognition.
\newblock In \emph{IEEE Conference on Computer Vision and Pattern Recognition},
  pp.\  770--778, Jun 2016.

\bibitem[Ilyas et~al.(2019)Ilyas, Santurkar, Tsipras, Engstrom, Tran, and
  Madry]{ilyas19}
Ilyas, A., Santurkar, S., Tsipras, D., Engstrom, L., Tran, B., and Madry, A.
\newblock Adversarial examples are not bugs, they are features.
\newblock In \emph{Advances in Neural Information Processing Systems}, pp.\
  125--136, Dec 2019.

\bibitem[Javanmard et~al.(2020)Javanmard, Soltanolkotabi, and
  Hassani]{javanmard20}
Javanmard, A., Soltanolkotabi, M., and Hassani, H.
\newblock Precise tradeoffs in adversarial training for linear regression.
\newblock In \emph{Conference on Learning Theory}, pp.\  2034--2078, Apr 2020.

\bibitem[Ji \& Telgarsky(2019)Ji and Telgarsky]{Ji19}
Ji, Z. and Telgarsky, M.
\newblock The implicit bias of gradient descent on nonseparable data.
\newblock In \emph{Conference on Learning Theory}, pp.\  1772--1798, Jun 2019.

\bibitem[Khim \& Loh(2018)Khim and Loh]{Khim18}
Khim, J. and Loh, P.-L.
\newblock Adversarial risk bounds via function transformation.
\newblock \emph{arXiv preprint arXiv:1810.09519}, 2018.

\bibitem[Laidlaw et~al.(2021)Laidlaw, Singla, and Feizi]{laidlaw21}
Laidlaw, C., Singla, S., and Feizi, S.
\newblock Perceptual adversarial robustness: Defense against unseen threat
  models.
\newblock In \emph{International Conference on Learning Representation}, Jun
  2021.

\bibitem[Lamb et~al.(2019)Lamb, Verma, Kannala, and Bengio]{lamb19}
Lamb, A., Verma, V., Kannala, J., and Bengio, Y.
\newblock Interpolated adversarial training: Achieving robust neural networks
  without sacrificing too much accuracy.
\newblock In \emph{ACM Workshop on Artificial Intelligence and Security}, pp.\
  95--103, 2019.

\bibitem[Lee et~al.(2020)Lee, Lee, and Yoon]{lee20}
Lee, S., Lee, H., and Yoon, S.
\newblock Adversarial {V}ertex {M}ixup: Toward better adversarially robust
  generalization.
\newblock In \emph{IEEE/CVF Conference on Computer Vision and Pattern
  Recognition}, pp.\  272--281, Jun 2020.

\bibitem[Li et~al.(2021)Li, Wang, Jana, and Carin]{li21}
Li, B., Wang, S., Jana, S., and Carin, L.
\newblock Towards understanding fast adversarial training.
\newblock \emph{arXiv preprint arXiv:2006.03089}, 2021.

\bibitem[Lin et~al.(2020)Lin, Lau, Levine, Chellappa, and Feizi]{Lin20}
Lin, W.-A., Lau, C.~P., Levine, A., Chellappa, R., and Feizi, S.
\newblock Dual manifold adversarial robustness: Defense against {L}p and
  non-{L}p adversarial attacks.
\newblock In \emph{Advances in Neural Information Processing Systems}, pp.\
  3487--3498, Dec 2020.

\bibitem[Liu et~al.(2020)Liu, Salzmann, Lin, Tomioka, and S\"{u}sstrunk]{liu20}
Liu, C., Salzmann, M., Lin, T., Tomioka, R., and S\"{u}sstrunk, S.
\newblock On the loss landscape of adversarial training: Identifying challenges
  and how to overcome them.
\newblock In \emph{Advances in Neural Information Processing Systems}, pp.\
  21476--21487, 2020.

\bibitem[Luo et~al.(2018)Luo, Liu, Wei, and Xu]{Luo18}
Luo, B., Liu, Y., Wei, L., and Xu, Q.
\newblock Towards imperceptible and robust adversarial example attacks against
  neural networks.
\newblock In \emph{AAAI Conference on Artificial Intelligence and Innovative
  Applications}, Feb 2018.

\bibitem[Madry et~al.(2018)Madry, Makelov, Schmidt, Tsipras, and
  Vladu]{madry18}
Madry, A., Makelov, A., Schmidt, L., Tsipras, D., and Vladu, A.
\newblock Towards deep learning models resistant to adversarial attacks.
\newblock In \emph{International Conference on Learning Representations}, 2018.

\bibitem[Mantecón et~al.(2019)Mantecón, del Blanco, Jaureguizar, and
  García]{Mantecon19}
Mantecón, T., del Blanco, C.~R., Jaureguizar, F., and García, N.
\newblock A real-time gesture recognition system using near-infrared imagery.
\newblock \emph{PLOS ONE}, pp.\  1--17, Oct 2019.

\bibitem[Moosavi-Dezfooli et~al.(2016)Moosavi-Dezfooli, Fawzi, and
  Frossard]{moosavi16}
Moosavi-Dezfooli, S.-M., Fawzi, A., and Frossard, P.
\newblock {D}eepfool: a simple and accurate method to fool deep neural
  networks.
\newblock In \emph{IEEE conference on computer vision and pattern recognition},
  pp.\  2574--2582, Jun 2016.

\bibitem[Mujahid et~al.(2021)Mujahid, Awan, Yasin, Mohammed, Damaševičius,
  Maskeliūnas, and Abdulkareem]{Mujahid21}
Mujahid, A., Awan, M.~J., Yasin, A., Mohammed, M.~A., Damaševičius, R.,
  Maskeliūnas, R., and Abdulkareem, K.~H.
\newblock Real-time hand gesture recognition based on deep learning {YOLO}v3
  model.
\newblock \emph{Applied Sciences}, 2021.

\bibitem[Nacson et~al.(2019)Nacson, Srebro, and Soudry]{nacson19}
Nacson, M.~S., Srebro, N., and Soudry, D.
\newblock Stochastic gradient descent on separable data: Exact convergence with
  a fixed learning rate.
\newblock In \emph{The 22th International Conference on Artificial Intelligence
  and Statistics}, pp.\  3051--3059, Apr 2019.

\bibitem[Nagarajan \& Kolter(2019)Nagarajan and Kolter]{kolter19}
Nagarajan, V. and Kolter, J.~Z.
\newblock Uniform convergence may be unable to explain generalization in deep
  learning.
\newblock In \emph{Advances in Neural Information Processing Systems}, pp.\
  11611--11622, Dec 2019.

\bibitem[Oudah et~al.(2020)Oudah, Al-Naji, and Chahl]{Oudah20}
Oudah, M., Al-Naji, A., and Chahl, J.
\newblock Hand gesture recognition based on computer vision: A review of
  techniques.
\newblock \emph{Journal of Imaging}, 2020.

\bibitem[Phan(2021)]{Phan21}
Phan, H.
\newblock huyvnphan/pytorch\_cifar10, 1 2021.

\bibitem[Raghunathan et~al.(2020)Raghunathan, Xie, Yang, Duchi, and
  Liang]{raghunathan20}
Raghunathan, A., Xie, S.~M., Yang, F., Duchi, J., and Liang, P.
\newblock Understanding and mitigating the tradeoff between robustness and
  accuracy.
\newblock In \emph{International Conference on Machine Learning}, pp.\
  7909--7919, Jul 2020.

\bibitem[Rice et~al.(2020)Rice, Wong, and Kolter]{rice20}
Rice, L., Wong, E., and Kolter, Z.
\newblock Overfitting in adversarially robust deep learning.
\newblock In \emph{International Conference on Machine Learning}, pp.\
  8093--8104, Jul 2020.

\bibitem[Sagawa* et~al.(2020)Sagawa*, Koh*, Hashimoto, and Liang]{Sagawa20}
Sagawa*, S., Koh*, P.~W., Hashimoto, T.~B., and Liang, P.
\newblock Distributionally robust neural networks.
\newblock In \emph{International Conference on Learning Representations}, Apr
  2020.

\bibitem[Sanyal et~al.(2020)Sanyal, Dokania, Kanade, and Torr]{sanyal20}
Sanyal, A., Dokania, P.~K., Kanade, V., and Torr, P.
\newblock How benign is benign overfitting?
\newblock In \emph{International Conference on Learning Representations}, Apr
  2020.

\bibitem[Schmidt et~al.(2018)Schmidt, Santurkar, Tsipras, Talwar, and
  Madry]{schmidt18}
Schmidt, L., Santurkar, S., Tsipras, D., Talwar, K., and Madry, A.
\newblock Adversarially robust generalization requires more data.
\newblock In \emph{Advances in Neural Information Processing Systems}, pp.\
  5019--5031, Dec 2018.

\bibitem[Schneider et~al.(2020)Schneider, Rusak, Eck, Bringmann, Brendel, and
  Bethge]{Schneider20}
Schneider, S., Rusak, E., Eck, L., Bringmann, O., Brendel, W., and Bethge, M.
\newblock Improving robustness against common corruptions by covariate shift
  adaptation.
\newblock In Larochelle, H., Ranzato, M., Hadsell, R., Balcan, M.~F., and Lin,
  H. (eds.), \emph{Advances in Neural Information Processing Systems}, pp.\
  11539--11551, Dec 2020.

\bibitem[Springer et~al.(2021)Springer, Mitchell, and Kenyon]{springer21}
Springer, J.~M., Mitchell, M., and Kenyon, G.~T.
\newblock Adversarial perturbations are not so weird: Entanglement of robust
  and non-robust features in neural network classifiers.
\newblock \emph{arXiv preprint arXiv:2102.05110}, 2021.

\bibitem[Stutz et~al.(2019)Stutz, Hein, and Schiele]{Stutz19}
Stutz, D., Hein, M., and Schiele, B.
\newblock Disentangling adversarial robustness and generalization.
\newblock In \emph{IEEE/CVF Conference on Computer Vision and Pattern
  Recognition}, pp.\  6967--6987, Jun 2019.

\bibitem[Szegedy et~al.(2014)Szegedy, Zaremba, Sutskever, Bruna, Erhan,
  Goodfellow, and Fergus]{szegedy14}
Szegedy, C., Zaremba, W., Sutskever, I., Bruna, J., Erhan, D., Goodfellow, I.,
  and Fergus, R.
\newblock Intriguing properties of neural networks.
\newblock In \emph{International Conference on Learning Representations}, apr
  2014.

\bibitem[Telgarsky(2013)]{telgarsky13}
Telgarsky, M.
\newblock Margins, shrinkage, and boosting.
\newblock In \emph{International Conference on Machine Learning}, pp.\
  307--315, Jun 2013.

\bibitem[Tsipras et~al.(2019)Tsipras, Santurkar, Engstrom, Turner, and
  Madry]{tsipras19}
Tsipras, D., Santurkar, S., Engstrom, L., Turner, A., and Madry, A.
\newblock Robustness may be at odds with accuracy.
\newblock In \emph{International Conference on Learning Representations}, May
  2019.

\bibitem[Vershynin(2010)]{vershynin12}
Vershynin, R.
\newblock Introduction to the non-asymptotic analysis of random matrices.
\newblock \emph{arXiv preprint arXiv:1011.3027}, 2010.

\bibitem[Welinder et~al.(2010)Welinder, Branson, Mita, Wah, Schroff, Belongie,
  and Perona]{Welinder10}
Welinder, P., Branson, S., Mita, T., Wah, C., Schroff, F., Belongie, S., and
  Perona, P.
\newblock {Caltech-UCSD Birds 200}.
\newblock Technical Report CNS-TR-2010-001, California Institute of Technology,
  2010.

\bibitem[Wong et~al.(2020)Wong, Rice, and Kolter]{Wong20Fast}
Wong, E., Rice, L., and Kolter, J.~Z.
\newblock Fast is better than free: Revisiting adversarial training.
\newblock In \emph{International Conference on Learning Representations}, Apr
  2020.

\bibitem[Wu et~al.(2020)Wu, Tong, and Vorobeychik]{Wu20}
Wu, T., Tong, L., and Vorobeychik, Y.
\newblock Defending against physically realizable attacks on image
  classification.
\newblock In \emph{International Conference on Learning Representations}, Apr
  2020.

\bibitem[Xu et~al.(2020)Xu, Zhang, Ni, Li, Wang, Tian, and Zhang]{xu20}
Xu, M., Zhang, J., Ni, B., Li, T., Wang, C., Tian, Q., and Zhang, W.
\newblock Adversarial domain adaptation with domain mixup.
\newblock In \emph{AAAI Conference on Artificial Intelligence}, pp.\
  6502--6509, Feb 2020.

\bibitem[Yang et~al.(2013)Yang, Premaratne, and Vial]{Yang13}
Yang, S., Premaratne, P., and Vial, P.
\newblock Hand gesture recognition: An overview.
\newblock In \emph{IEEE International Conference on Broadband Network
  Multimedia Technology}, pp.\  63--69, 2013.

\bibitem[Yin et~al.(2019)Yin, Kannan, and Bartlett]{Yin19}
Yin, D., Kannan, R., and Bartlett, P.
\newblock Rademacher complexity for adversarially robust generalization.
\newblock In \emph{International conference on machine learning}, pp.\
  7085--7094, Jun 2019.

\bibitem[Zhai et~al.(2019)Zhai, Cai, He, Dan, He, Hopcroft, and Wang]{zhai20}
Zhai, R., Cai, T., He, D., Dan, C., He, K., Hopcroft, J., and Wang, L.
\newblock Adversarially robust generalization just requires more unlabeled
  data.
\newblock \emph{arXiv preprint arXiv:1906.00555}, 2019.

\bibitem[Zhang et~al.(2019)Zhang, Yu, Jiao, Xing, Ghaoui, and Jordan]{zhang19}
Zhang, H., Yu, Y., Jiao, J., Xing, E., Ghaoui, L.~E., and Jordan, M.
\newblock Theoretically principled trade-off between robustness and accuracy.
\newblock In \emph{International Conference on Machine Learning}, pp.\
  7472--7482, Jun 2019.

\bibitem[Zhao et~al.(2020)Zhao, Liu, and Larson]{zhao20}
Zhao, Z., Liu, Z., and Larson, M.
\newblock Towards large yet imperceptible adversarial image perturbations with
  perceptual color distance.
\newblock In \emph{IEEE/CVF Conference on Computer Vision and Pattern
  Recognition}, pp.\  1039--1048, 2020.

\bibitem[Zhou et~al.(2017)Zhou, Lapedriza, Khosla, Oliva, and Torralba]{zhou17}
Zhou, B., Lapedriza, A., Khosla, A., Oliva, A., and Torralba, A.
\newblock Places: A 10 million image database for scene recognition.
\newblock \emph{IEEE Transactions on Pattern Analysis and Machine
  Intelligence}, 2017.

\bibitem[Zhou et~al.(2020)Zhou, Liang, and Chen]{Zhou20}
Zhou, J., Liang, C., and Chen, J.
\newblock Manifold projection for adversarial defense on face recognition.
\newblock In \emph{European Conference on Computer Vision}, pp.\  288–305,
  Aug 2020.

\end{thebibliography}

\appendix

\section{Theoretical statements for the linear model}

\label{sec:app_theorylinear}
Before we present the proof of the theorem, we introduce two lemmas are of separate interest that are used throughout the proof of Theorem 1. Recall that the definition of the (standard normalized) maximum-$\ell_2$-margin solution (max-margin solution in short) of a dataset $\data =\{(x_i, y_i)\}_{i=1}^n$ corresponds to
\begin{equation}
  \label{eq:stdmaxmargin}
  \thetahat{} := \argmax_{\|\theta\|_2\leq 1} \min_{i\in [n]} y_i \theta^\top x_i,
\end{equation}
by simply setting $\epstrain = 0$ in Equation~\eqref{eq:maxmargin}. The $\ell_2$-margin of $\thetahat{}$ then reads $\min_{i\in[n]} y_i \thetahat{\top} x_i$. Furthermore for a dataset $\data = \{(x_i, y_i)\}_{i=1}^n$ we refer to the induced dataset $\datanonsig$ as the dataset with covariate vectors stripped of the first element, i.e.
\begin{equation}
  \datanonsig = \{(\xnonsig_i, y_i)\}_{i=1}^n :=  \{ ((x_i)_{[2:d]}, y_i) \}_{i=1}^n, 
\end{equation}
where $(x_i)_{[2:d]}$ refers to the last $d-1$ elements of the vector $x_i$. Furthermore, remember that for any vector $z$, $\indof{z}{j}$ refers to the $j$-th element of $z$ and $e_j$ denotes the $j$-th canonical basis vector.
Further, recall the distribution $\prob_\sigsep$ as defined in
Section~\ref{logreg_linear_model}: the label $y \in \{+1, -1\}$ is
drawn with equal probability and the covariate vector is sampled as $x
= [y\frac{\sigsep}{2}, \xnonsig]$ where $\xnonsig \in \R^{\dims-1}$ is
a random vector drawn from a standard normal distribution,
i.e. $\xnonsig \sim \Normal(0, \sigma^2 I_{d-1})$. We generally allow
$\sigsep$, used to sample the training data, to differ from $\sigseptest$, which is
used during test time.

The following lemma derives a closed-form expression for the normalized max-margin solution for any dataset with fixed separation $\sigsep$ in the signal component, and that is linearly separable in the last $d-1$ coordinates with margin $\marginnonsig$.

\begin{lemma}
\label{lem:maxmargin}
Let $\data = \{(x_i,y_i)\}_{i=1}^{\numsamp}$ be a dataset that
consists of points $(x,y) \in \mathbb{R}^{\dims}\times\{\pm 1\}$ and
$\xind{1} = y\frac{\sigsep}{2}$, i.e. the covariates $x_i$ are
deterministic in their first coordinate given $y_i$ with
separation distance $\sigsep$. Furthermore, let the induced dataset
$\datanonsig$ also be linearly separable by the normalized
max-$\ell_2$-margin solution $\thetatilde$ with an $\ell_2$-margin 
$\marginnonsig$. Then, the normalized max-margin solution of the
original dataset $\data$ is given by
\begin{equation}
\label{eq:lemmaxmargin}
\thetahat{} = \frac{1}{\sqrt{\sigsep^2 + 4 \marginnonsig^{2}}}\left[\sigsep,  2 \marginnonsig \thetatilde \right].
\end{equation}
Further, the standard accuracy of $\thetahat{}$ for data drawn from $\prob_{\sigseptest}$ reads
\begin{equation}
  \label{eq:stdaccmaxmargin}
  \prob_{\sigseptest}(Y \thetahat{\top} X > 0) = \Phi\left(
  \frac{\sigsep \:\sigseptest }{4\mixvar\: \marginnonsig} \right).
\end{equation}
\end{lemma}
The proof can be found in Section~\ref{sec:maxmarginproof}. The next lemma provides high probability upper and lower bounds
for the margin $\marginnonsig$ of $\datanonsig$ when $\xnonsig_i$ are drawn from the normal distribution.
\begin{lemma}
\label{lem:boundsmaxmargin}
Let $\datanonsig=\{(\Tilde{x}_i,y_i)\}_{i=1}^{\numsamp}$ be a random dataset where $y_i \in \{\pm 1\}$ are equally distributed and $\xnonsig_i \sim \Normal(0,\sigma I_{d-1})$ for all $i$, and $\marginnonsig$ is the maximum $\ell_2$ margin that can be written as
\begin{equation*}
  \marginnonsig= \max_{\|\thetatilde\|_2 \leq 1} \min_{i \in [\numsamp]} y_i \thetatilde^{\top} \Tilde{x}_i .
\end{equation*}
Then, for any $t \geq 0$, with probability greater than $1-2e^{-\frac{t^2}{2}}$, we have $\minmargin(t) \leq \marginnonsig \leq \maxmargin(t)$ where
\begin{align*}
  \label{Crude_bounds_subsequent_maxmar}
  &\maxmargin(t) = \mixvar \left( \sqrt{\frac{\dims-1}{\numsamp}} + 1  + \frac{t}{\sqrt{n}}\right), \:\: \minmargin(t)= \mixvar \left( \sqrt{\frac{\dims-1}{\numsamp}} -1 - \frac{t}{\sqrt{n}}\right).
\end{align*}  
\end{lemma}

\subsection{Proof of Theorem~\ref{thm:linlinf}}
\label{sec:thmproof}

Given a dataset $\data = \{(x_i, y_i)\}$ drawn from $\prob_\sigsep$, it is easy to see that the (normalized) $\epstrain$-robust max-margin solution~\eqref{eq:maxmargin} of $\data$ with respect to signal-attacking perturbations $\pertset{\epstrain}{x_i}$ as defined in Equation~\eqref{eq:linfmaxpert}, can be written as
\begin{equation}
\begin{aligned}
  \label{eq:robmaxmargin}
  \thetahat{\epstrain} &= \argmax_{\|\theta\|_2\leq 1}  \min_{i\in [n], x_i' \in \pertset{x_i}{\epstrain}} y_i \theta^\top x'_i \\
  &= \argmax_{\|\theta\|_2\leq 1}\min_{i\in [n],|\beta|\leq \epstrain}y_i \theta^\top (x_i + \beta e_1) \nonumber\\
  &= \argmax_{\|\theta\|_2\leq 1} \min_{i\in [n]} y_i \theta^\top (x_i - y_i \epstrain \sign(\thetaind{1}) e_1). \nonumber
\end{aligned}
\end{equation}
Note that by definition, it is equivalent to the (standard normalized)
max-margin solution $\thetahat{}$ of the shifted dataset ${\Dshift =
  \{(x_i - y_i \epstrain \sign(\thetaind{1}) e_1,
  y_i)\}_{i=1}^n}$. Since $\Dshift$ satisfies the assumptions of
Lemma~\ref{lem:maxmargin}, it then follows directly that the
normalized $\epstrain$-robust max-margin solution reads
\begin{equation}
  \label{eq:appmaxmargin}
  \thetahat{\epstrain} = \frac{1}{\sqrt{(\sigsep -2\epstrain)^2 + 4 \marginnonsig^{2}}}\left[\sigsep-2\epstrain,  2 \marginnonsig \thetatilde \right],
\end{equation}
by replacing $\sigsep$ by $\sigsep - 2\epstrain$ in
Equation~\eqref{eq:lemmaxmargin}. Similar to above, $\thetatilde \in
R^{d-1}$ is the (standard normalized) max-margin solution of
$\{(\xnonsig_i, y_i)\}_{i=1}^n$ and $\marginnonsig$ the corresponding
margin.

\paragraph{Proof of 1.}
We can now compute the $\epstest$-robust accuracy of the
$\epstrain$-robust max-margin estimator $\thetahat{\epstrain}$ for a
given dataset $\data$ as a function of $\marginnonsig$. Note that in
the expression of $\thetahat{\epstrain}$, all values are fixed for a
fixed dataset, while $0\leq \epstrain\leq \sigsep-2\maxmargin$ can be chosen.
First note that for a test distribution $\prob_\sigsep$, the
$\epstest$-robust accuracy, defined as one minus the robust error (Equation~\eqref{eq:roberr}), for a classifier
associated with a vector $\theta$, can be written as
\begin{align}
  \label{eq:robacc_closed}
  \robacc{\theta} &= \EE_{X,Y\sim \prob_\sigsep} \left[\Indi{\min_{x'
        \in \pertset{X}{\epstest}} Y \theta^\top x'>0}\right] \\
  &=   \EE_{X,Y\sim \prob_{\sigsep}} \left[ \Indi{ Y \theta^\top X -
      \epstest \thetaind{1} >0}\right] = \EE_{X,Y\sim \prob_{\sigsep}}
  \left[\Indi{ Y \theta^\top (X - Y\epstest \sign(\thetaind{1}) e_1) >0}\right]
  \nonumber
\end{align}
Now, recall that
by Equation~\eqref{eq:appmaxmargin} and the assumption in the
theorem, we have $\sigsep-2\epstrain>0$, so that $\sign(\thetahat{\epstrain})=1$.
Further, using the definition of the $\pertset{\epstrain}{x}$ in
Equation~\eqref{eq:linfmaxpert} and by definition of the
distribution $\prob_\sigsep$, we have $\indof{X}{1} = Y
\frac{\sigsep}{2}$.
Plugging into Equation~\eqref{eq:robacc_closed} then yields
\begin{align*}
  \robacc{\thetahat{\epstrain}}&= \EE_{X,Y\sim \prob_{\sigsep}} \left[\Indi{ Y \thetahat{\epstrain \top} (X - Y\epstest  e_1) >0}\right] \\
  &=   \EE_{X,Y\sim \prob_{\sigsep}}\left[\Indi{ Y \thetahat{\epstrain \top} (X_{-1} + Y\left(\frac{\sigsep}{2} - \epstest\right)  e_1) >0}\right] \\
  &= \prob_{\sigsep- 2 \epstest} (Y\thetahat{\epstrain \top} X >0 )
\end{align*}
where $X_{-1}$ is a shorthand for the random vector $X_{-1} = (0;
  \indof{X}{2}, \dots, \indof{X}{d})$.  The assumptions in
Lemma~\ref{lem:maxmargin} ($\Dshift$ is linearly separable) are
satisfied whenever the $n<d-1$ samples are distinct, i.e. with
probability one. Hence applying Lemma~\ref{lem:maxmargin} with
$\sigseptest = \sigsep - 2\epstest$ and $\sigsep = \sigsep -
2\epstrain$ yields
\begin{equation}
  \label{eq:arsenal}
  \robacc{\thetahat{\epstrain}} =
  \Phi\left(\frac{\sigsep(\sigsep-2\epstest)}{4\mixvar \marginnonsig}
  - \epstrain \frac{\sigsep-2\epstest}{2\mixvar \marginnonsig}\right).
\end{equation}
Theorem statement a) then follows by noting that
$\Phi$ is a monotically decreasing function in $\epstrain$.
The expression for the robust error then follows by noting that $1-\Phi(-z) = \Phi(z)$ for any $z \in \R$
and defining
\begin{equation}
  \label{eq:varphidef}
  \randvarphi = \frac{\sigma \marginnonsig}{\sigsep/2 - \epstest}.
\end{equation}

\paragraph{Proof of 2.}
First define $\varphimin, \varphimax$ using $\minmargin, \maxmargin$ as in Equation~\eqref{eq:varphidef}. Then we have by Equation~\eqref{eq:arsenal}
\begin{align*}
  \roberr{\thetahat{\epstrain}} - \roberr{\thetahat{0}} &= \robacc{\thetahat{0}} - \robacc{\thetahat{\epstrain}}\\
  &=   \Phi\left(\frac{\sigsep/2}{\randvarphi}\right) - \Phi\left(\frac{\sigsep/2 - \epstrain}{\randvarphi}\right)\\
  &= \int_{r/2-\epstrain}^{r/2} \frac{1}{\sqrt{2\pi}\randvarphi} \E^{- \frac{x^2 }{\randvarphi^2}} d x
\end{align*}

By plugging in $t = \sqrt{\frac{2 \log 2/\delta}{\numsamp}}$ in
Lemma~\ref{lem:boundsmaxmargin}, we obtain that with probability at
least $1-\delta$ we have
\begin{equation*}
   \minmargin := \mixvar 
                \left[\sqrt{\frac{d-1}{n}} - \left(1+\sqrt{\frac{2 \log (2/\delta)}{\numsamp}}\right)\right] \leq \marginnonsig \leq \mixvar 
                \left[\sqrt{\frac{d-1}{n}} + \left(1+\sqrt{\frac{2 \log (2/\delta)}{\numsamp}}\right)\right] =: \maxmargin
\end{equation*}
and equivalently $\varphimin \leq \randvarphi \leq \varphimax$.

Now note the general fact that for all
$\randvarphi \leq \sqrt{2} x$ the density function  
$f(\randvarphi; x) = \frac{1}{\sqrt{2\pi}\randvarphi} \E^{- \frac{x^2 }{\randvarphi^2}} $
is monotonically increasing in $\randvarphi$.

By assumption of the theorem, $\randvarphi \leq \sqrt{2} (\sigsep/2-\epstrain)(\sigsep/2-\epstest)$ so that $f(\randvarphi; x) \geq f(\varphimin;x)$ for all $x\in [\sigsep/2-\epstrain,\sigsep/2]$ and therefore
\begin{equation*}
   \int_{r/2-\epstrain}^{r/2} \frac{1}{\sqrt{2\pi}\randvarphi} \E^{- \frac{x^2 }{\randvarphi^2}} d x \geq  \int_{r/2-\epstrain}^{r/2} \frac{1}{\sqrt{2\pi}\varphimin} \E^{- \frac{x^2 }{\randvarphi^2}} d x = \Phi\left(\frac{r/2}{\varphimin}\right) - \Phi\left(\frac{r/2-\epstrain}{\varphimin}\right).
\end{equation*}
and the statement is proved.


\subsection{Proof of Corollary~\ref{cor:l1extension}}
We now show that Theorem~\ref{thm:linlinf} also holds for
$\ell_1$-ball perturbations with at most radius $\eps$.  Following
similar steps as in Equation~\eqref{eq:appmaxmargin}, the
$\epstrain$-robust max-margin solution for $\ell_1$-perturbations can
be written as
\begin{equation}
  \label{eq:maxmarginl1}
  \thetahat{\epstrain} := \argmax_{\|\theta\|_2 \leq 1}\min_{i\in [n]}  y_i \theta^\top (x_i  - y_i  \epstrain \sign(\indof{\theta}{\maxind(\theta)}) e_{\maxind(\theta)})
\end{equation}
where $\maxind(\theta) := \argmax_j |\theta_j|$ is the index of the maximum absolute value of $\theta$.
We now prove by contradiction that the robust max-margin solution for
this perturbation set~\eqref{eq:l1maxpert} is equivalent to the solution~\eqref{eq:appmaxmargin} for the perturbation set~\eqref{eq:linfmaxpert}.
We start by assuming that $\thetaA$ does not solve
Equation~\eqref{eq:appmaxmargin}, which is equivalent to assuming $1\not \in
\maxind(\thetaA)$ by definition. We now show how this assumption leads
to a contradiction.

Define the shorthand $\maxindA := \maxind(\thetaA) -1$. Since
$\thetaA$ is the solution of~\eqref{eq:maxmarginl1}, by definition, we
have that $\thetaA$ is also the max-margin solution of the shifted
dataset $\Dshift :=(x_i - y_i \epstrain \sign(\thetaind{\maxindA+1})
e_{\maxindA+1}, y_i)$.  Further, note that by the assumption that $1
\not \in \maxind(\thetaA)$, this dataset $\Dshift$ consists of input
vectors $x_i = (y_i \frac{\sigsep}{2}, \xnonsig_i - y_i \epstrain
\sign(\thetaind{\maxindA+1}) e_{\maxindA+1} )$.  Hence via
Lemma~\ref{lem:maxmargin}, $\thetaA$ can be written as
\begin{equation}
  \label{eq:sml}
       \thetaA = \frac{1}{\sqrt{\sigsep^2 - 4 (\marginnonsig^{\epstrain})^2}} [\sigsep, 2 \marginnonsig^{\epstrain} \thetatilde^{\epstrain}],
\end{equation}
where $\thetatilde^{\epstrain}$ is the normalized max-margin solution
of  $\datanonsig := (\xnonsig_i
- y_i \epstrain \sign(\indof{\thetatilde}{\maxindA}) e_{\maxindA},
y_i)$.

We now characterize $\thetatilde^{\epstrain}$. Note that by
assumption, $\maxindA = \maxind(\thetatilde^{\epstrain}) = \argmax_j
|\indof{\thetatilde^{\epstrain}}{j}|$. Hence, the normalized max-margin
solution $\thetatilde^{\epstrain}$ is the solution of
\begin{equation}
  \label{eq:maxmarginsmall}
  \thetatilde^{\epstrain} := \argmax_{\|\thetatilde\|_2 \leq 1}
  \min_{i\in [n]} y_i \thetatilde^\top \xnonsig_i - \epstrain
  |\indof{\thetatilde}{\maxindA}| 
\end{equation}
Observe that the minimum margin of this estimator
$\marginnonsig^{\epstrain}=\min_{i\in [n]} y_i
(\thetatilde^{\epstrain})^\top \xnonsig_i - \epstrain
|\indof{\thetatilde^{\epstrain}}{\maxindA}|$ decreases with
$\epstrain$ as the problem becomes harder $\marginnonsig^{\epstrain}
\leq \marginnonsig$, where the latter is equivalent to the margin of
$\thetatilde^{\epstrain}$ for $\epstrain = 0$.  Since $\sigsep >
2\maxmargin$ by assumption in the Theorem, by Lemma~\ref{lem:boundsmaxmargin}
 with probability at least $1-2\E^{-\frac{\tconst^2 (d-1)}{n}}$, we then have that $\sigsep> 2\marginnonsig \geq 2\marginnonsig^{\epstrain}$. Given
the closed form of $\thetaA$ in Equation~\eqref{eq:sml}, it
directly follows that $\indof{\thetaA}{1} = \sigsep >
2\marginnonsig^{\epstrain} \|\thetatilde^{\epstrain}\|_2 =
\|\indof{\thetaA}{2:d}\|_2$ and hence $1\in \maxind(\thetaA)$. This
contradicts the original assumption $1\not \in \maxind(\thetaA)$ and
hence we established that $\thetahat{\epstrain}$ for the
$\ell_1$-perturbation set~\eqref{eq:l1maxpert} has the same closed
form~\eqref{eq:robmaxmargin} as for the perturbation
set~\eqref{eq:linfmaxpert}.

The final statement is proved by using the analogous steps as in
the proof of 1. and 2. to obtain the closed form of the robust accuracy of
$\thetahat{\epstrain}$.

\subsection{Proof of Lemma~\ref{lem:maxmargin}}
\label{sec:maxmarginproof}

We start by proving that $\thetahat{}$ is of the form
\begin{equation}
\label{Eq:max_margin_param_form_total_D}
\thetahat{} = \left[\constone, \constwo \thetatilde \right],
\end{equation}
for $\constone, \constwo > 0$. Denote by $\decplanegen{\theta}$ the plane through the origin with normal $\theta$. We define $d\left((x,y), \decplanegen{\theta} \right)$ as the signed euclidean distance from the point $(x,y) \in \data \sim \prob_{\sigsep}$ to the plane $\decplanegen{\theta}$. The signed euclidean distance is the defined as the euclidean distance from x to the plane if the point $(x,y)$ is correctly predicted by $\theta$, and the negative euclidean distance from $x$ to the plane otherwise. We rewrite the definition of the max $l_2$-margin classifier. It is the classifier induced by the  normalized vector $\thetahat{}$, such that 
\begin{equation*}
\max_{\theta \in \mathbb{R}^{\dims}} \min_{(x,y) \in \data}d\left( \left(x,y\right),\decplanegen{\theta}\right)  = \min_{(x,y) \in \data} d\left( \left(x,y \right),\decplanegen{\thetahat }\right).
\end{equation*}
We use that $\data$ is deterministic in its first coordinate and get
\begin{equation*}
\begin{split}
	\max_{\theta}\min_{(x,y) \in \data}d\left(\left(x,y\right), \decplanegen{\theta} \right) &= \max_{\theta}\min_{(x,y) \in \data} y (\thetaind{1} \xind{1} + \thetatilde^{\top} \xnonsig)\\
	&= \max_{\theta}  \theta_1  \frac{r}{2} + \min_{(x,y) \in \data}  y \thetatilde^{\top} \Tilde{x}.
	\end{split}
\end{equation*}
Because $\sigsep >0$, the maximum over all $\theta$ has $\thetahatind{}{1} \geq 0$. Take any $a > 0$ such that $\|\thetatilde\|_2 = a$.  By definition the max $l_2$-margin classifier, $\thetatilde$, maximizes $\min_{(x,y) \in \data} d\left(\left(x,y\right), \decplanegen{\theta} \right)$. Therefore, $\thetahat{}$ is of the form of Equation \eqref{Eq:max_margin_param_form_total_D}. 

Note that all classifiers induced by vectors of the form of Equation \eqref{Eq:max_margin_param_form_total_D} classify $\data$ correctly.  Next, we aim to find expressions for $\constone$ and $\constwo$ such that Equation \eqref{Eq:max_margin_param_form_total_D} is the normalized max $l_2$-margin classifier. The distance from any $x \in \data$ to $\decplanegen{\thetahat{}}$ is
\begin{equation*}
d\left(x,\decplanegen{\thetahat{}} \right) = \left| \constone \xind{1}  + \constwo \thetatilde^{\top} \xnonsig \right|.
\end{equation*}
Using that $\xind{1} = y \frac{\sigsep}{2}$ and that the second term equals $\constwo d\left(x, \decplanegen{\thetatilde}\right)$, we get
\begin{equation}
\label{eq:distance_to_opt_intermidate}
d\left(x,  \decplanegen{\thetahat{}}\right) =  \left| \constone \frac{\sigsep}{2}  + \constwo d\left(x, \decplanegen{\thetatilde}\right) \right| = \constone \frac{\sigsep}{2}  + \sqrt{1-\constone^2} d\left(x, \decplanegen{\thetatilde}\right).
\end{equation}
Let $(\xnonsig,y) \in \datanonsig$ be the point closest in Euclidean
distance to $\thetatilde$. This point is also the closest point in
Euclidean distance to $\decplanegen{\thetahat{}}$, because by Equation
\eqref{eq:distance_to_opt_intermidate} $d\left(x,
\decplanegen{\thetahat{}}\right)$ is strictly decreasing for
decreasing $d\left(x, \decplanegen{\thetatilde}\right)$. We maximize
the minimum margin $d\left(x, \decplanegen{\thetahat{}} \right)$ with
respect to $\constone$. Define the vectors $a = \left[\constone,
  \constwo\right]$ and $v = \left[\frac{\sigsep}{2}, d\left(x,
  \decplanegen{\thetatilde}\right)\right]$. We find using the dual
norm that
\begin{equation*}
a = \frac{v}{\|v\|_2}.
\end{equation*}
Plugging the expression of $a$ into Equation
\eqref{Eq:max_margin_param_form_total_D} yields that $\thetahat{}$ is
given by
\begin{equation*}
	\thetahat{} = \frac{1}{\sqrt{\sigsep^2 + 4 \marginnonsig^2}}\left[\sigsep,  2 \marginnonsig\thetatilde \right].
\end{equation*}

For the second part of the lemma we first decompose
\begin{equation}
  \label{eq:jacob}
\prob_{\sigseptest} (Y\thetahat{\top} X >0 ) = \frac{1}{2}\prob_{\sigseptest} \left[ \thetahat{\top} X >0 \mid Y=1 \right]  +\frac{1}{2}\prob_{\sigseptest} \left[\thetahat{\top} X <0 \mid Y=-1\right]\nonumber
\end{equation}
We can further write 
\begin{align}
  \label{eq:cumul1}
\prob_{\sigseptest} \left[\thetahat{\top} X > 0 \mid
  Y = 1\right] &=\prob_{\sigseptest} \left[\sum_{i=2}^{\dims}\indof{\thetahat{}}{i} \indof{X}{i} > -
  \indof{\thetahat{}}{1} \: \indof{X}{1} \mid Y=1\right]\\
&= \prob_{\sigseptest} \left[2 \marginnonsig \sum_{i=1}^{\dims-1}\indof{\thetatilde}{i} \indof{X}{i} > -
  \sigsep \: \frac{\sigseptest}{2} \mid Y=1\right]\nonumber\\
&= 1-\Phi\left(-\frac{\sigsep\: \sigseptest}{4\mixvar \marginnonsig} \right) =
\Phi\left(\frac{\sigsep \: \sigseptest}{4\mixvar \marginnonsig} \right) \nonumber
\end{align}
where $\Phi$ is the cumulative distribution function. The second equality
follows by multiplying by the normalization constant on both sides and the
third equality is due to the fact that $\sum_{i=1}^{\dims-1}\indof{\thetatilde}{i} \indof{X}{i}$ is
a zero-mean Gaussian with variance $\sigma^2\|\thetatilde\|^2_2 = \sigma^2$ since $\thetatilde$ is normalized.
Correspondingly we can write
\begin{align}
  \label{eq:cumul2}
\prob_{\sigseptest} \left[\thetahat{\top} X < 0 \mid
  Y = -1\right] &=\prob_{\sigseptest} \left[2\marginnonsig
  \sum_{i=1}^{\dims-1}\indof{\thetatilde}{i} \indof{X}{i} < -
  \sigsep \left(- \frac{\sigseptest}{2}\right) \mid Y=-1\right] = \Phi\left(\frac{\sigsep \:\sigseptest}{4\mixvar \marginnonsig}\right) 
\end{align}
so that we can
combine~\eqref{eq:jacob} and~\eqref{eq:cumul1} and \eqref{eq:cumul2} to obtain
$\prob_{\sigseptest} (Y\thetahat{\top} X >0 ) = \Phi \left(\frac{\sigsep \:\sigseptest}{4\mixvar \marginnonsig}\right)$. This concludes the proof of the lemma.

\subsection{Proof of Lemma \ref{lem:boundsmaxmargin}}
\label{sec:boundsmaxmargin}

The proof plan is as follows. We start from the definition of the max
$\ell_2$-margin of a dataset. Then, we rewrite the
max $\ell_2$-margin as an expression that includes a random matrix with independent
standard normal entries. This allows us to prove the upper and lower bounds for the
max-$\ell_2$-margin in Sections~\ref{sec:gammaupperbound} and ~\ref{sec:gammalowerbound}
respectively, using non-asymptotic estimates on the singular values of
Gaussian random matrices.

Given the dataset $\datanonsig =  \{(\xnonsig_i, y_i)\}_{i=1}^{\numsamp}$, we define the random matrix
\begin{equation}
\label{eq:randmatrixsamples}
\randdatamatr = \begin{pmatrix}
\xnonsig_1^{\top}\\
\xnonsig_2^{\top}\\
...\\
\xnonsig_{\numsamp}^{\top}
\end{pmatrix}.
\end{equation}
where $\xnonsig_i \sim \Normal(0,\sigma I_{d-1})$. 
Let $\mathcal{V}$ be the class of all perfect predictors of $\datanonsig$. For a matrix $A$ and vector $b$ we also denote by $|Ab|$ the vector whose entries correspond to the absolute values of the entries of $Ab$. 
Then, by definition
\begin{equation}
\label{maxmargindefgammaproof}
\marginnonsig = \max_{v \in \mathcal{V}, \|v\|_2=1} \min_{j \in [\numsamp]} \indof{|\randdatamatr v|}{j} = \max_{v \in \mathcal{V}, \|v\|_2=1} \min_{j \in [\numsamp]} \mixvar \indof{|\gaussianmatrix v|}{j},
\end{equation}
where $\gaussianmatrix = \frac{1}{\sigma} \randdatamatr$ is the scaled data matrix.

In the sequel we will use the operator norm of a matrix $A \in \mathbb{R}^{\numsamp \times \dims-1}$.
\begin{equation*}
\| A\|_2 = \sup_{v \in \mathbb{R}^{\dims-1} \mid \|v\|_2=1} \|A v \|_2
\end{equation*}
and denote the maximum singular value of a matrix $A$ as $s_{\text{max}} (A)$ and the minimum singular value as $s_{\text{min}}(A)$.

\subsubsection{Upper bound}
\label{sec:gammaupperbound}

Given the maximality of the
operator norm and since the minimum entry of the vector
$|\gaussianmatrix v|$ must be smaller than $\frac{\|\gaussianmatrix\|_2}{\sqrt{\numsamp}}$, we can upper bound
$\marginnonsig$ by
\begin{equation*}
\marginnonsig  \leq \mixvar  \frac{1}{\sqrt{\numsamp}} \|\gaussianmatrix{}\|_2.
\end{equation*}
Taking the expectation on both sides with respect to the draw of
$\datanonsig$ and noting $\|\gaussianmatrix\|_2 \leq
s_{\text{max}}\left(\gaussianmatrix\right)$,
it follows from
Corollary 5.35  of \cite{vershynin12} 
that for all $t\geq 0$:
\begin{equation*}
\prob \left[\sqrt{\dims-1}+\sqrt{\numsamp}+t \geq s_{\text{max}}\left(\gaussianmatrix\right) \right] \geq 1-2e^{-\frac{t^2}{2}}.
\end{equation*}
Therefore, with a probability greater than $1-2e^{-\frac{t^2}{2}}$,
\begin{equation*}
\marginnonsig \leq  \mixvar \left(1+ \frac{t+\sqrt{\dims-1}}{\sqrt{\numsamp}}\right).
\end{equation*}

\subsubsection{Lower bound}
\label{sec:gammalowerbound}
By the definition in Equation \eqref{maxmargindefgammaproof}, if we
find a vector $v \in \mathcal{V}$ with $\|v\|_2=1$ such that for an
$a>0$, it holds that $\Hquad \min_{j \in \numsamp} \sigma
\indof{|\randdatamatr v|}{j} > a$, then $\marginnonsig > a$.


Recall the definition of the max-$\ell_2$-margin
as in Equation \ref{eq:randmatrixsamples}.
As $\numsamp < \dims-1$, the random matrix $\gaussianmatrix$ is a wide
matrix, i.e. there are more columns than rows and therefore the
minimal singular value is $0$.
Furthermore, $\gaussianmatrix$ has rank $\numsamp$ almost surely and hence 
for all $\cst >0$, there exists a $v \in \mathbb{R}^{\dims-1}$ such that
\begin{equation}
\label{eq:existencerhs}
 \mixvar \gaussianmatrix v= 1_{\{ \numsamp\}}\cst> 0,
\end{equation}
where $ 1_{\{ \numsamp \}}$ denotes the all ones vector of dimension $\numsamp$. The smallest non-zero singular value of $\gaussianmatrix$, $s_{\text{min, nonzero}}(\gaussianmatrix)$, equals the smallest non-zero singular value of its transpose $\gaussianmatrix^{\top}$. Therefore, there also exists a $v \in \mathcal{V}$ with $\|v\|_2=1$ such that
\begin{equation}
\label{minimum_step_gamma}
\marginnonsig \geq  \min_{j \in [n]} \mixvar \indof{|\gaussianmatrix v|}{j} \geq \mixvar s_{\text{min,nonzeros}}\left(\gaussianmatrix^{\top}\right)\frac{1}{\sqrt{\numsamp}},
\end{equation}
where we used the fact that any vector $v$ in the span of non-zero eigenvectors satisfies $\|\gaussianmatrix  v \|_2 \geq s_{\text{min, nonzeros}}(\gaussianmatrix)$ and the existence of a solution $v$ for any right-hand side as in Equation \ref{eq:existencerhs}.
Taking the expectation on both sides,
Corollary 5.35 of \cite{vershynin12} yields that with a probability greater than $1-2e^{-\frac{t^2}{2}}, t\geq 0$ we have
\begin{equation}
\marginnonsig \geq \mixvar\left( \frac{\sqrt{\dims-1}-t}{\sqrt{\numsamp}}-1\right).
\end{equation}



\section{Bounds on the susceptibility score}
\label{app:susc}
In Theorem \ref{thm:linlinf}, we give non-asymptotic bounds on the robust and standard error of a linear classifier trained with adversarial logistic regression. Moreover, we use the robust error decomposition in susceptibility and standard error to gain intuition about how adversarial training may hurt robust generalization. In this section, we complete the result of Theorem \ref{thm:linlinf} by also deriving non-asymptotic bounds on the susceptibility score of the max $\ell_2$-margin classifier.

Using the results in Appendix \ref{sec:app_theorylinear}, we can prove following Corollary \ref{cor:robustness}, which gives non asymptotic bounds on the susceptibility score.
\begin{corollary}
\label{cor:robustness}
  Assume $d-1>n$. For the $\epstest$-susceptibility on test samples from $\prob_{\sigsep}$ with $2 \epstest < \sigsep$ and perturbation sets in Equation~\eqref{eq:linfmaxpert} and~\eqref{eq:l1maxpert} the following holds:

For $\epstrain < \frac{\sigsep}{2} - \maxmargin$, with probability at least $1-2\E^{-\frac{\tconst^2 (d-1)}{2}}$ for any $0<\tconst<1$, over the draw of a dataset $\data$ with $n$ samples from $\prob_{\sigsep}$, the $\epstest$-susceptibility is upper and lower bounded by
  \begin{equation}
  \begin{split}
       &\suscept{\thetaA} \leq \Phi \left(\frac{(\sigsep-2 \epstrain) (\epstest - \frac{\sigsep}{2})}{2 \maxmargin \sigma}\right) - \Phi \left( \frac{(\sigsep-2 \epstrain)( -\epstest - \frac{\sigsep}{2})}{2 \minmargin\sigma} \right)\\ 
       &\suscept{\thetaA} \geq   \Phi \left(\frac{(\sigsep-2 \epstrain) (\epstest - \frac{\sigsep}{2})}{2 \minmargin\sigma}\right) - \Phi \left( \frac{(\sigsep-2 \epstrain)( -\epstest - \frac{\sigsep}{2})}{2 \maxmargin \sigma} \right)
        \end{split}
  \end{equation}
\end{corollary}

We give the proof in Subsection \ref{sec:proof_robust_cor}. Observe that the bounds on the susceptibility score in Corollary \ref{cor:robustness} consist of two terms each, where the second term decreases with $\epstrain$, but the first term increases. We recognise following two regimes: the max $\ell_2$-margin classifier is close to the ground truth $e_1$ or not. Clearly, the ground truth classifier has zero susceptibility and hence classifiers close to the ground truth also have low susceptibility. On the other hand, if the max $l_2$-margin classifier is not close to the ground truth, then putting less weight on the first coordinate increases invariance to the perturbations along the first direction. Recall that by Lemma \ref{lem:maxmargin}, increasing $\epstrain$, decreases the weight on the first coordinate of the max $\ell_2$-margin classifier. Furthermore, in the low sample size regime, we are likely not close to the ground truth. Therefore, the regime where the susceptibility decreases with increasing $\epstrain$ dominates in the low sample size regime.

\begin{figure*}[!b]
  \centering
\begin{subfigure}[b]{0.4\textwidth}
 \centering
  \includegraphics[width=0.99\linewidth]{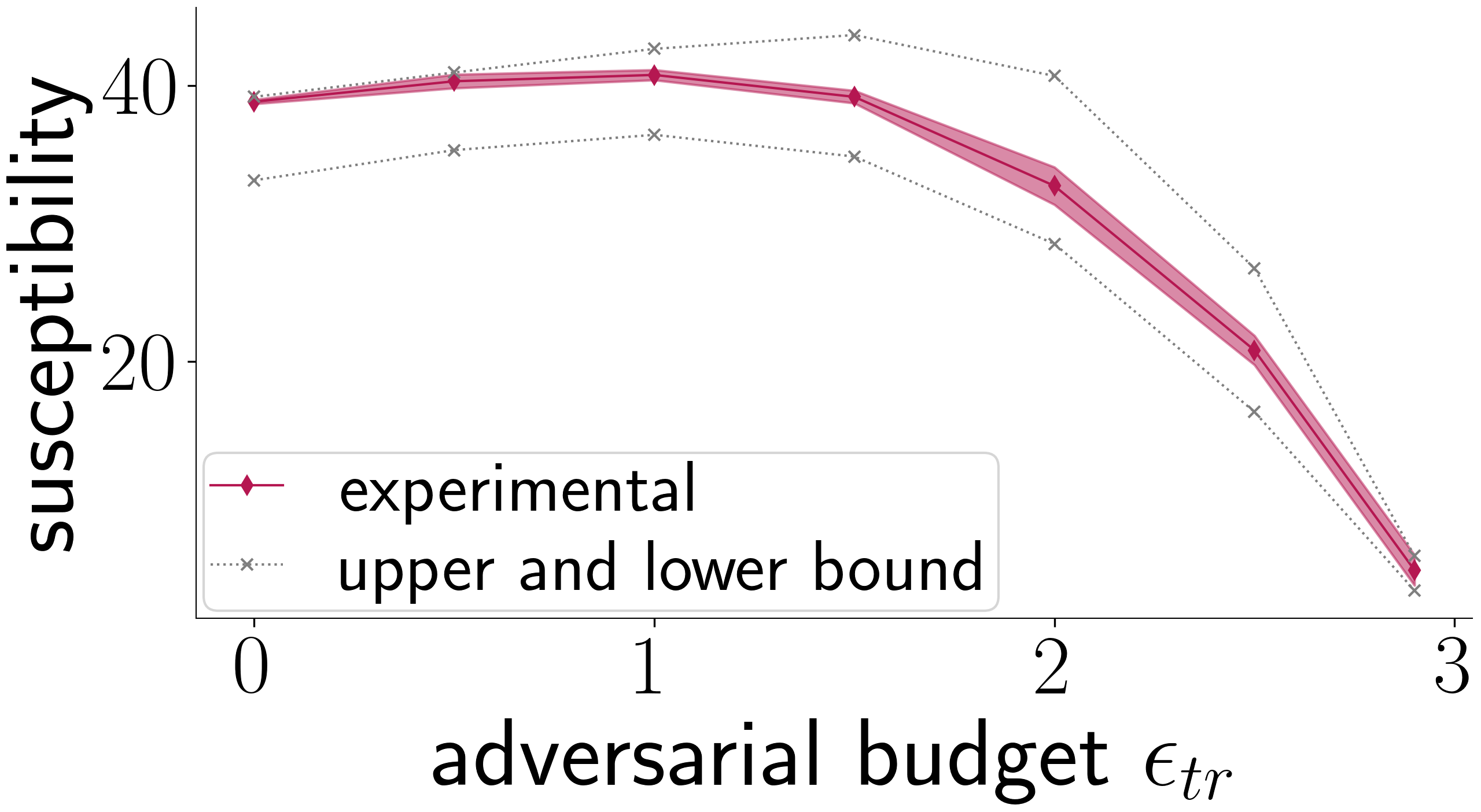}
  \caption{Susceptibility score decreases with $\epstrain$}
  \label{fig:app_robustness}
\end{subfigure}
\begin{subfigure}[b]{0.4\textwidth}
 \centering
  \includegraphics[width=0.99\linewidth]{plotsAistats/logreg_trade_off_plot.png}
  \caption{Robust error decomposition}
  \label{fig:app_tradeoff_logreg}
\end{subfigure}
\caption{We set $\sigsep = 6$, $\dims = 1000$, $\numsamp = 50$ and $\epstest = 2.5$. (a) We plot the average susceptibility score and the standard deviation over 5 independent experiments. Note how the bounds closely predict the susceptibility score. (b) For comparison, we also plot the robust error decomposition in susceptibility and standard error. Even though the susceptibility decreases, the robust error increases with increasing adversarial budget $\epstrain$.}
  \vspace{-0.2in}
\label{fig:logreg_robust}
\end{figure*}

To confirm the result of Corollary \ref{cor:robustness}, we plot the mean and standard deviation of the susceptibility score of $5$ independent experiments. The results are depicted in Figure \ref{fig:logreg_robust}. We see that for low standard error, when the classifier is reasonably close to the optimal classifier, the susceptibility increases slightly with increasing adversarial budget. However, increasing the adversarial training budget, $\epstrain$, further, causes the susceptibility score to drop greatly. Hence, we can recognize both regimes and validate that, indeed, the second regime dominates in the low sample size setting.

\subsection{Proof of Corollary \ref{cor:robustness}}
\label{sec:proof_robust_cor}
We proof the statement by bounding the robustness of a linear classifier. Recall that the robustness of a classifier is the probability that a classifier does not change its prediction under an adversarial attack. The susceptibility score is then given by 
\begin{equation}
\label{eq:rob_sus}
\suscept{\thetaA} = 1 - \robness{\thetaA}.
\end{equation}

The proof idea is as follows: since the perturbations are along the first basis direction, $e_1$, we compute the distance from the robust $l_2$-max margin $\thetaA$ to a point $(X,Y) \sim \prob$. Then, we note that the robustness of $\thetaA$ is given by the probability that the distance along $e_1$, from $X$ to the decision plane induced by $\thetaA$ is greater then $\epstest$. Lastly, we use the non-asymptotic bounds of Lemma \ref{lem:boundsmaxmargin}.

Recall, by Lemma \ref{lem:maxmargin}, the max $l_2$-margin classifier is of the form of
\begin{equation}
\label{eq:robustmaxmarg}
\thetaA = \frac{1}{\sqrt{(\sigsep-2 \epstrain)^2 + 4 \marginnonsig^{2}}}\left[\sigsep-2\epstrain,  2 \marginnonsig \thetatilde \right].
\end{equation}
Let $(X, Y) \sim \prob$. The distance along $e_1$ from $X$ to the decision plane induced by $\thetaA$, $\decplanegen{\thetaA }$, is given by
\begin{equation*}
d_{e_1}(X, \decplanegen{\thetaA}) = \left| \indof{X}{1}+ \frac{1}{ \indof{\thetaA}{0}} \sum_{i=2}^{ \dims }  \indof{\thetaA}{i} \indof{X}{i} \right|. 
\end{equation*}
Substituting the expression of $\thetaA$ in Equation \ref{eq:robustmaxmarg} yields
\begin{equation*}
d_{e_1}(X, \decplanegen{\thetaA}) = \left| \indof{X}{1} + 2 \marginnonsig \frac{1}{(\sigsep-\epstrain)} \sum_{i=2}^{\dims}  \indof{\thetatilde}{i}   \indof{X}{i} \right|. 
\end{equation*}
Let $N$ be a standard normal distributed random variable. By definition $\| \thetatilde\|_2^2 = 1$ and using that a sum of Gaussian random variables is again a Gaussian random variable, we can write 
\begin{equation*}
d_{e_1}(X,\decplanegen{\thetaA}) = \left| \indof{X}{1} + 2 \marginnonsig \frac{\sigma}{(\sigsep-\epstrain)} N \right|. 
\end{equation*}
The robustness of $\thetaA$ is given by the probability that $d_{e_1}(X,\decplanegen{\thetaA}) > \epstest$. Hence, using that $X_1 = \pm \frac{\sigsep}{2}$ with probability $\frac{1}{2}$, we get
\begin{equation}
\label{eq:robustness_form}
\robness{\thetaA} = P\left[ \frac{\sigsep}{2} + 2 \marginnonsig \frac{\sigma}{(\sigsep-2\epstrain)}  N > \epstest \right] + P \left[ \frac{\sigsep}{2} + 2 \marginnonsig \frac{\sigma}{(\sigsep-\epstrain)}  N < -\epstest \right].
\end{equation}
We can rewrite Equation \ref{eq:robustness_form} in the form
\begin{equation*}
\robness{\thetaA}  = P \left[ N > \frac{(\sigsep-2\epstrain) (\epstest - \frac{\sigsep}{2})}{2 \marginnonsig\sigma} \right] + P \left[  N <  \frac{(\sigsep-2\epstrain)( -\epstest -\frac{ \sigsep}{2})}{2 \marginnonsig\sigma} \right].
\end{equation*}
Recall, that $N$ is a standard normal distributed random variable and denote by $\Phi$ the cumulative standard normal density. By definition of the cumulative denisity function, we find that
\begin{equation*}
\robness{\thetaA} = 1 - \Phi \left(\frac{(\sigsep-2\epstrain) (\epstest - \frac{\sigsep}{2})}{2 \marginnonsig\sigma}\right) + \Phi \left( \frac{(\sigsep-2 \epstrain)( -\epstest - \frac{\sigsep}{2})}{2 \marginnonsig\sigma} \right).
\end{equation*}
Substituting the bounds on $\marginnonsig$ of Lemma \ref{lem:boundsmaxmargin} gives us the non-asymptotic bounds on the robustness score and by Equation \ref{eq:rob_sus} also on the susceptibility score.

\section{Experimental details on the linear model}

\label{sec:logregapp}
In this section, we provide detailed experimental details to Figures \ref{fig:main_theorem} and \ref{fig:lineartradeoff}.

We implement adversarial logistic regression using stochastic gradient descent with a learning rate of $0.01$. Note that logistic regression converges logarithmically to the robust max $l_2$-margin solution. As a consequence of the slow convergence, we train for up to $10^7$ epochs. Both during training and test time we solve $\max_{x_i' \in \pertset{x_i}{\epstrain}} \loss(f_\theta(x_i') y_i)$ exactly. Hence, we exactly measure the robust error. 
Unless specified otherwise, we set $\mixvar= 1$,  $\sigsep = 12$ and $\epstest = 4$. 

\paragraph{Experimental details on Figure \ref{fig:main_theorem}} (a) We draw $5$ datasets with $\numsamp= 50$ samples and input dimension $\dims=1000$ from the distribution $\prob$. We then run adversarial logistic regression on all $5$ datasets with adversarial training budgets, $\epstrain = 1$ to $5$. To compute the resulting robust error gap of all the obtained classifiers, we use a test set of size $10^{6}$. Lastly, we compute the lower bound given in part 2. of Theorem \ref{thm:linlinf}. (b) We draw $5$ datasets with different sizes $\numsamp$ between $50$ and $10^4$. We take an input dimension of $d = 10^4$ and plot the mean and standard deviation of the robust error after adversarial and standard logistic regression over the $5$ samples.(c) We again draw $5$ datasets for each $d/n$ constellation and compute the robust error gap for each dataset.

\paragraph{Experimental details on Figure \ref{fig:lineartradeoff}} For both (a) and (b) we set $\dims = 1000$, $\epstest = 4$, and vary the adversarial training budget ($\epstrain$) from $1$ to $5$. For every constellation of $\numsamp$ and $\epstrain$, we draw $10$ datasets and show the average and standard deviation of the resulting robust errors. In (b), we set $\numsamp = 50$.


\section{Experimental details on the Waterbirds dataset}
\label{sec:waterbirds}
In this section, we discuss the experimental details and construction of the Waterbirds dataset in more detail. We also provide ablation studies of attack parameters such as the size of the motion blur kernel, plots of the robust error decomposition with increasing $\numsamp$, and some experiments using early stopping.

\paragraph{The waterbirds dataset}

To build the Waterbirds dataset, we use the CUB-200 dataset \cite{Welinder10}, which contains images and labels of $200$ bird species, and $4$ background classes (forest, jungle/bamboo, water ocean, water lake natural) of the Places dataset \cite{zhou17}.The aim is to recognize whether or not the bird, in a given image, is a waterbird (e.g. an albatros) or a landbird (e.g. a woodpecker). To create the dataset, we randomly sample equally many water- as landbirds from the CUB-200 dataset. Thereafter, we sample for each bird image a random background image. Then, we use the segmentation provided in the CUB-200 dataset to segment the birds from their original images and paste them onto the randomly sampled backgrounds. The resulting images have a size of $256 \times 256$. Moreover, we also resize the segmentations such that we have the correct segmentation profiles of the birds in the new dataset as well. For the concrete implementation, we use the code provided by \cite{Sagawa20}.

\paragraph{Experimetal training details}
Following the example of \cite{Sagawa20}, we use a ResNet50 pretrained on the ImageNet dataset for all experiments, a weight-decay of $10^{-4}$, and train for $300$ epochs using the Adam optimizer. Extensive fine-tuning of the learning rate resulted in an optimal learning rate of $0.006$ for all experiments in the low sample size regime. Adversarial training is implemented as suggested in \cite{madry18}: at each iteration we find the worst case perturbation with an exact or approximate method. In all our experiments, the resulting classifier interpolates the training set. We plot the mean over all runs and the standard deviation of the mean. 

\begin{figure*}[!t]
\centering
\begin{subfigure}[b]{0.19\textwidth}
  \includegraphics[width=0.99\linewidth]{plotsAistats/waterbird_original_example.png}
  \caption{Original}
  \label{fig:motion_blur_or}
\end{subfigure}
\begin{subfigure}[b]{0.19\textwidth}
  \includegraphics[width=0.99\linewidth]{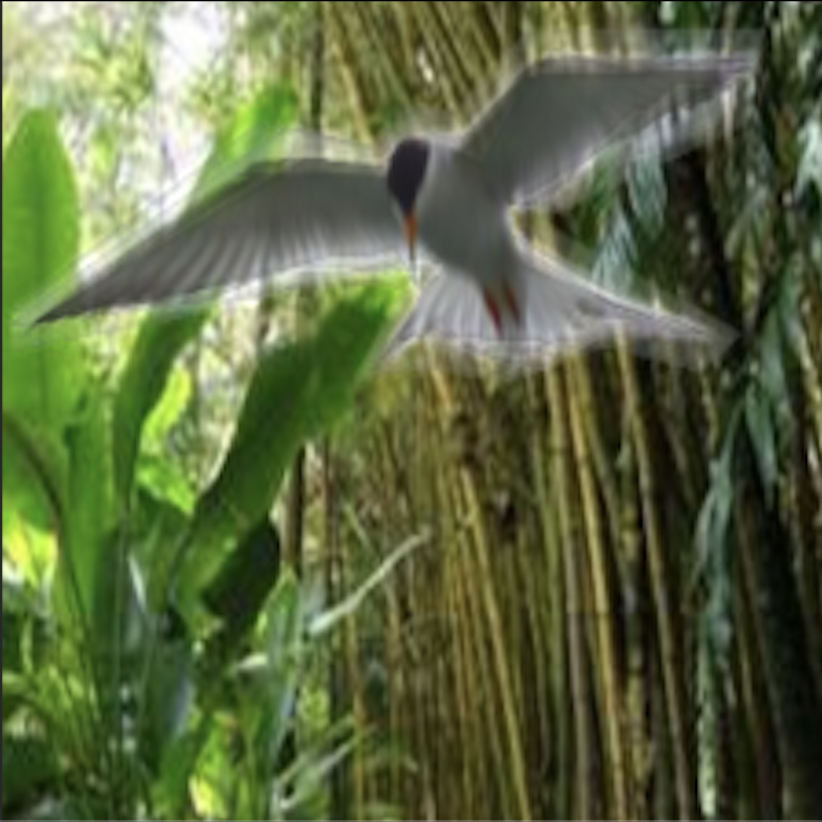}
  \caption{$\motionblurkernel = 5$}
  \label{fig:motion_blur_5}
\end{subfigure}
\begin{subfigure}[b]{0.19\textwidth}
  \includegraphics[width=0.99\linewidth]{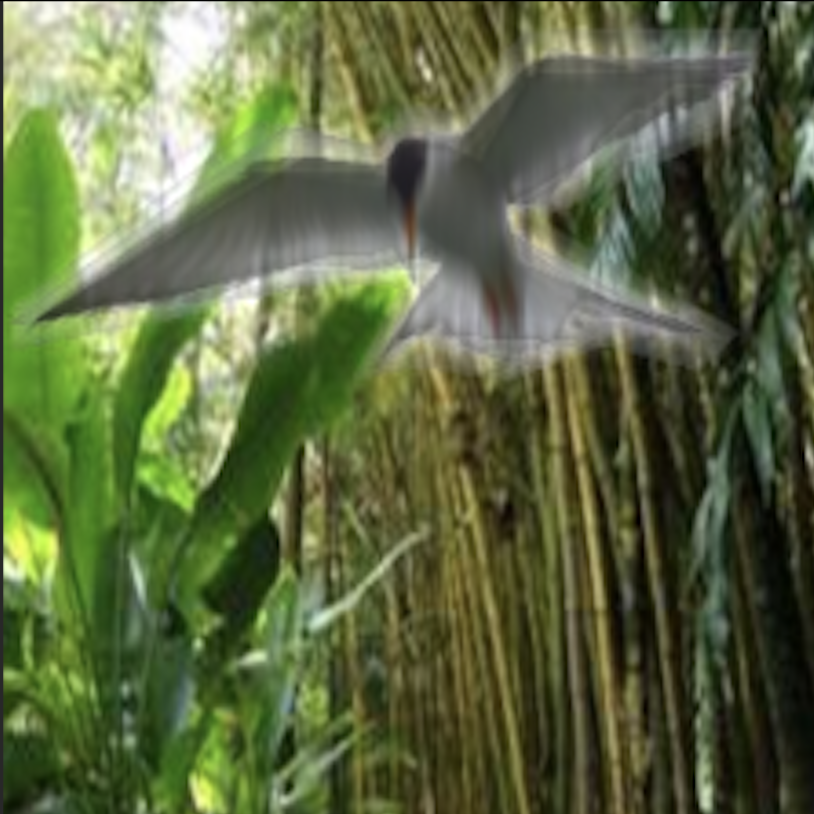}
  \caption{$\motionblurkernel = 10$}
  \label{fig:motion_blur_10}
\end{subfigure}
\begin{subfigure}[b]{0.19\textwidth}
  \includegraphics[width=0.99\linewidth]{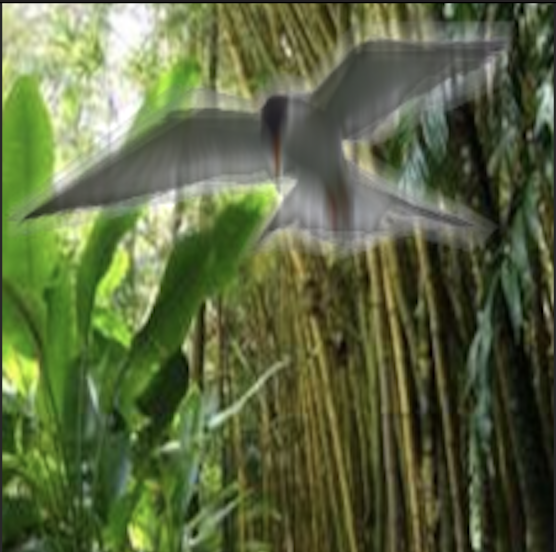}
  \caption{$\motionblurkernel = 15$}
  \label{fig:motion_blur_15}
\end{subfigure}
\begin{subfigure}[b]{0.19\textwidth}
  \includegraphics[width=0.99\linewidth]{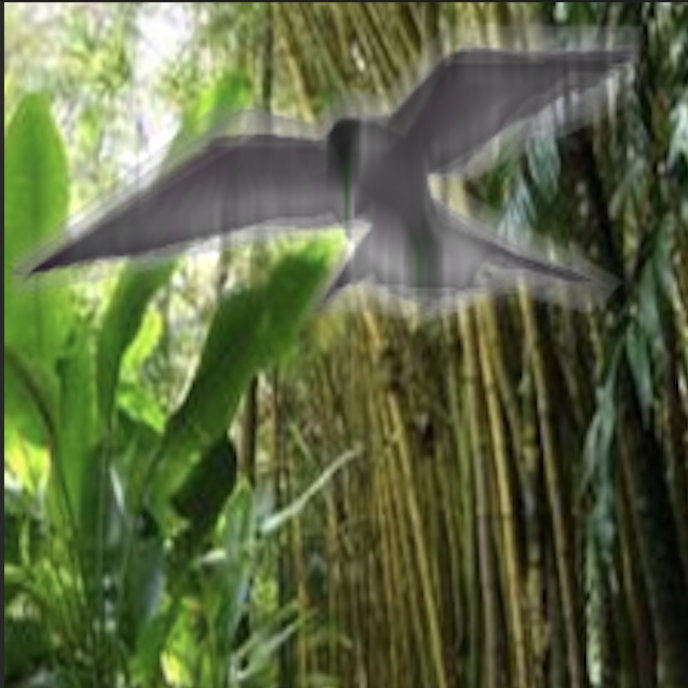}
  \caption{$\motionblurkernel = 20$}
  \label{fig:motion_blur_20}
\end{subfigure}
\caption{We perform an ablation study of the motion blur kernel size, which corresponds to the severity level of the blur. We see that for increasing $\motionblurkernel$, the severity of the motion blur increases. In particular, note that for $\motionblurkernel = 15$ and even $\motionblurkernel = 20$, the bird remains recognizable: we do not semantically change the class, i.e. the perturbations are consistent.}
\label{fig:motion_blur_panel}
\end{figure*}

\paragraph{Specifics to the motion blur attack}
Fast moving objects or animals are hard to photograph due to motion blur. Hence, when trying to classify or detect moving objects from images, it is imperative that the classifier is robust against reasonable levels of motion blur. We implement the attack as follows. First, we segment the bird from the original image, then use a blur filter and lastly, we paste the blurred bird back onto the background. We are able to apply more severe blur, by enlarging the kernel of the filter. See Figure \ref{fig:motion_blur_panel} for an ablation study of the kernel size. 

The motion blur filter is implemented as follows. We use a kernel of size $\motionblurkernel \times \motionblurkernel$ and build the filter as follows: we fill the row $(\motionblurkernel-1)/2$ of the kernel with the value $1/\motionblurkernel$. Thereafter, we use the 2D convolution implementation of OpenCV (filter2D) \cite{opencv_library} to convolute the kernel with the image. Note that applying a rotation before the convolution to the kernel, changes the direction of the resulting motion blur. Lastly, we find the most detrimental level of motion blur using a list-search over all levels up to $\motionblurkernel_{max}$.

\begin{figure*}[!b]
\centering
\begin{subfigure}[b]{0.136\textwidth}
  \includegraphics[width=0.99\linewidth]{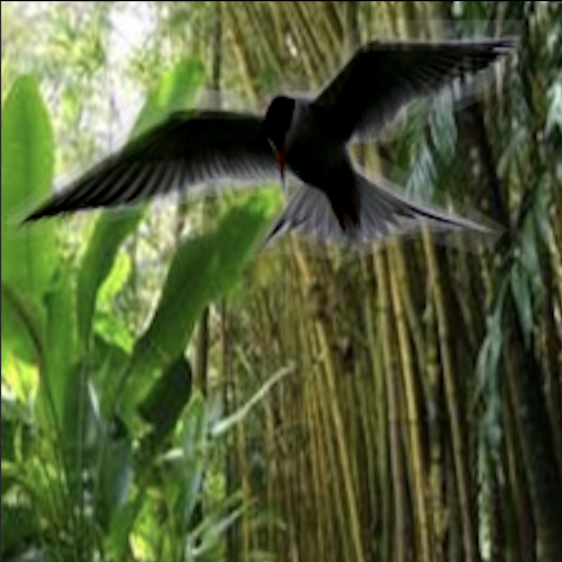}
  \caption{$\epsilon = -0.3$}
  \label{fig:dark_03}
\end{subfigure}
\begin{subfigure}[b]{0.136\textwidth}
  \includegraphics[width=0.99\linewidth]{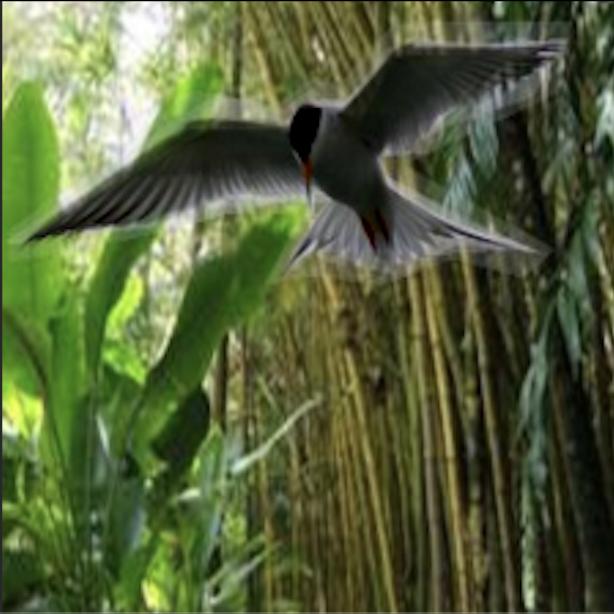}
  \caption{$\epsilon = -0.2$}
  \label{fig:dark_02}
\end{subfigure}
\begin{subfigure}[b]{0.136\textwidth}
  \includegraphics[width=0.99\linewidth]{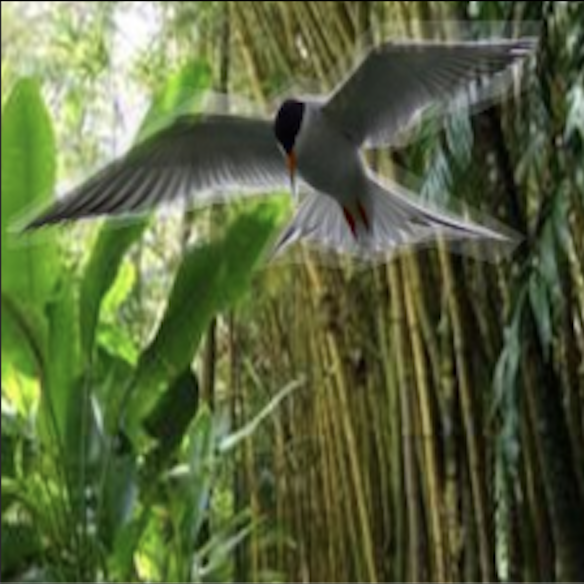}
  \caption{$\epsilon = -0.1$}
  \label{fig:dark_01}
\end{subfigure}
\begin{subfigure}[b]{0.136\textwidth}
  \includegraphics[width=0.99\linewidth]{plotsAistats/waterbird_original_example.png}
  \caption{Original}
  \label{fig:light_or}
\end{subfigure}
\begin{subfigure}[b]{0.136\textwidth}
  \includegraphics[width=0.99\linewidth]{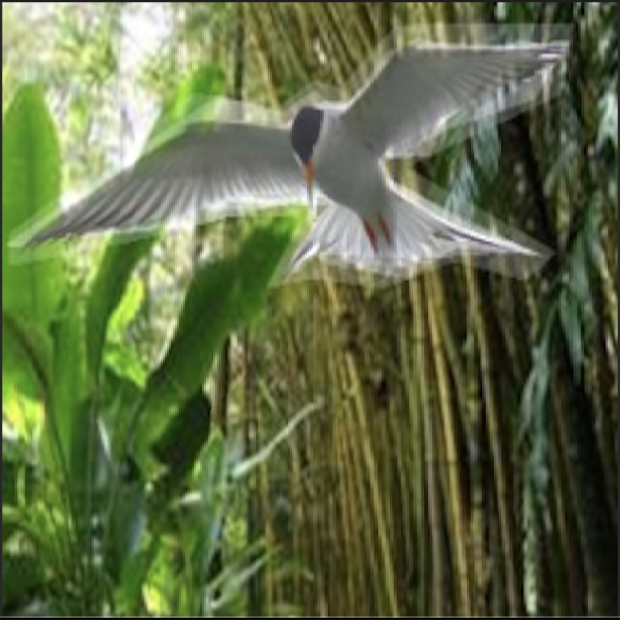}
  \caption{$\epsilon = 0.1$}
  \label{fig:light_01}
 \end{subfigure}
\begin{subfigure}[b]{0.136\textwidth}
  \includegraphics[width=0.99\linewidth]{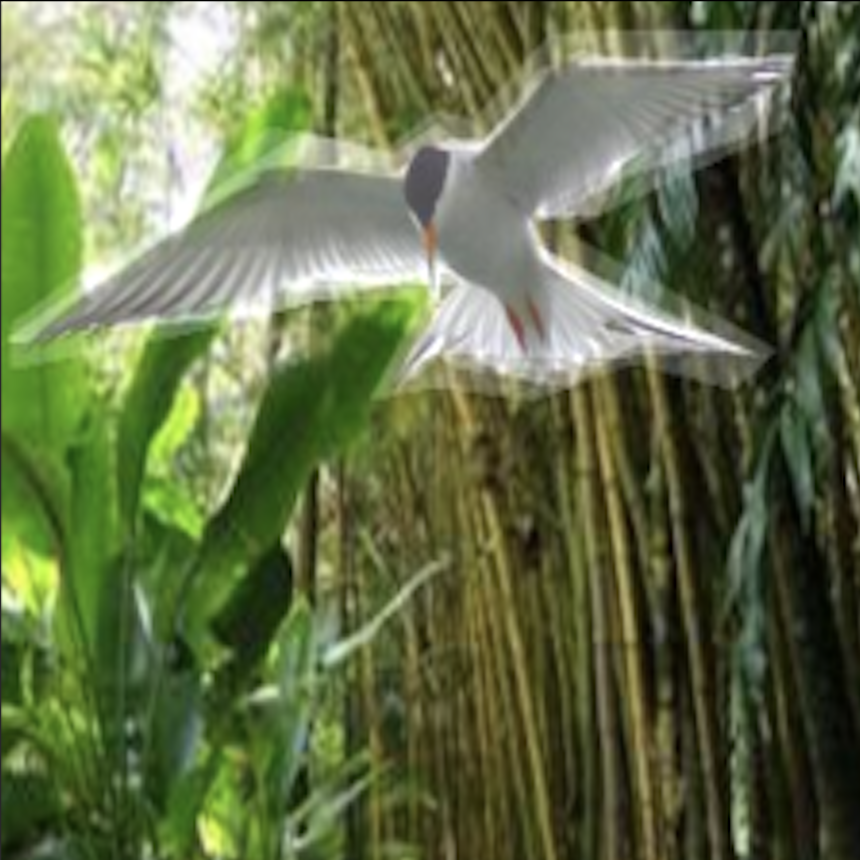}
  \caption{$\epsilon = 0.2$}
  \label{fig:light_02}
\end{subfigure}
\begin{subfigure}[b]{0.136\textwidth}
  \includegraphics[width=0.99\linewidth]{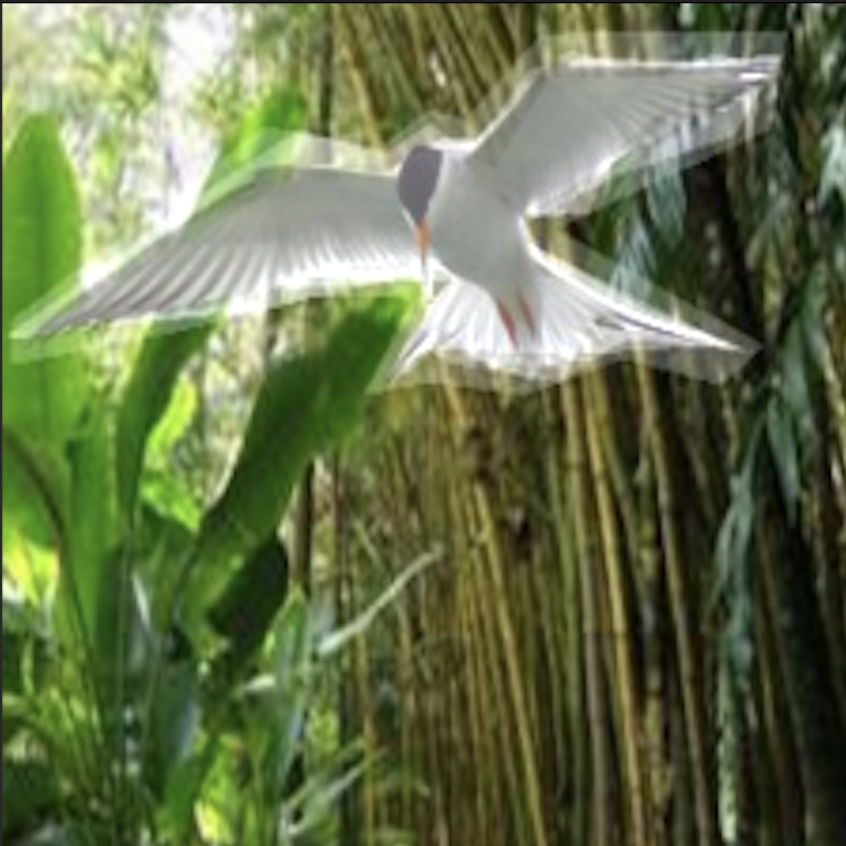}
  \caption{$\epsilon = 0.3$}
  \label{fig:light_03}
\end{subfigure}
\caption{We perform an ablation study of the different lighting changes of the adversarial illumination attack. Even though the \nameofattack attacks the signal component in the image, the bird remains recognizable in all cases.}
\label{fig:light_panel}
\end{figure*}

\paragraph{Specifics to the adversarial illumination attack} 
An adversary can hide objects using poor lightning conditions, which can for example arise from shadows or bright spots. To model poor lighting conditions on the object only (or targeted to the object), we use the adversarial illumination attack. 
The attack is constructed as follows: First, we segment the bird from their background. Then we apply an additive constant $\epsilon$ to the bird, where the absolute size of the constant satisfies $|\epsilon| < \epstest = 0.3$. Thereafter, we clip the values of the bird images to $[0, 1]$, and lastly, we paste the bird back onto the background. See Figure \ref{fig:light_panel} for an ablation of the parameter $\epsilon$ of the attack. It is non-trivial how to (approximately) find the worst perturbation. We find an approximate solution by searching over all perturbations with increments of size $\epstest/K_{\text{max}}$. Denote by \segmentation, the segmentation profile of the image $x$. We consider all perturbed images in the form of
\begin{equation*}
x_{pert} = (1-seg) x + seg (x + \epsilon \frac{K}{K_{\text{max}}}  1_{255 \times 255}), \Hquad K \in [-K_{max}, K_{max}].
\end{equation*} 
During training time we set $K_{max} = 16$ and therefore search over $33$ possible images. During test time we search over $65$ images ($K_{max} = 32$).

\paragraph{Early stopping} In all our experiments on the Waterbirds dataset, a parameter search lead to an optimal weight-decay and learning rate of $10^{-4}$ and $0.006$ respectively. Another common regularization technique is early stopping, where one stops training on the epoch where the classifier achieves minimal robust error on a hold-out dataset. To understand if early stopping can mitigate the effect of adversarial training aggregating robust generalization in comparison to standard training, we perform the following experiment. On the Waterbirds dataset of size $n = 20$ and considering the adversarial illumination attack, we compare standard training with early stopping and adversarial training $(\epstrain = \epstest = 0.3)$ with early stopping. Considering several independent experiments, early stopped adversarial training has an average robust error of $33.5$ a early stopped standard training $29.1$. Hence, early stopping does decrease the robust error gap, but does not close it. 

\paragraph{Error decomposition with increasing $n$}

\begin{figure*}[!t]
\centering
\begin{subfigure}[b]{0.32\textwidth}
  \includegraphics[width=0.99\linewidth]{plotsAistats/numsamp_waterbirds_light.png}
  \caption{Robust error}
  \label{fig:app_waterbirds_robust_error}
\end{subfigure}
\begin{subfigure}[b]{0.32\textwidth}
  \includegraphics[width=0.99\linewidth]{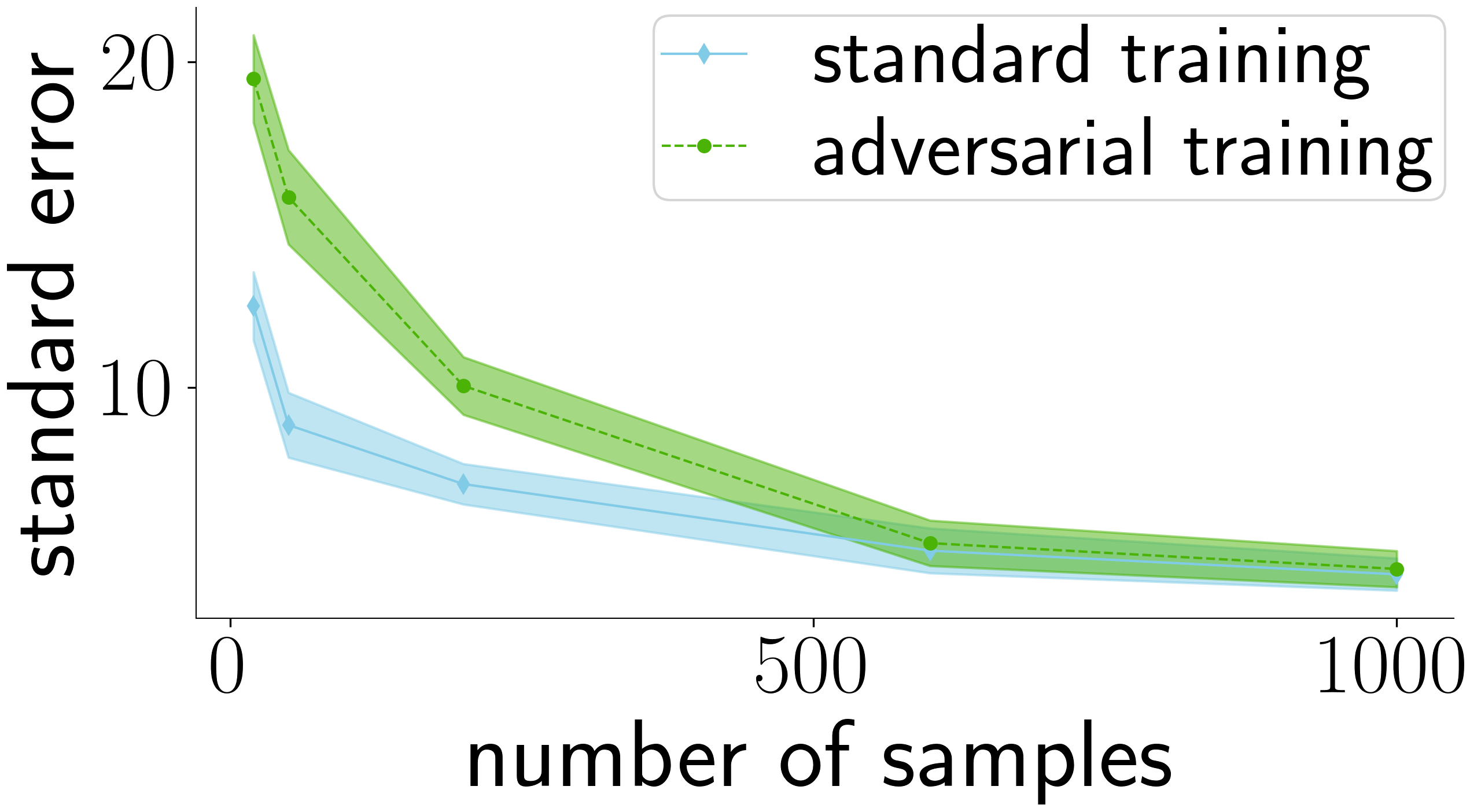}
  \caption{Standard error}
  \label{fig:app_waterbirds_standard_error}
\end{subfigure}
\begin{subfigure}[b]{0.32\textwidth}
  \includegraphics[width=0.99\linewidth]{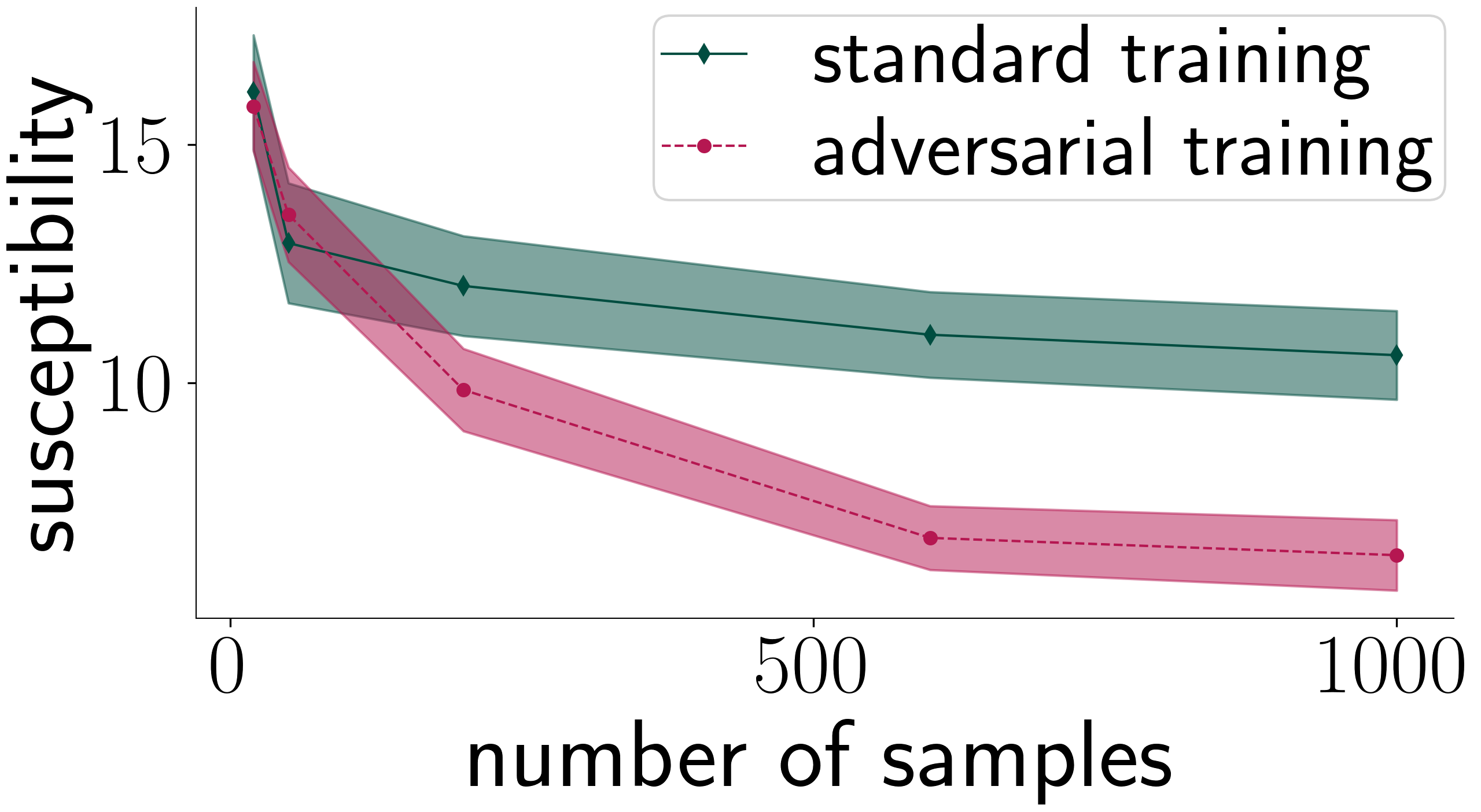}
  \caption{Susceptibility}
  \label{fig:app_waterbirds_susceptibility}
\end{subfigure}
\caption{We plot the robust error decomposition of the experiments depicted in Figure \ref{fig:waterbirds_light_numobs}. The plots depict the mean and standard deviation of the mean over several independent experiments. We see that, in comparison to standard training, the reduction in susceptibility for adversarial training is minimal in the low sample size regime. Moreover, the increase in standard error of adversarial training is quite severe, leading to an overall increase in robust error in the low sample size regime.}
\label{fig:light_numsamp_decomposition}
\end{figure*}

In Figure \ref{fig:waterbirds_light_numobs}, we see that adversarial training hurts robust generalization in the small sample size regime. For completeness, we plot the robust error composition for adversarial and standard training in Figure \ref{fig:light_numsamp_decomposition}. We see that in the low sample size regime, the drop in susceptibility that adversarial training achieves in comparison to standard training, is much lower than the increase in standard error. Conversely, in the high sample regime, the drop of susceptibility from adversarial training over standard training is much bigger than the increase in standard error. 

\section{Experimental details on CIFAR10}
\label{sec:app_cifar10}

\begin{wrapfigure}{r}{0.4\textwidth}
\includegraphics[width=0.99\linewidth]{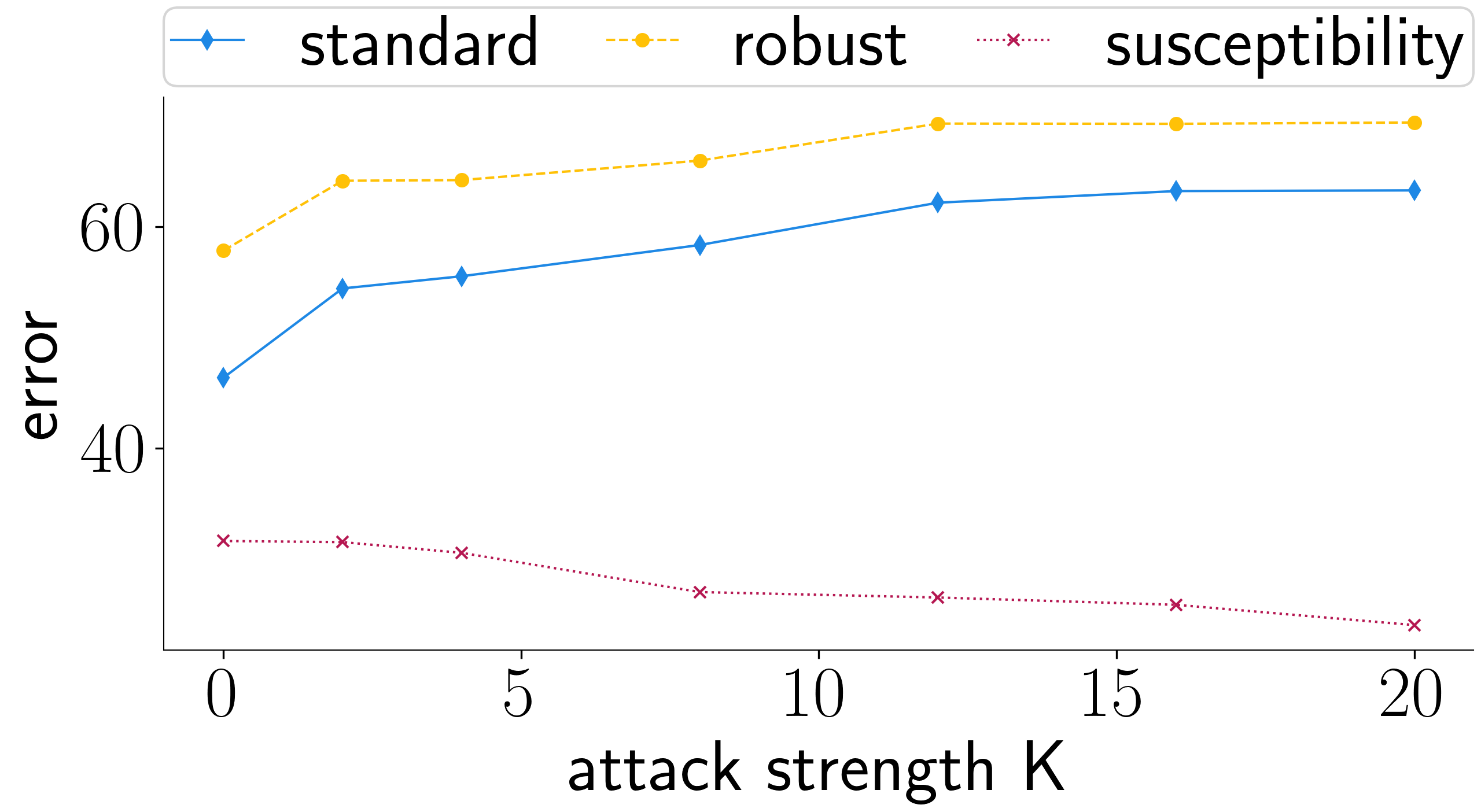}
\caption{We plot the standard error, robust error and susceptibility for varying attack strengths $K$. We see that the larger $K$, the lower the susceptibility, but the higher the standard error.}
\label{fig:K_plot}
\end{wrapfigure}

In this section, we give the experimental details on the CIFAR10-based experiments shown in Figures \ref{fig:teaserplot} and \ref{fig:K_plot}. Moreover, we also conduct similar experiments using different neural network architectures. First, we give the full experimental details and then provide the results of the experiments using the different architectures.

\paragraph{Subsampling CIFAR10}
In all our experiments we subsample CIFAR10 to simulate the low sample size regime. We ensure that for all subsampled versions the number of samples of each class are equal. Hence, if we subsample to $500$ training images, then each class has exactly $50$ images, which are drawn uniformly from the $5k$ training images of the respective class.

\begin{figure}[!ht]
\centering
  \includegraphics[width=0.8\linewidth]{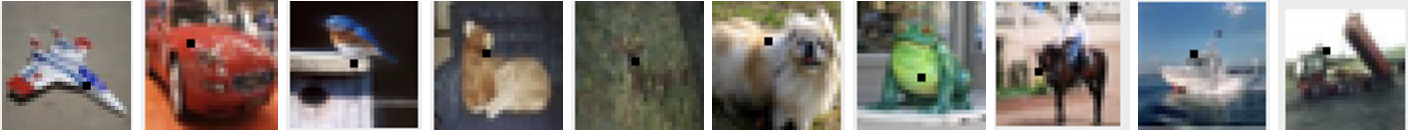}
  \caption{We show an example of a mask perturbation for all $10$ classes of CIFAR10. Even though the attack occludes part of the images, a human can still easily classify all images correctly.}
\label{fig:cifar10_masks}
\end{figure}

\paragraph{Mask perturbation on CIFAR10}
We consider square black-mask perturbations; the attacker can set in the image a patch of size $2 \times 2$ to zero. The attack is a simplification of the patch-attack as considered in \cite{Wu20}. We show an example of a black-mask attack on each of the classes in CIFAR10 in Figure \ref{fig:cifar10_masks}. Clearly, the mask reduces the information about the class in the image as it occludes part of the object in the image.

During test time, we evaluate the attack exactly by means of a full grid search over all possible windows. Note that a full grid search requires $900$ forward passes to evaluate one image, which computationally too expensive during training time. Therefore, we use the same approximation as in \cite{Wu20} at training time. For each image in the training batch, we compute the gradient from the loss with respect to the input. Using that gradient, which is a tensor in $\mathbb{R}^{3 \times 32 \times 32}$, we compute the $l_1$-norm of each patch by a full grid search and save the upper left coordinates of the $K$ windows with largest $l_1$-norm. The intuition is that windows with high $l_1$-norm are more likely to change the prediction. Out of the $K$ identified candidate windows, we take the most loss worsening by means of a full list-search. 

\paragraph{Experimental training details}
For all our experiments on CIFAR10, we adjusted the code provided by \cite{Phan21}. As typically done for CIFAR10, we augment the data with random cropping and horizontal flipping. For the experiments with results depicted in Figures \ref{fig:teaserplot} and \ref{fig:K_plot}, we use a ResNet18 network and train for $100$ epochs. We tune the parameters learning rate and weight decay for low robust error. For standard standard training, we use a learning rate of $0.01$ with equal weight decay. For adversarial training, we use a learning rate of $0.015$ and a weight decay of $10^{-4}$. We run each experiment three times for every dataset with different initialization seeds, and plot the average and standard deviation over the runs. 

\begin{figure*}[!b]
\centering
\begin{subfigure}[b]{0.32\textwidth}
  \includegraphics[width=0.99\linewidth]{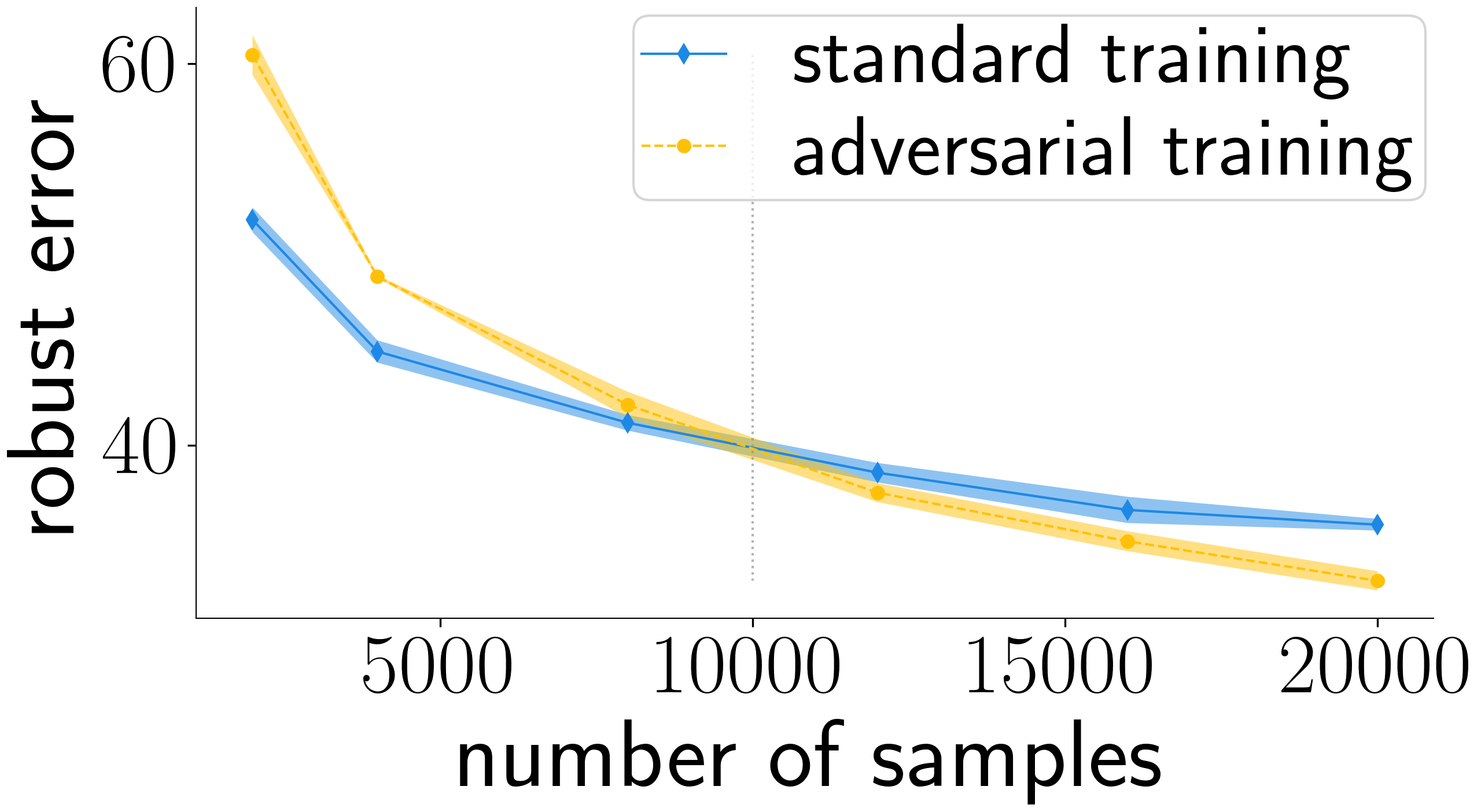}
  \caption{Robust error}
  \label{fig:RA_CIFAR_10_n}
\end{subfigure}
\begin{subfigure}[b]{0.32\textwidth}
  \includegraphics[width=0.99\linewidth]{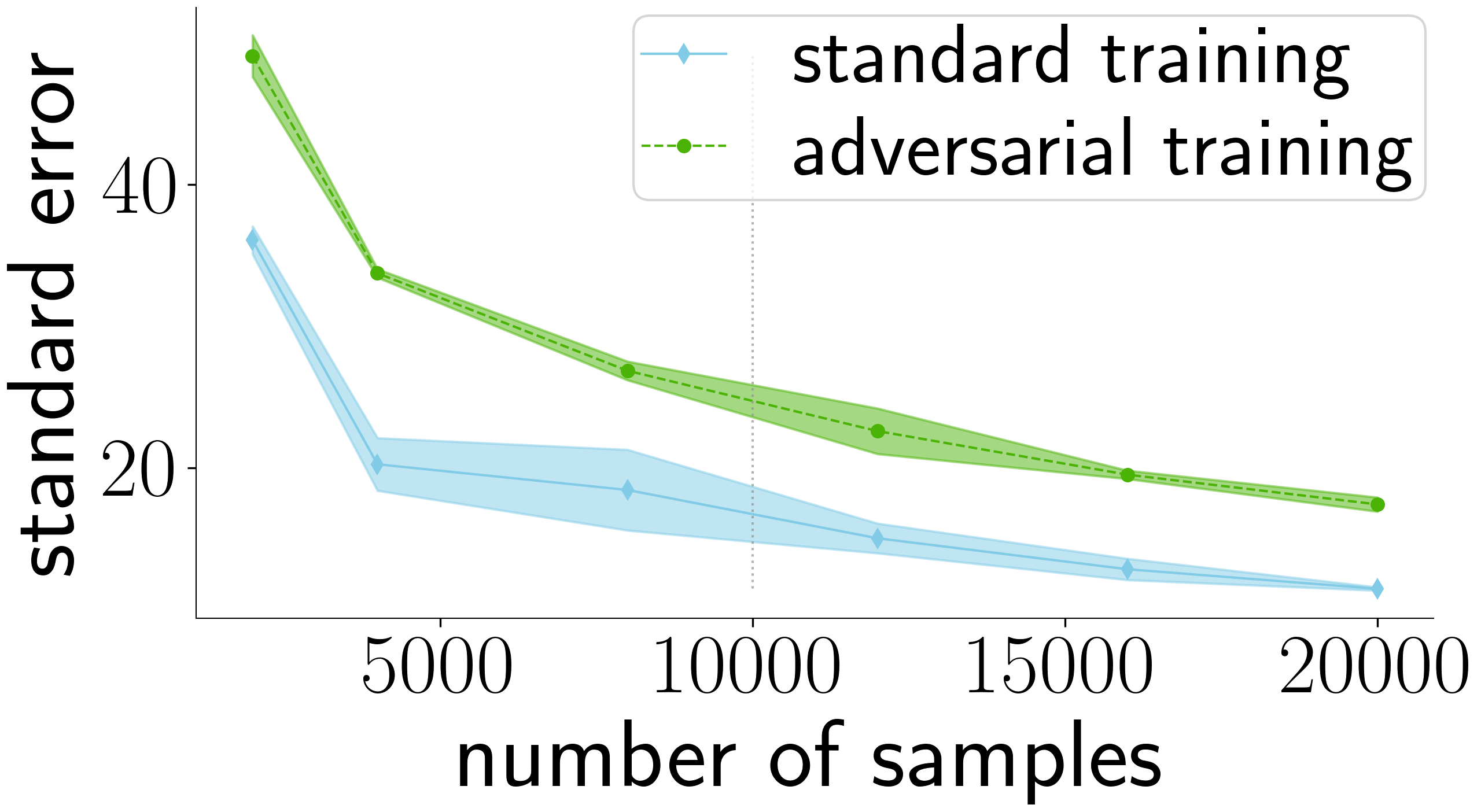}
  \caption{Standard error}
  \label{fig:SA_CIFAR_10_n}
\end{subfigure}
\begin{subfigure}[b]{0.32\textwidth}
  \includegraphics[width=0.99\linewidth]{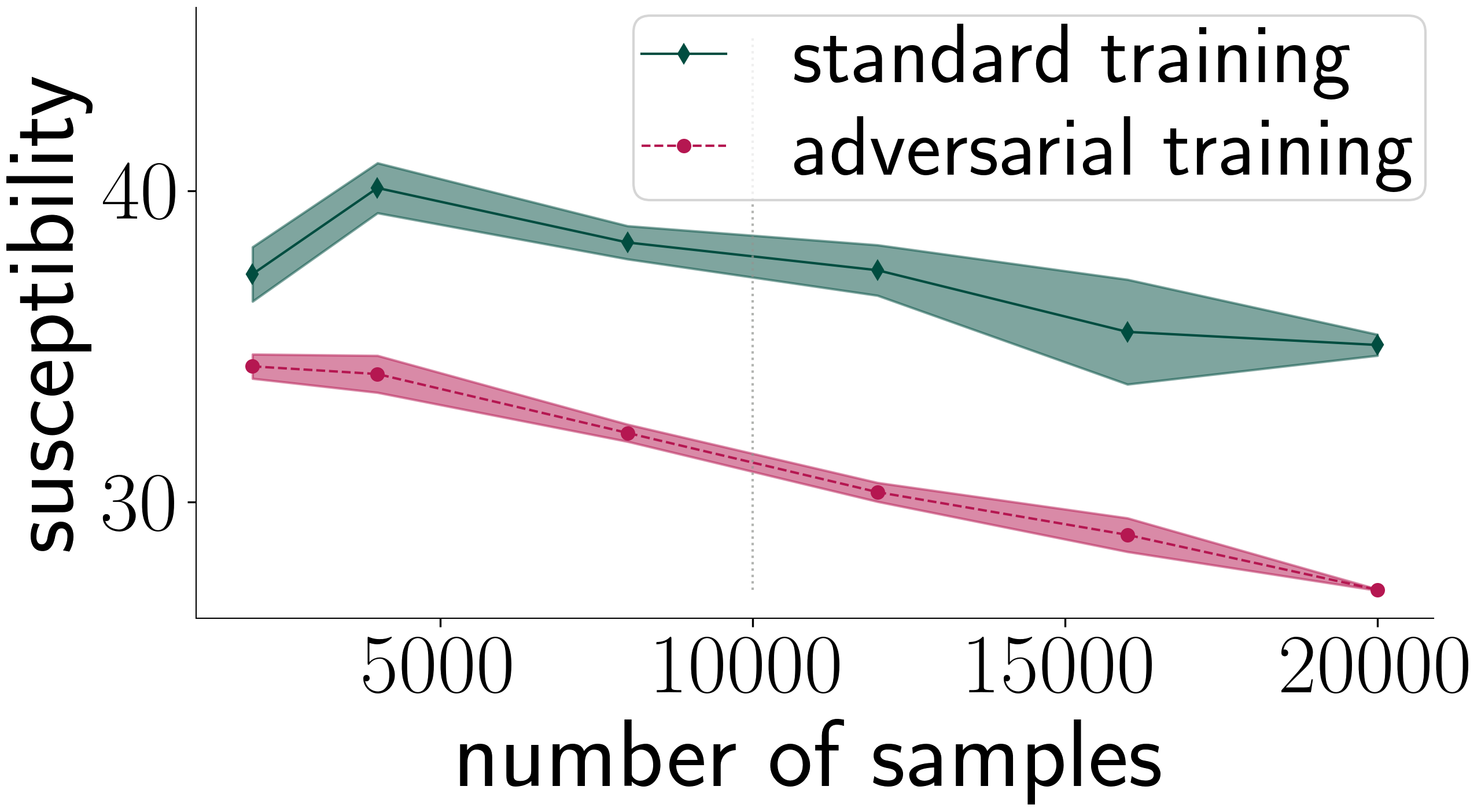}
  \caption{Susceptibility}
  \label{fig:Robustness_n}
\end{subfigure}

\caption{We plot the standard error, robust error and susceptibility of the subsampled datasets of 
CIFAR10 after adversarial and standard training. For small sample size, adversarial 
training has higher robust error then standard training. We see that the increase in standard error in comparison to the drop in susceptibility of standard versus robust training, switches between the low and high sample size regimes.}
\label{fig:num_obs_CIFAR}
\end{figure*}

For the experiments in Figure \ref{fig:teaserplot} and \ref{fig:num_obs_CIFAR} we use an attack strength of $K = 4$. Recall that we perform a full grid search at test time and hence have a good approximation of the robust accuracy and susceptibility score. 

\paragraph{Increasing training attack strength} We investigate the influence of the attack strength $K$ on the robust error for adversarial training. We take $\epstrain = 2$ and $\numsamp = 500$ and vary $K$. The results are depicted in Figure \ref{fig:K_plot}. We see that for increasing $K$, the susceptibility decreases, but the standard error increases more severely, resulting in an increasing robust error.

\paragraph{Robust error decomposition}
In Figure \ref{fig:teaserplot}, we see that the robust error increases for adversarial training compared to standard training in the low sample size regime, but the opposite holds when enough samples are available. For completeness, we provide a full decomposition of the robust error in standard error and susceptibility for standard and adversarial training. We plot the decomposition in Figure \ref{fig:num_obs_CIFAR}.

\paragraph{Multiple networks on CIFAR10}
We run adversarial training for multiple network architectures on subsampled CIFAR10 ($n=500$) with mask perturbations of size $2 \times 2$ and an attack strength of $K=4$.  We plot the results in Table \ref{CIFAR10_diffArchitectures}. For all the different architectures, we notice a similar increase in robust error when trained with adversarial training instead of standard training.

\begin{table}[!ht]
\centering
\caption{We subsample CIFAR10 to a dataset of sample size $500$ and perform both standard training (ST) and adversarial training (AT) using different networks. We evaluate the resulting susceptibility score and the robust and standard error. }
\begin{tabular}{ |p{2cm}||p{2cm}||p{1cm}||p{1cm}|p{2cm}|p{2cm}|p{2cm}|}
 \hline
 \multicolumn{7}{|c|}{Adversarial training on CIFAR10} \\
 \hline
Architecture & learning rate & weight decay & Train type & standard error & robust error & Susceptibility\\
 \hline
 ResNet34 &   $ 0.02$  & $0.025$ &   ST  & 44 & 64 & 50 \\
 ResNet34 &   $0.015$  & $10^{-4}$ &   AT & 52 & 66 & 40\\
 ResNet50 &  $0.015$  & $0.03$  &   ST &  45 & 62 & 47\\
 ResNet50 &  $0.015$  &  $10^{-4}$ &   AT &  53 & 68 & 45\\
VGG11bn &  $0.03$ & $0.01$ & ST & 40 & 55 & 43\\
VGG11bn &   $0.015$  & $10^{-4}$ & AT & 48 &63 & 34\\
VGG16bn &  $0.02$ & $0.01$ & ST & 41 & 60 & 48\\
VGG16bn &   $0.015$  & $10^{-4}$ & AT & 50 & 65  & 42\\
 \hline
\end{tabular}
\label{CIFAR10_diffArchitectures}
\end{table}


\section{Static hand gesture recognition}
\label{sec:handgestures}

The goal of static hand gesture or posture recognition is to recognize hand gestures such as a pointing index finger or the okay-sign based on static data such as images \cite{Oudah20, Yang13}. The current use of hand gesture recognition is primarily in the interaction between computers and humans \cite{Oudah20}. More specifically, typical practical applications can be found in the environment of games, assisted living, and virtual reality \cite{Mujahid21}. In the following, we conduct experiments on a hand gesture recognition dataset constructed by \cite{Mantecon19}, which consists of near-infrared stereo images obtained using the Leap Motion device. First, we crop or segment the images after which we use logistic regression for classification. We see that adversarial logistic regression deteriorates robust generalization with increasing $\epstrain$.

\paragraph{Static hand-gesture dataset}
We use the dataset made available by \cite{Mantecon19}. This dataset consists of near-infrared stereo images taken with the Leap Motion device and provides detailed skeleton data. We base our analysis on the images only. The size of the images is $640 \times 240$ pixels. The dataset consists of $16$ classes of hand poses taken by $25$ different people. We note that the variety between the different people is relatively wide; there are men and women with different posture and hand sizes. However, the different samples taken by the same person are alike.

\begin{figure}
    \centering
    \begin{subfigure}{0.49\textwidth}
    \includegraphics[width=.80\linewidth]{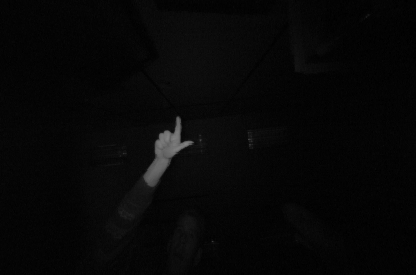}
    \caption{L pose}
    \label{fig:L_pose_or_example}
    \end{subfigure}
    \begin{subfigure}{0.49\textwidth}
    \includegraphics[width=.80\linewidth]{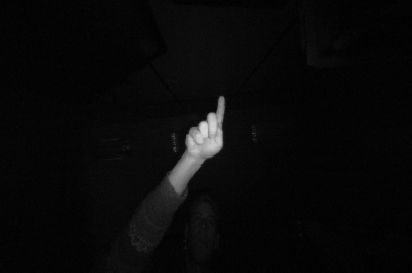}
    \caption{Index pose}
    \label{fig:index_pose_or_example}
    \end{subfigure}
    \caption{We plot two images, where both correspond to the two different classes. We recognize the "L"-sign in Figure \ref{fig:L_pose_or_example} and the index sign in Figure \ref{fig:index_pose_or_example}. Observe that the near-infrared images highlight the hand pose well and blends out much of the non-useful or noisy background. }
\label{fig:original_examples}
\end{figure}

We consider binary classification between the index-pose and L-pose, and take as a training set $30$ images of the users $16$ to $25$. This results in a training dataset of $300$ samples. We show two examples of the training dataset in Figure \ref{fig:original_examples}, each corresponding to a different class. Observe that the near-infrared images darken the background and successfully highlight the hand-pose. As a test dataset, we take $10$ images of each of the two classes from the users $1$ to $10$ resulting in a test dataset of size $200$.

\paragraph{Cropping the dataset}
To speed up training and ease the classification problem, we crop the images from a size of $640 \times 240$ to a size of $200 \times 200$. We crop the images using a basic image segmentation technique to stay as close as possible to real-world applications. The aim is to crop the images such that the hand gesture is centered within the cropped image.

For every user in the training set, we crop an image of the L-pose and the index pose by hand. We call these images the training masks $\{\text{masks}_i \}_{i=1}^{20}$. We note that the more a particular window of an image resembles a mask, the more likely that the window captures the hand gesture correctly. Moreover, the near-infrared images are such that the hands of a person are brighter than the surroundings of the person itself. Based on these two observations, we define the best segment or window, defined by the upper left coordinates $(i,j)$, for an image $x$ as the solution to the following optimization problem:

\begin{equation}
\label{preprocessing}
    \argmin_{i \in [440], \Hquad j \in [40]} \sum_{l=1}^{20}\|\text{masks}_l-x_{\{i:i+200,j:j+200\}}\|^2_2 - \frac{1}{2}\|x_{\{i+w,j+h\}}\|_1.
\end{equation}
Equation \ref{preprocessing} is solved using a full grid search. We use the result to crop both training and test images. Upon manual inspection of the cropped images, close to all images were perfectly cropped. We replace the handful poorly cropped training images with hand-cropped counterparts.

\begin{figure}[!ht]
\centering
\begin{subfigure}{0.31\textwidth}
    \centering
    \includegraphics[width=.80\linewidth]{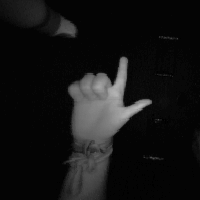}
    \caption{Cropped L pose}
    \label{fig:cropped_L}
\end{subfigure}
\begin{subfigure}{0.31\textwidth}
    \centering
    \includegraphics[width=.80\linewidth]{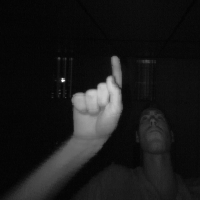}
    \caption{Cropped index pose}
    \label{fig:cropped_index}
\end{subfigure}
\begin{subfigure}{0.31\textwidth}
    \centering
    \includegraphics[width=.80\linewidth]{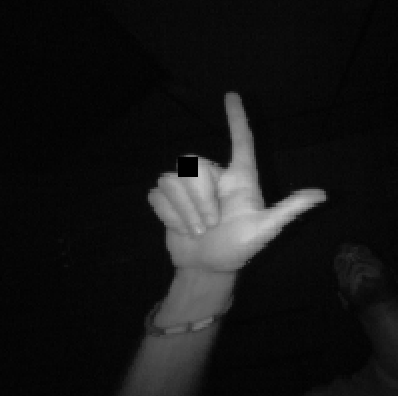}
    \caption{Black-mask perturbation}
    \label{fig:cropped_L_mask}
\end{subfigure}
    \caption{In Figure \ref{fig:cropped_L} and \ref{fig:cropped_index} we show an example of the images cropped using Equation \ref{preprocessing}. We see that the hands are centered and the images have a size of $200 \times 200$. In Figure \ref{fig:cropped_L_mask} we show an example of the square black-mask perturbation.}
    \label{fig:preprocessing}
\end{figure}

\paragraph{Square-mask perturbations}
 Since we use logistic regression, we perform a full grid search to find the best adversarial perturbation at training and test time. For completeness, the upper left coordinates of the optimal black-mask perturbation of size $\epstrain \times \epstrain$ can be found as the solution to
\begin{equation}
\label{square_perturbations_logistic_regression}
    \text{arg}\max_{i \in [200-\epstrain], \Hquad j \in [200-\epstrain]} \sum_{l,m \in [\epstrain]}\theta_{[i:i+l,j:j+m]}.
\end{equation}
The algorithm is rather slow as we iterate over all possible windows. We show a black-mask perturbation on an $L$-pose image in Figure \ref{fig:cropped_L_mask}.

\begin{wrapfigure}{r}{0.4\textwidth}
\includegraphics[width=0.99\linewidth]{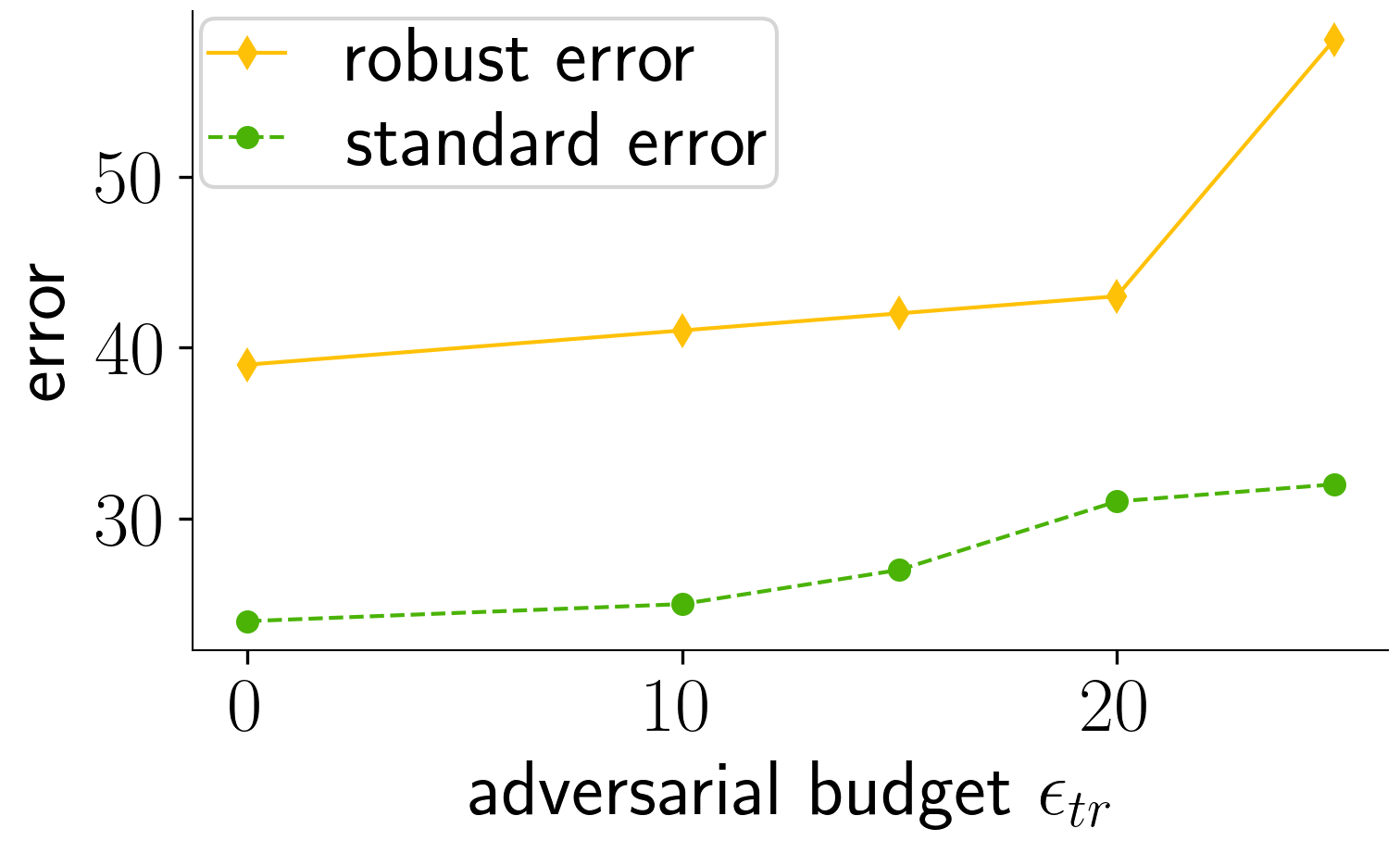}
\caption{We plot the standard error and robust error for varying adversarial training budget $\epstrain$. We see that the larger $\epstrain$ the higher the robust error.}
\label{fig:eps_mask}
\end{wrapfigure}

\paragraph{Results} We run adversarial logistic regression with square-mask perturbations on the cropped dataset and vary the adversarial training budget and plot the result in Figure \ref{fig:eps_mask}. We observe attack that adversarial logistic regression deteriorates robust generalization. 

Because we use adversarial logistic regression, we are able to visualize the classifier. Given the classifier induced by $\theta$, we can visualize how it classifies the images by plotting $\frac{\theta - \min_{i \in [\dims]}\theta_{[i]}}{\max_{i \in [\dims]}\theta_{[i]}} \in [0,1]^{\dims}$. Recall that the class-prediction of our predictor for a data point $(x,y)$ is given by $\text{sign}(\theta^{\top} x) \in \{\pm 1\}$. The lighter parts of the resulting image correspond to the class with label $1$ and the darker patches with the class corresponding to label $-1$.

\begin{figure}[!ht]
\centering
\begin{subfigure}{0.31\textwidth}
    \centering
    \includegraphics[width=.80\linewidth]{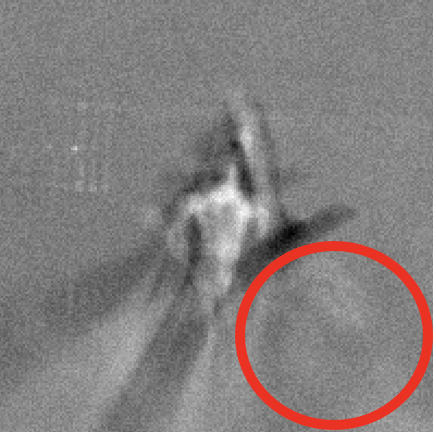}
    \caption{$\epstrain = 0 $}
    \label{fig:log_natural}
\end{subfigure}
\begin{subfigure}{0.31\textwidth}
    \centering
    \includegraphics[width=.80\linewidth]{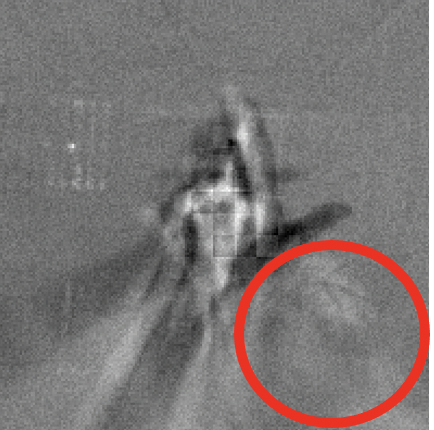}
    \caption{$\epstrain = 10 $}
    \label{fig:log_e10}
\end{subfigure}
\begin{subfigure}{0.31\textwidth}
    \centering
    \includegraphics[width=.80\linewidth]{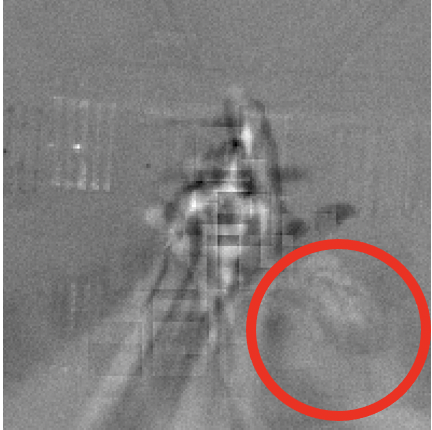}
    \caption{$\epstrain = 25$}
    \label{fig:log_e25}
\end{subfigure}
    \caption{We visualize the logistic regression solutions. In Figure \ref{fig:log_natural} we plot the vector that induces the classifier obtained after standard training. In Figure \ref{fig:log_e10} and Figure \ref{fig:log_e25} we plot the vector obtained after training with square-mask perturbations of size $10$ and $25$, respectively. We note the non-signal enhanced background correlations at the parts highlighted with the red circles in the image projection of the adversarially trained classifiers. }
    \label{fig:visulation_log}
\end{figure}

We plot the classifiers obtained by standard logistic regression and adversarial logistic regression with training adversarial budgets $\epstrain$ of $10$ and $25$ in Figure \ref{fig:visulation_log}. The darker parts in the classifier correspond to patches that are typically bright for the $L$-pose. Complementary, the lighter patches in the classifier correspond to patches that are typically bright for the index pose. We see that in the case of adversarial logistic regression, the background noise is much higher than for standard logistic regression. In other words, adversarial logistic regression puts more weight on non-signal parts in the images to classify the training dataset and hence exhibits worse performance on the test dataset.
 
 \newpage

\end{document}